**Title**

Large Language Models show both individual and collective creativity comparable to humans

**Authors**

Luning Sun[1,2]†, Yuzhuo Yuan[3,4]†, Yuan Yao[4], Yanyan Li[4], Hao Zhang[4], Xing Xie[5], Xiting Wang[5], Fang Luo[4]*, David Stillwell[1,2]*

**Affiliations**

[1]The Psychometrics Centre, University of Cambridge; Cambridge, CB2 1AG, UK.

[2]Cambridge Judge Business School; Cambridge, CB2 1AG, UK.

[3]Collaborative Innovation Center of Assessment for Basic Education Quality, Beijing Normal University, Beijing, 100875, China.

[4]Faculty of Psychology, Beijing Normal University; Beijing, 100875, China.

[5]Microsoft Research Asia; Beijing, 100190, China.

*Corresponding authors. Email: luof@bnu.edu.cn and d.stillwell@jbs.cam.ac.uk.

†These authors contributed equally to this work.

**Abstract**

Artificial intelligence has, so far, largely automated routine tasks, but what does it mean for the future of work if Large Language Models (LLMs) show creativity comparable to humans? To measure the creativity of LLMs holistically, the current study uses 13 creative tasks spanning three domains. We benchmark the LLMs against individual humans, and also take a novel approach by comparing them to the collective creativity of groups of humans. We find that the best LLMs (Claude and GPT-4) rank in the 52$^{nd}$ percentile against humans, and overall LLMs excel in divergent thinking and problem solving but lag in creative writing. When questioned 10 times, an LLM's collective creativity is equivalent to 8-10 humans. When more responses are requested, two additional LLM responses equal one extra human. Ultimately, LLMs, when optimally applied, may compete with a small group of humans in the future of work.

**Introduction**

Large language models (LLMs), such as Generative Pre-trained Transformers (GPTs), are increasingly being piloted in the workplace. It is estimated that 80% of the U.S. workforce belong to an occupation where at least 10% of its tasks could be affected by the introduction of LLMs[1]. According to a World Economic Forum white paper[2], routine and repetitive tasks have the highest potential for automation by LLMs. However, remarkable capabilities of LLMs are being reported across a range of domains and tasks[3], and so the belief that only routine tasks can be performed by LLMs is being overturned[4]. It is speculated that the most affected occupations are in the content creation industry[5], which requires a high level of creativity, a skill that used to be considered unique to humans[6]. Therefore, to anticipate and plan for the future of work we need to answer the question of how creative LLMs are.

Creativity is a multi-facet construct and generally defined as the generation of products or ideas that are both novel and useful[7,8]. Guilford and Vaughan[9] identified four creativity dimensions: fluency, flexibility, originality, and elaboration, which were subsequently measured in the Torrance Tests of Creative Thinking, one of the most used creativity tests[10]. It is also widely accepted that creativity performance may vary depending on the specific domain[11,12]. Therefore, the PISA 2022 Creative Thinking assessment (https://www.oecd.org/pisa/innovation/creative-thinking/) includes four different domains: written expression, visual expression, social problem solving, and scientific problem solving.

There have been several attempts at comparing LLMs with humans on creative tasks. As summarised in Table 1[13-33], the findings are not always consistent. While LLMs achieve comparable or stronger performance on average than humans on divergent thinking tasks, creative writing tasks seem to be more challenging for LLMs. One may argue that average creativity performance is not important because usually only one idea or creative output (or a few) will be put into practice. In terms of the best performance, mixed results are reported as well.

**Table 1. Existing studies benchmarking LLMs' creativity against humans (by 1st November 2024).**

| Study | Task | Main findings |
|---|---|---|
| Haase & Hanel (2023)[13] | Alternate uses task | • No mean difference was found between human and AI-generated ideas in terms of human-rated originality.<br>• 9.4% of humans were more creative than GPT-4. |
| Koivisto & Grassini (2023)[14] | Alternate uses task | • AI chatbots performed better than humans on average in terms of mean scores and max scores.<br>• AI chatbots did not consistently outperform the best human performers. |
| Stevenson, Smal, Baas, Grasman, & can der Maas (2022)[15] | Alternate uses task | • Human responses were rated higher on originality and surprise.<br>• GPT-3's responses were rated as more useful. |
| Cropley (2023)[16] | Divergent association task | • ChatGPT (GPT-3.5 and GPT-4) had statistically significant, higher mean scores than humans. |
| Girotra, Meincke, Terwiesch, & Ulrich (2023)[17] | Idea generation task | • In comparison to human-generated ideas, the average quality of ideas generated by GPT-4 was higher as measured by purchase intent and lower as measured by rated novelty.<br>• Majority (87.5%) of the best ideas in the pooled sample were generated by GPT-4 and not by humans. |
| Guzik, Byrge, & Gilde (2023)[18] | Torrance Tests of Creative Thinking - Verbal | • GPT-4 scored within the top 1% for originality and fluency. A significant difference was found between GPT-4 and humans.<br>• Overall flexibility scores were higher for GPT-4 than humans, although GPT-4 scored lower on flexibility on certain activities. |
| Vicente-Yagüe-Jara, López-Martínez, Navarro-Navarro, & Cuéllar-Santiago (2023)[19] | Test of Creative Imagination for Adults | • AI systems scored higher than humans on indicators of fluency, flexibility and originality in Game 2, but no statistically significant differences were found in the indicators of Game 3. |
| Chakrabarty, Laban, Agarwal, Muresan, & Wu, (2024)[20] | Torrance Test of Creative Writing | • LLM-generated stories showed lower passing rates (between a third and a tenth) than stories written by professionals. |
| Cox, Abdul, & Ooi (2023)[21] | Generating motivational messages | • GPT-4 did not produce a corpus of messages as diverse as those from humans. |
| Tian et al. (2024)[22] | Unconventional everyday problems | • LLMs with single effort achieved lower chances of success than humans.<br>• GPT-4 with multiple efforts underperformed humans in terms of both average and best performance. |

| Summers-Stay, Lukin, & Voss (2023)[23] | Alternate uses task | • GPT-3 achieved higher than average human performance when given a sequence of prompts that included both brainstorming and selection phases. |
|---|---|---|
| Orwig, Edenbaum, Greene, & Schacter (2024)[24] | Five-sentence creative story task | • Compared to human stories, stories generated by both GPT-3 and GPT-4 scored lower in creativity, though this difference was not significant. |
| Nath, Dayan, & Stevenson (2024)[25] | Alternate uses task | • LLMs scored higher on overall mean response sequence originality compared to humans. |
| Chen & Ding (2023)[26] | Divergent association task | • When using the greedy search, GPT-4 outperformed 96% of humans, while GPT-3.5-turbo exceeded the average human level. |
| Castelo, Katona, Li, & Sarvary (2024)[27] | Idea generation task | • Ideas generated by GPT-4 were rated as more creative than those generated by laypeople and creative professionals.<br>• GPT-4 outperformed humans in both creative form and creative substance. |
| Hubert, Awa, & Zabeline (2024)[28] | Alternate uses tasks, consequences task, & divergent associations task | • GPT-4 was more original and elaborate than humans on each task, even when controlling for fluency of responses. |
| Marco, Gonzalo, & Rello (2023)[29] | Creative writing (synopsis) | • Synopses produced by transformers (BART, which is not an LLM) were 3% less creative than human synopses, which was not statistically significant. |
| Grassini & Koivisto (2024)[30] | Figural Interpretation Quest | • GPT-4 on average demonstrated higher flexibility in generating creative interpretations but lower creativity than humans.<br>• The most creative human responses were higher than those of AI in both flexibility and subjective creativity. |
| Bellemare-Pepin et al. (2024)[31] | Divergent association task and creative writing tasks | • GPT-4 was the only LLM that outperformed humans on the divergent association task.<br>• Humans outperformed all LLMs on the creative writing tasks. |
| Gómez-Rodríguez & Williams (2023)[32] | Creative writing | • Human writers outperformed all LLMs on creativity and originality.<br>• On overall rating no significant differences between humans and the top 6 LLMs. |
| Si, Yang, & Hashimoto (2024)[33] | Research ideation | • LLM-generated ideas were judged as more novel than human expert ideas. |

These existing studies suffer from one major limitation, as they are mostly based on one task from a single domain, which is not sufficient to shed light on the overall picture of LLMs' creativity. This study takes a multi-facet approach to the measurement of LLMs' creativity. Specifically, we designed multiple tasks tapping into different dimensions of creativity (e.g., novelty, usefulness, originality, flexibility, and diversity) in three distinctive domains, namely divergent thinking, problem solving, and creative writing. To our knowledge there is no existing dataset about any of these tasks available online that could have been included in the LLMs' training set, reducing concerns about data contamination[34-36]. We compared the responses collected from several LLMs with those from a large, diverse group of human participants, who completed the tasks as part of a high-stakes Master's degree admission assessment. All responses were rated by humans following the Consensual Assessment Technique[37].

Another limitation of existing studies is that all of them focus on individual responses of LLMs. Considering that LLM users frequently request multiple responses for one problem, in the second part of this study we turn to the concept of collective creativity, which refers to the novelty and usefulness of ideas developed by a group of individuals[38,39]. To operationalise the measurement of collective creativity, we requested 10 responses from each LLM and examined the contribution of top ideas when these LLM responses were combined with those by a group of humans. By varying the number of humans in the group, we were able to quantitatively assess LLMs' collective creativity and understand how many humans the LLMs equalled. Based on results of both individual and collective creativity, we discuss the potential of LLMs as capable assistants for creative tasks in the future of work.

**Results**

We used 13 creative tasks spanning over three distinctive domains, including divergent thinking, problem solving, and creative writing. A total of 467 human participants from China (330 female, 137 male; age ranging from 21 to 53 years, with a mean of 27.56 and SD of 6.03; all reported to have at least a Bachelor's degree) completed the tasks as part of an admission assessment for a Master's degree. The same tasks were administered to GPT-3.5 and GPT-4 (https://openai.com) in April and May, 2023, respectively, via OpenAI's API, where responses were collected under five different temperature settings. To probe potential differences in the LLMs, we further recorded responses from Claude (https://claude.ai) as well as two LLMs developed in China, namely Qwen (http://tongyi.aliyun.com) and SparkDesk (https://xinghuo.xfyun.cn) using the relevant web portal in June, 2023. We chose two LLMs developed in China to ensure that our evaluation of LLMs' performance of creative tasks was not biased by the language, since the tasks were administered in Chinese.

Responses of both humans and LLMs were rated by five trained judges independently, who were not aware of which responses were written by whom. Inter-rater reliability was acceptable-to-good (with above .7 Fleiss' Kappa or Intraclass Correlations). Detailed information about the data collection and rating procedure can be found in the Materials and Methods section and Table S1. Fig. 1 and Figs. S1-S3 show the frequencies of the creativity ratings for each task.

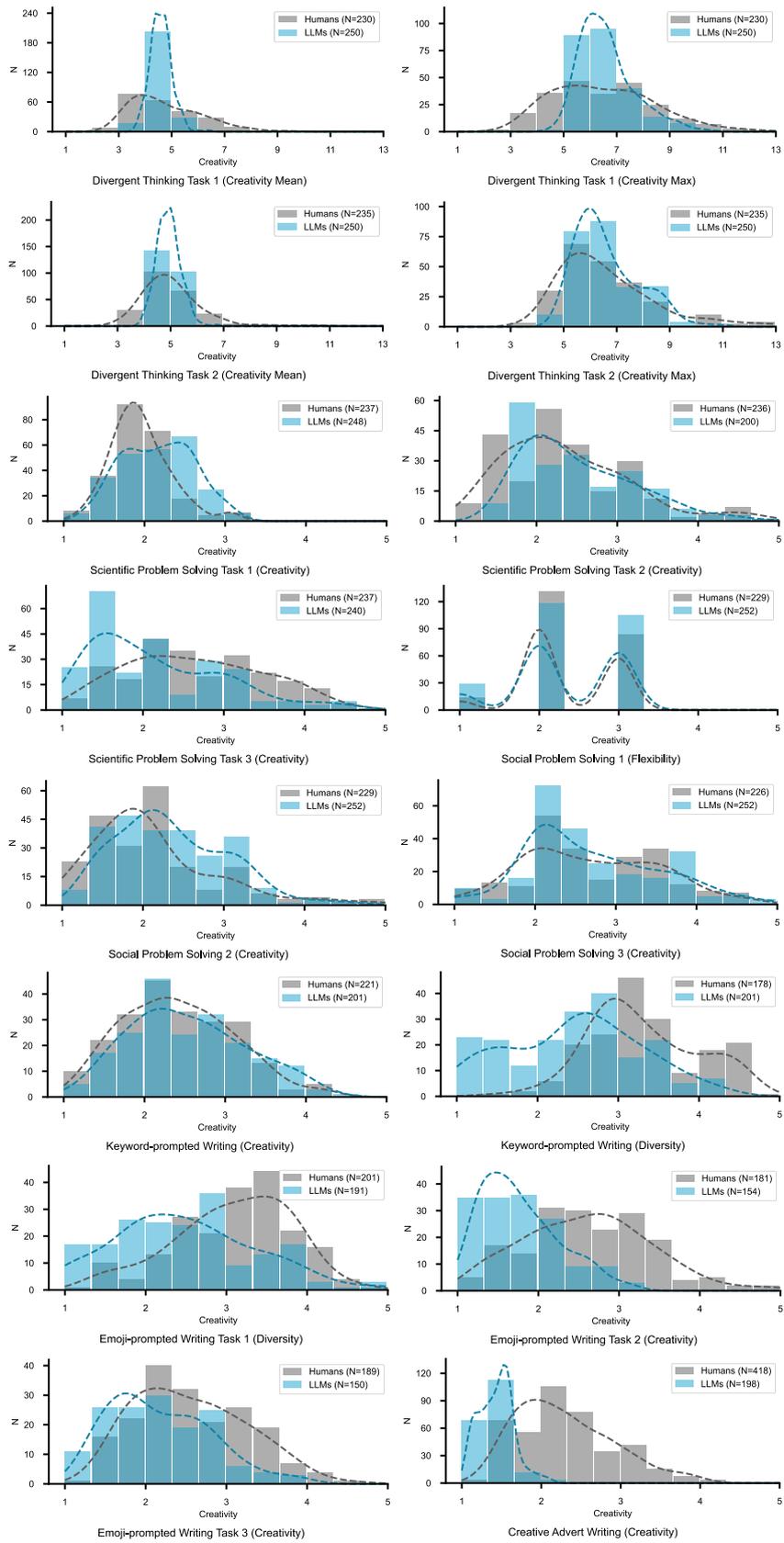

**Fig. 1. Frequencies of the creativity ratings for all creative tasks.**

Below we present the overall results benchmarking average performance of each LLM against the human participants, followed by a more detailed description of the results in each domain. Then we move our focus onto the top-rated responses for each task and evaluate LLMs' capability of generating the most creative ideas, based on which we investigate LLMs' collective creativity.

*Part 1: Individual Creativity*

*Benchmarking LLMs against humans*

To understand how creative the LLMs are in comparison to humans, we rank all the human participants on their performance in each task, and then benchmark the LLMs against humans in terms of their percentiles in the humans' distributions. As shown in Fig. 2, across all indicators of the 13 tasks, the five LLMs ranked on average in the 46th percentile against the human participants. Claude and GPT-4 ranked in the 52nd percentile, SparkDesk in the 44th percentile, Qwen in the 42nd percentile, and GPT-3.5 in the 37th percentile. Domain-wise results suggest that the LLMs performed well in the divergent thinking and problem solving tasks, as on average they ranked in the 55th and 59th percentile. None of the LLMs reached 50th percentile in the domain of creative writing (in the 25th percentile on average), which is consistent with findings reported in previous studies[20].

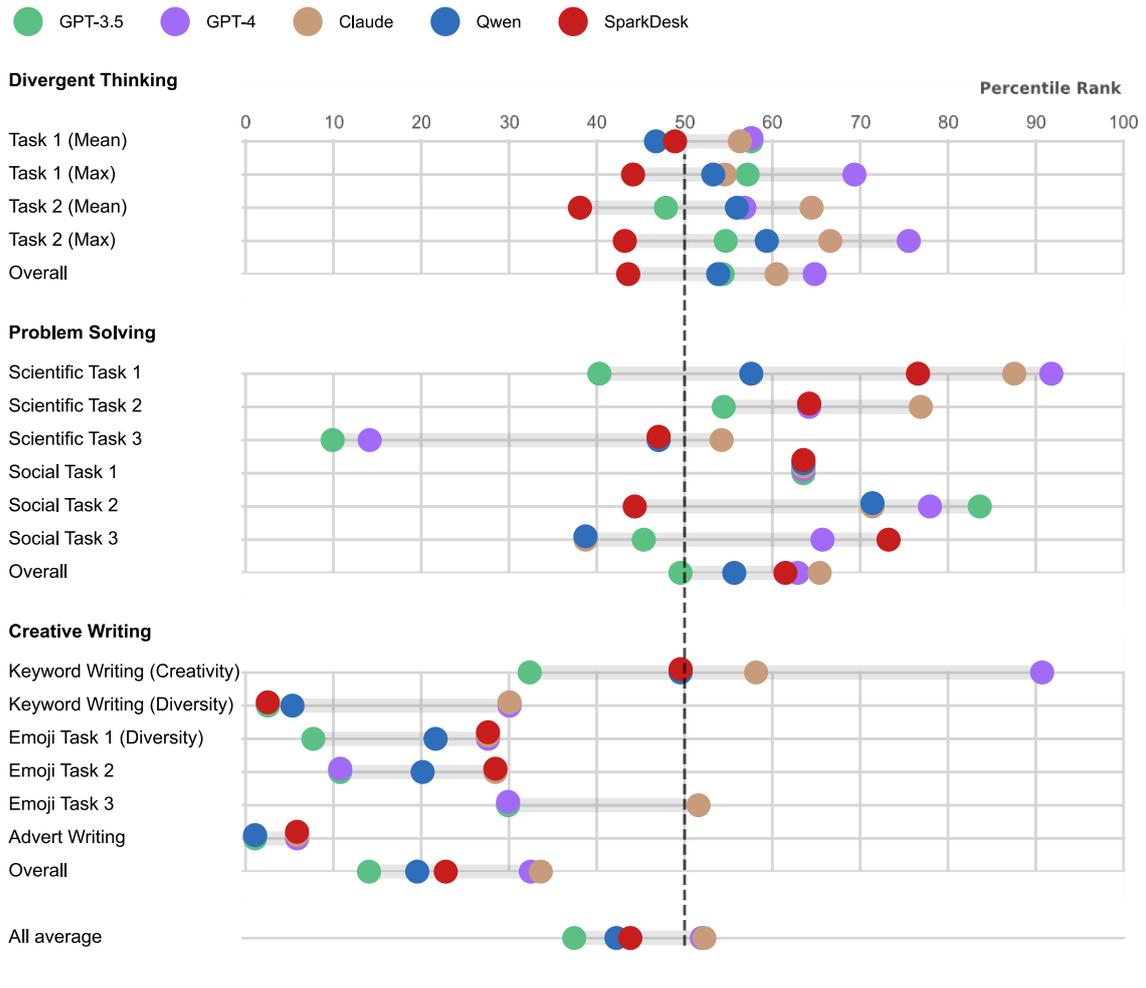

**Fig. 2. Percentile ranks of LLMs in the distribution of human participants on each creative task.**

*Divergent thinking*

Unlike existing studies that mainly make use of Alternate Uses Tasks, we designed two scenario-based divergent thinking tasks, where participants were asked to generate as many ideas as possible within a limited period of time. On average each human participant generated 3.68 valid ideas. In contrast, LLMs generated a mean of 8.85 valid ideas in each response, ranging from 5.43 in SparkDesk to 12.47 in GPT-4.

All ideas were rated on two dimensions: novelty and usefulness[7,8]. In both tasks, all LLMs showed comparable or higher novelty and usefulness than humans (except SparkDesk on Novelty in Task 1 and Qwen on Usefulness in Task 2; see details in Table S2). In both LLMs and humans, ideas appearing later in each response tend to be rated as more novel but less useful (see Figs. S4-S8). This is also confirmed by the paired sample t-tests, which suggest that ideas in the second half of each response show significantly higher ratings of novelty but lower ratings of usefulness (see details in Table S3). Notably, a significant, negative correlation was observed (see Fig. 3) in the LLMs (-0.43 on Task 1 and -0.81 on Task 2) and in the humans (-0.44 on Task 2). Given this trade-off between novelty and usefulness[40,41], we used the product of novelty and usefulness ratings as a proxy of creativity for each idea in the subsequent analysis.

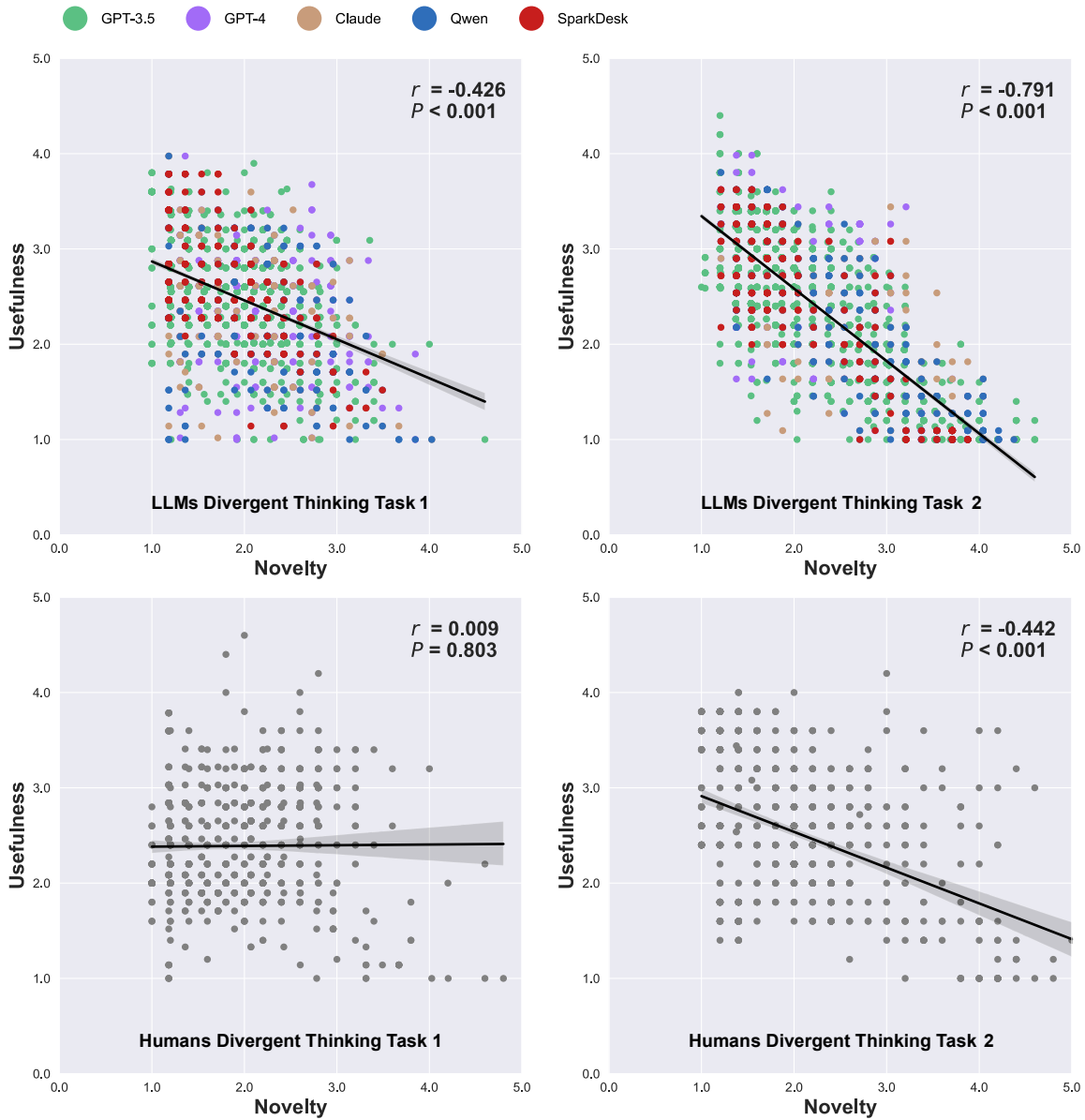

**Fig. 3. Scatter plots of the novelty and usefulness ratings for ideas generated by the LLMs and humans in Divergent Thinking Tasks 1 and 2.**

In terms of the mean score for each response, humans were more creative than Qwen and SparkDesk on Task 1 and GPT-3.5 and SparkDesk on Task 2. If we use the most creative idea as the indicator for each response, GPT-4 would become significantly more creative than humans on both tasks, while SparkDesk on both tasks and GPT-3.5 on Task 2 underperformed humans. All other comparisons were not statistically significant (see Table S4 for details).

*Problem solving*

In this domain, three scientific problem solving tasks and three social problem solving tasks were included, following examples of the PISA 2022 Creative Thinking assessment. Participants were asked to come up with either one or three ideas to solve a real-life problem within limited time. One of the tasks measured the creativity dimension of flexibility, whereas the other five focused on the dimension of originality. Notably, Qwen refused to answer Scientific Problem Solving Task 2, likely due to the mention of "theft" in the question.

Independent sample *t*-tests between LLMs and humans show mixed results (see Table S4). In scientific problem solving, GPT-3.5 underperformed humans on Tasks 1 and 3, whereas GPT-4 exhibited better performance than humans on Task 1 and worse performance on Task 3. Claude, Qwen, and SparkDesk showed comparable or better performance than humans across all three tasks. In social problem solving, GPT-4 outperformed humans on all three tasks. GPT-3.5 showed better performance on Task 2 but worse performance on Task 3. SparkDesk showed exactly the opposite pattern. Both Claude and Qwen failed to outperform humans on all tasks. In a nutshell, no clear pattern is observed: depending on the specific task and the model, LLMs may outperform, underperform, or show comparable performance to humans.

*Creative writing*

We introduced five creative writing tasks using different forms of writing prompts. Specifically, participants were asked to create an advert, write two different stories according to a set of keywords and a set of emojis, and write a full story and a continuation of an existing story based on several emojis provided, taking the example of the PISA 2022 Creative Thinking assessment. The judges provided ratings on their creativity and additionally on diversity in the two tasks where two different stories were required.

The comparison between LLMs and humans (see Table S4) suggests that humans outperformed all LLMs in the Creative Advert Writing Task. Humans' slogans were more emotionally touching and memorable whereas those by LLMs tended to be generic statements. In the Keyword-prompted Writing Task, compared to humans, GPT-3.5 showed lower creativity, GPT-4 higher creativity, whereas all other models had comparable creativity. In the Emoji-prompted Writing Tasks, all LLMs failed to outperform humans. Notably, Qwen and SparkDesk were not able to complete the story continuation task.

In terms of the pairwise diversity in both tasks requiring two stories, judges rated significantly higher diversity for stories generated by humans than those generated by the LLMs except for those generated by Claude in the Keyword-prompted Writing Task.

*Diversity within and between responses*

To replicate findings of recent studies[42], we further investigated if the responses generated the LLMs are less diverse than those by humans. In the six tasks where more than one idea (or story in the creative writing tasks) is required, we calculated the text similarity using both cosine similarity and Levenshtein distance to quantify the diversity among the ideas in each response generated by the participants and LLMs. The comparisons (see Table S5 for details) indicate that in most cases, the ideas within each response generated by LLMs showed higher similarity than those by humans. Furthermore, we examined the text similarity between the responses in all tasks and discovered that consistently across all tasks, responses generated by humans exhibited the lowest text similarity (see Table S6). These results suggest less variation or diversity both within and between the responses generated by the LLMs, replicating recent findings that in comparison to humans, LLM outputs lack diversity and result in the homogenising effect[43,44].

*Effect of temperature*

Temperature is a parameter that controls how random or diverse the output of an LLM is, which might be expected to be related to creativity. As shown in Fig. 4 (and see Tables S7 and S8), the effect of temperature on creativity performance was not consistent. The optimal temperature varied among the tasks even within the same domain. On average, there seems to be an upward trend for GPT-3.5, with temperature 1 generating more creative responses than temperature 0, whereas for GPT-4 no clear pattern could be identified. Therefore, while it might be possible to improve the performance of LLMs in creative tasks by changing the temperature, there is no simple rule for what the optimal temperature should be across creative tasks. Unless specified, all results for GPT-3.5 and GPT-4 presented in this paper are aggregated across responses at different temperatures.

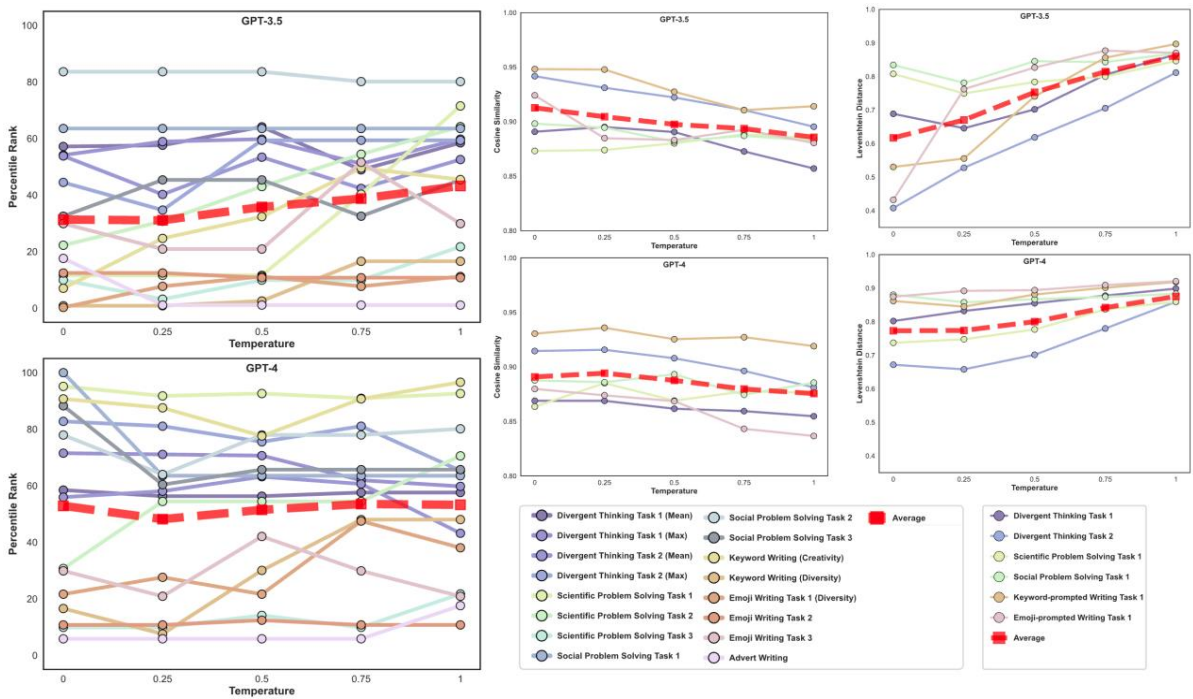

**Fig. 4. Performance rank (left) and diversity in the responses (right) of GPT-3.5 (top) and GPT-4 (bottom) in the creative tasks at different temperatures**

Nonetheless, temperature appears to have an impact on the diversity of the responses by the LLMs. In all of the six tasks that required more than one idea (or story), temperature had a significant effect on either cosine similarity or Levenshtein distance (or both) for both GPT-3.5 and GPT-4 (see Tables S9 and S10). As shown in Fig. 4, lower similarities and higher distances are observed among the ideas within each response as the temperature increases. It seems that temperature should be better interpreted as a parameter for diversity rather than a parameter for creativity.

*Part 2: Collective Creativity*

In Part 1 we looked at the creativity of each LLM response on average. In reality there is no reason why one cannot repeatedly ask for multiple responses from an LLM to the same task. We extend our analysis by looking at the collective creativity of a set of LLM responses. Our approach to LLMs' collective creativity is to examine how many of the top ideas are contributed by LLMs relative to humans when their responses are pooled together. This is particularly valuable in real-world applications, because no matter how many reasonably good ideas are presented for a particular task, only the most creative ones will be implemented. When an equal number of top ideas comes from the LLM responses and humans, we take the number of humans as the indicator of the collective creativity of the LLMs. In other words, these LLMs can replace this number of humans in collectively generating creative ideas.

Specifically, we analyse the 10 most highly rated responses for each task and identify whether humans or LLMs generated these responses. If there is a tie, we include all responses with the same rating, hence the number of top responses can be higher than 10. Social Problem Solving Task 1 and Emoji-prompted Creative Writing Task 1 are not included here as they measure flexibility and diversity, respectively, which focus on the disparity within each response rather than creativity in the ideas.

*Top 10 responses among all responses*

We first pool together all responses by the LLMs and human participants, and examine the top 10 responses for each task. As shown in Fig. 5 (and Table S11), the top responses for the divergent thinking tasks mostly came from human participants. In the domain of problem solving, the five LLMs jointly contributed around 33% - 64% of the top responses in the social problem solving tasks and around 40% - 57% of the top responses in the scientific problem solving tasks. While 79% of the most creative keyword-prompted stories were written by LLMs (primarily GPT-4), stories generated by humans dominated the other creative writing tasks. Averaged across all tasks, the LLMs contributed approximately a third (33%) of the top 10 responses.

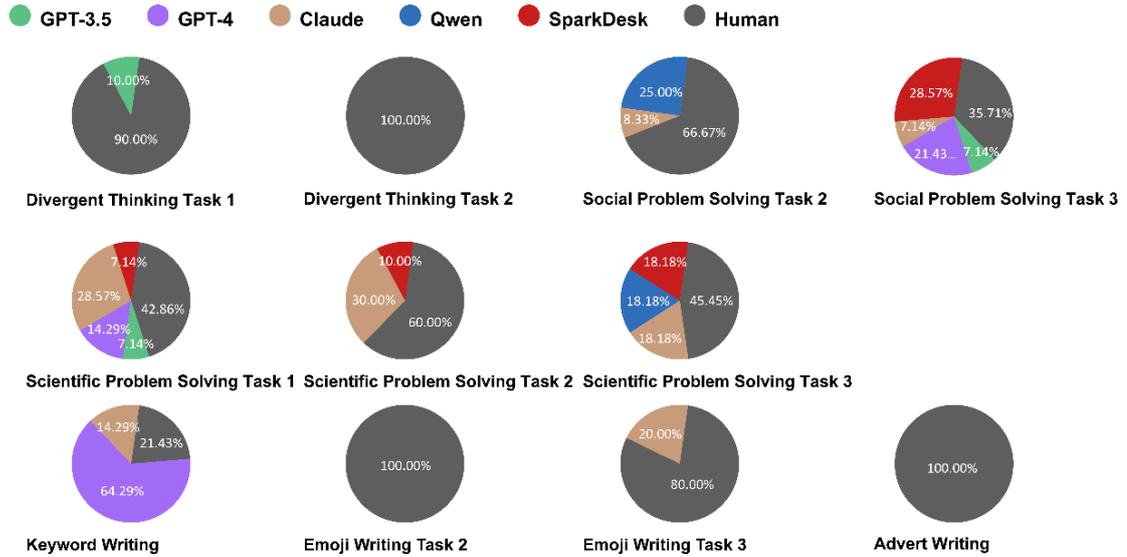

**Fig. 5. Proportion of the top 10 responses in each creative task generated by humans and LLMs.**

*When one LLM is asked 10 times*

The above results indicate that humans are still collectively more likely to come up with the best responses on creative tasks. However, our sample included 467 humans whereas most creative brainstorming sessions in the real world include fewer than 10[45]. We therefore analyse our results further to see what size human group would be necessary to equal one LLM when it is asked 10 times. First, we split the responses of each LLM into five groups of 10 responses. Then we randomly select N human participants (N is an integer starting from 1) and combine their responses with one group of 10 LLM responses. To ensure robustness we adopt a bootstrapping approach, where we repeat the random selection 1000 times and take the average result. If the LLM and the humans contribute the same number of responses in the top 10 responses (i.e., 50% each), we would take the current N value and average them across the five groups as an indicator of the LLM's collective creativity.

The results for each LLM averaged across all creative tasks are presented in Fig. 6 (see detailed results for each task in Table S12). GPT-4 and Claude appear to be the most creative LLMs for top responses, as their collective creativity is equivalent to a group of 10 humans. GPT-3.5 shows the poorest collective creativity but is still equivalent to 8 humans. Qwen and SparkDesk, both equivalent to 9 humans, lie in the middle.

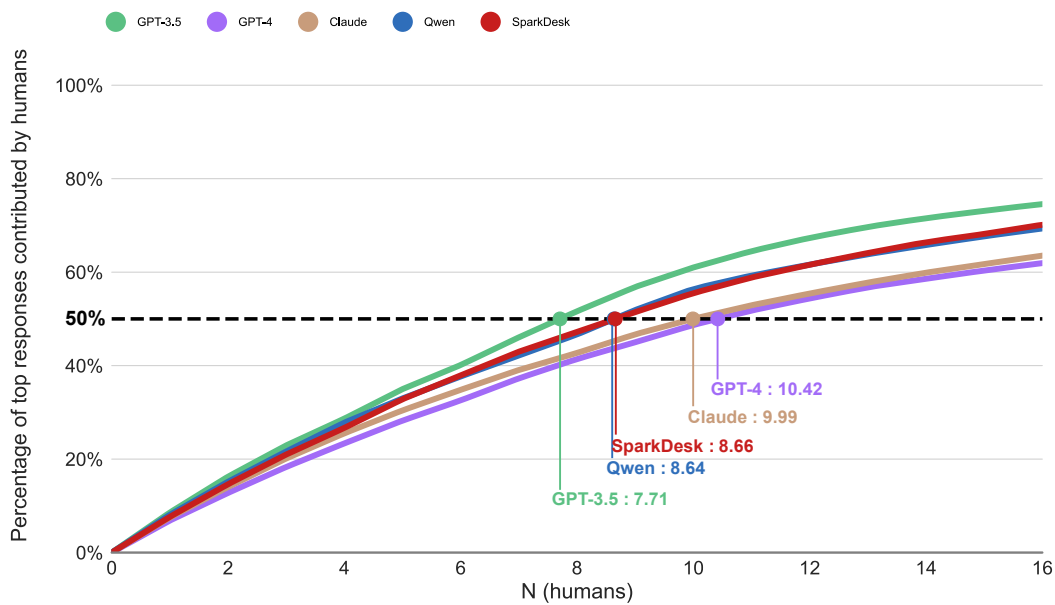

**Fig. 6. Collective creativity of the LLMs averaged across the creative tasks when asked 10 times**

*When more responses are requested from LLMs*

  We continue to probe LLMs' collective creativity when more responses are requested. Specifically, we mix the responses of all LLMs and randomly select a certain number of responses, ranging from 10 responses to 50 responses with an interval of five. Following the method described above, we find out what size human group this number of LLM responses is equivalent to. We repeat this procedure 100 times and take the average of human group sizes as the collective creativity associated with this number of LLM responses (see Table S13 for details). As shown in Fig. 7, a linear relationship is observed. In terms of the incremental improvement, roughly two additional LLM responses (1/0.52) equal one extra human in the group.

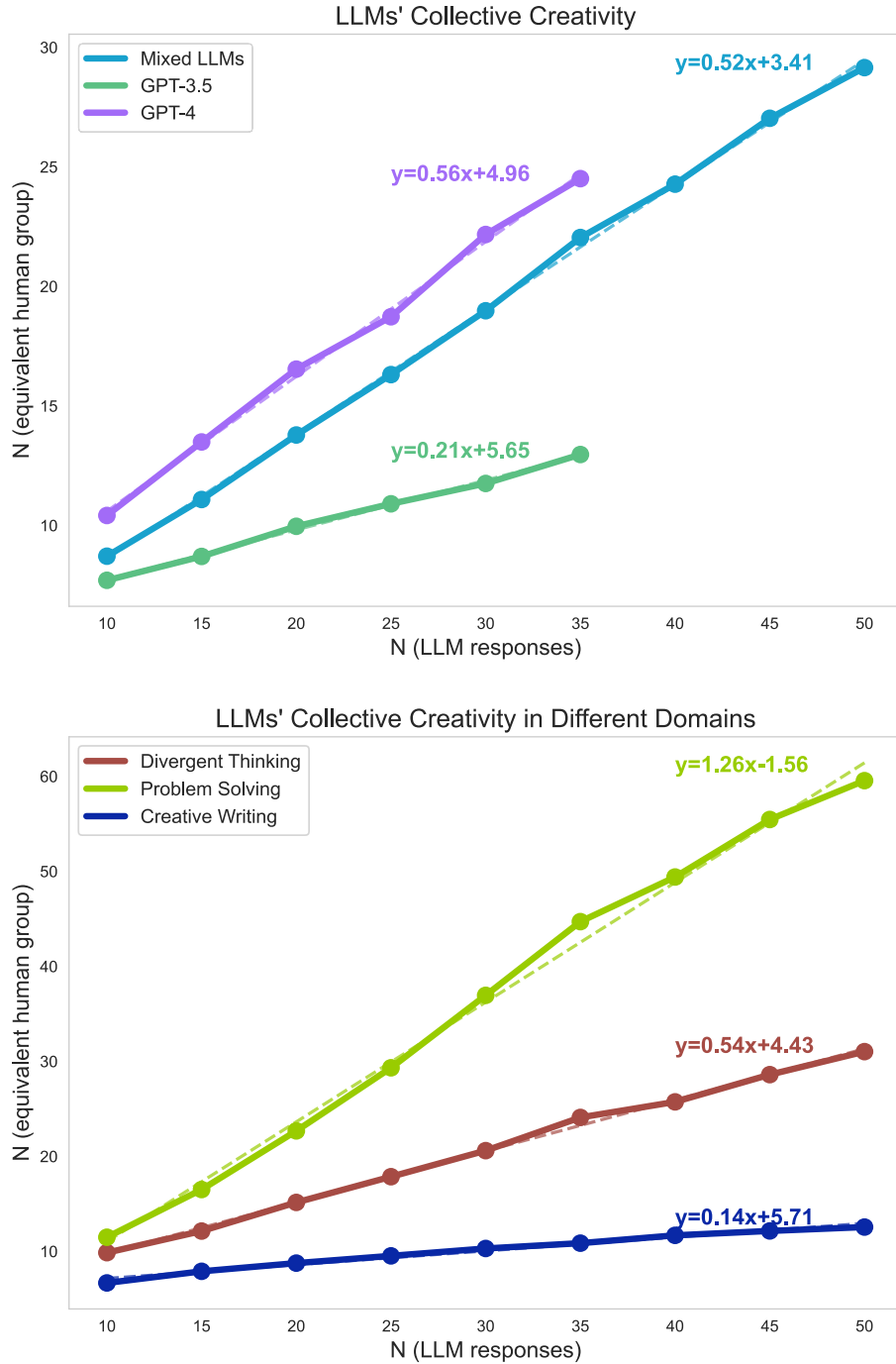

**Fig. 7. Collective creativity of the LLMs when more than 10 responses are requested.**

Consistent with our findings of LLMs' individual creativity, LLMs show excellent collective creativity in the domain of problem solving, as one extra human in the group equals only 0.79 LLM responses (1/1.26). On the other hand, 1.85 LLM responses (1/0.54) are needed to compare to one extra human in the domain of divergent thinking. LLMs are not particularly strong in creative writing, as even 7 more LLMs responses (1/0.14) could not compete with one extra human in the group.

Due to the limited number of responses available for each LLM, we are only able to examine the collective creativity for GPT-3.5 and GPT-4 when more than 10 responses are requested. As shown in Fig. 7, on average, 1.79 more responses of GPT-4 (1/0.56) equal one extra human in the group, whereas up to 4.76 more responses of GPT-3.5 (1/0.21) are needed to compete with one extra human.

**Discussion**

The Organisation for Economic Co-operation and Development suggests that creativity is one of the key 21$^{st}$ Century skills that young people must possess to become effective workers in the knowledge society[46]. With LLMs available to the public, everyone can harness the power of LLMs to generate creative ideas. To understand LLMs' creativity, this study took a multi-facet approach and measured LLMs' individual and collective creativity in multiple dimensions across different domains. Depending on the task and the model, differential performances were reported. On average, the LLMs ranked in the 46$^{th}$ percentile against humans, exhibiting relatively strong performance on divergent thinking and problem solving tasks but suffering from weaknesses in creative writing. Claude and GPT-4 ranked the highest in the 52$^{nd}$ percentile and GPT-3.5 ranked the lowest in the 38$^{th}$ percentile. In most tasks, at least one LLM ranked above the 50$^{th}$ percentile against humans, suggesting that LLMs can be as creative as humans if not more so. Consistent with previous literature, our results also suggested that the responses generated by LLMs lack diversity in comparison to humans.

Considering that it is common to request multiple responses from LLMs while only the best ideas are put into practice, we further examined the collective creativity of the LLMs. When one LLM was asked 10 times, our findings suggest that one LLM could be considered equivalent on average to a group of 8 to 10 humans. When more responses were requested, in terms of incremental improvement, roughly two additional LLM responses equalled one extra human. Taken together, it is reasonable to expect an LLM to generate ideas as creative as a small group of humans when LLM responses are repeatedly collected. This suggests that LLMs, when optimally applied, may greatly contribute to the collective creativity of many organisations, where a typical brainstorming session would not involve more than 10 employees. This is especially beneficial, considering the cost effectiveness of LLMs. However, in a creative endeavour pooling the efforts of thousands of humans, such as the pursuit of scientific knowledge, it is still unlikely that the ideas generated by LLMs would be considered the best[47]. Indeed, we found that when pooling all LLM responses (up to 50 responses from each of the five LLMs) and all human responses (ranging from 200 to 467 responses per task), around two thirds of the top responses were contributed by humans.

Importantly, LLMs are very efficient at generating ideas and can serve as effective assistants, but currently it is still down to humans to evaluate the novelty and usefulness of the ideas generated and select the most creative ones. While initial attempts have reported evidence on the productivity effects of humans working with LLMs[42,48,49], further research is warranted to explore how much incremental creativity the LLMs would bring to AI-human teaming and more importantly what the best strategy for AI-human collaboration is.

This study did not involve extensive prompt engineering, which could potentially improve the performance of LLMs[22,50,51]. We intentionally kept the prompts identical to the instructions human participants received, as we were simulating a real-world scenario where a non-AI expert interacts with an LLM, expecting to receive high-quality responses without spending a long time fine-tuning the prompt. Moreover, it is anticipated that future LLMs will become better at understanding users' intentions, reducing the need for prompt engineering[52,53].

Another limitation of our study is the operationalisation of the measurement of LLMs' collective creativity. In our procedure of human data collection, participants completed the tasks independently, which represents a nominal group setting. For the sake of comparability, we adopted a similar nominal group setting for the LLMs and did not simulate an interactive group setting. While there is abundant evidence that nominal groups outperform interactive groups in both the quantity and the quality of the ideas generated in brainstorming sessions[54-56], some literature on organisational creativity suggests that collective creativity is nurtured by interactions between the group members[39]. Future research should explore the collective creativity of LLMs within an interactive group setting (e.g., a multi-agent system) in comparison to human collaborative efforts.

Although our human sample is relatively large and diverse in comparison to those reported in previous studies that compare LLMs with humans (mostly students or samples collected from Prolific), it is not representative of the human population. In order to find out how generalisable our results are, we tested the demographic differences in humans' performance. The results (see details in Supplementary Text) suggest that in general there is very little demographic difference in humans' creativity performance. This concurs with the fact that most creativity tests, such as the Torrance Tests of Creative Thinking, do not provide adult norms separated according to demographics. Therefore, we are cautiously confident that our findings based on a non-representative but large and diverse human sample can be generalised to a wider population.

LLMs are being regularly updated and, as revealed here, GPT-4 outperformed GPT-3.5 on most tasks. So we expect that future LLMs (including those published after our data collection) would exhibit higher creativity performance and our study will not be the last word on the creative performance of LLMs. However, this does not diminish the contribution of our study to the ongoing discussion about the impact of AI on the future of work, as it achieves an important milestone towards a better understanding of current LLMs' creativity across a range of domains and tasks, offering practical guidance on the application of current LLMs in the workplace.

With public accessibility of LLMs, we foresee radical changes to the future of work. Our findings of LLMs' performance on creative tasks provide evidence for their high-level cognitive capabilities, challenging the notion that only occupations intense in routine tasks would be exposed to AI automation. If LLMs and other AI systems empowered by continued advancement in AI technology deliver on their promised capabilities, they could potentially automate at least parts of the jobs of hundreds of millions of human workers[57]. It becomes ultimately important for us to understand what the role is for humans in the future of work.

**Materials and Methods**

Experimental Design

This study aims to holistically measure LLMs' individual and collective creativity. We use 13 creative tasks spanning over three distinctive domains, including divergent thinking, problem solving, and creative writing. All the tasks and the corresponding instructions are available from https://doi.org/10.6084/m9.figshare.24878421.

- Divergent Thinking Task 1/2 (time limit: 5 minutes)

Participants are presented with a hypothetical scenario, where they are trapped on a desert island alone after encountering a storm on a sailing trip. To keep them warm (Task 1) or pick fruit from a tall tree (Task 2), they are asked to give as many different ideas as possible. Additional instruction is given to encourage original, unique ideas.

Each response is split into ideas, which are rated separately on two dimensions: novelty and usefulness. Novelty is defined as the extent to which the idea is uncommon and unique, whereas usefulness reflects how practical and effective the idea is at addressing the current problem.

- Social Problem Solving Tasks 1, 2, and 3 (time limit: 7 minutes each)

In Task 1, participants are asked to describe three different ideas of what people can do to save water. In Task 2, participants are asked to think of an original idea to promote a smartphone application that is designed to reward users for actions they take to save water. In Task 3, participants are asked to think of an improvement to the application that keeps people using it for longer time. Additional instruction is given to encourage original, unique ideas.

Task 1 measures the dimension of flexibility, which examines if the ideas are sufficiently different from each other. It is operationalised as the number of categories the three ideas in each response fall into. Tasks 2 and 3 measure the dimension of originality, which is defined as the extent to which the response is uncommon and unique.

- Scientific Problem Solving Tasks 1, 2, and 3 (time limit: 7 minutes each)

In Task 1, participants are asked to imagine a 'bicycle of the future' and think of three original improvements to an ordinary bicycle. In Task 2, participants are presented with an idea, which proposes that a camera with facial recognition software could be installed to prevent bicycle theft, and asked to suggest an improvement to make it more effective. In Task 3, participants are asked to suggest an original way to reuse or repurpose a bicycle pedal. Additional instruction is given to encourage original, unique ideas.

The responses in all three tasks are rated on the dimension of originality, which is defined as the extent to which the response is uncommon and unique. In Task 1, the three ideas in each response are rated separately and the average rating is used to indicate the originality of the response.

- Creative Advert Writing Task (time limit: 5 minutes)

Participants are asked to come up with an advert (no more than 30 words) for a hypothetical product: a robot that can provide companionship like a nanny to the elderly.

Responses are rated on their overall creativity that takes into account both novelty (i.e., uniqueness and originality) and usefulness (as an informative advert). Additional ratings are provided on three sub-scales, i.e., conception, expression, and emotion. The average rating across the three sub-scales shows a high correlation with the overall creativity rating (r = 0.886, $P < 0.001$), hence the overall creativity rating is used in the analysis.

- Keyword-prompted Creative Writing Task (time limit: 10 minutes + 10 minutes)

Participants are asked to create an imaginative and creative story (no less than 100 characters) using three keywords (in any order) with no restriction on the genre, theme or style. It is highlighted that the story should be unique, original, distinctive and unexpected. Then participants are asked to create another story (no less than 100 characters) that is different from the previous one using the same set of keywords.

The stories are rated on their overall creativity as well as five sub-scales, including originality, imagination, surprisingness, the interpretation of the keywords and the connection of the keywords. A unidimensional factorial model is constructed with the five sub-scales, which shows good model fit (CFI = 0.937) and generates a factor that is highly correlated with the overall creativity rating (r = 0.948, $P < 0.001$). Hence, the overall rating is used in the analysis. Additionally, the two stories created by each participant are rated on how different they are from each other.

- Emoji-prompted Creative Writing Tasks 1, 2, and 3 (time limit: 7 minutes each)

In Task 1, participants are asked to create two different stories (no more than 120 characters each) that connect two emojis. In Task 2, participants are asked to write one creative story (no more than 120 characters) that connects six emojis in the order they appear. It is highlighted that the story should be original, demonstrates a rich imagination and is well structured. In Task 3, participants are presented with a story that is based on six emojis and asked to write a continuation (no more than 120 characters) based on three more emojis.

Similar to the Keyword-prompted Creative Writing Task, the two stories created by each participant in Task 1 are rated on how different they are from each other, whereas the stories in Tasks 2 and 3 are rated on their overall creativity as well as five sub-scales, including originality, imagination, surprisingness, the interpretation of the emojis and the connection of the emojis. Unidimensional factorial models are constructed with the five sub-scales for Tasks 2 and 3, respectively, both of which show good model fit (CFIs = 0.948 and 0.974) and generate factors that are highly correlated with the overall creativity ratings (rs = 0.949 and 0.929, $P$s $< 0.001$). Hence, the overall ratings are used in the analysis.

*Human data collection*

The tasks above were grouped into two test forms. Form A consisted of Divergent Thinking Task 2, Scientific Problem Solving Tasks 1, 2, and 3, Emoji-prompted Creative Writing Tasks 1, 2, and 3, and Creative Advert Writing Task. Form B consisted of Divergent Thinking Task 1, Social Problem Solving Tasks 1, 2, and 3, Keyword-prompted Creative Writing, and Creative Advert Writing Task.

The test, embedded in the admission assessment for a Master's programme, was administered as a computer-based exam to 467 human participants from China (330 female, 137 male) simultaneously at a designated testing centre. The programme was in a social science subject which has no formal element of creativity training. The average age was 27.56 years (ranging from 21 to 53, SD = 6.03). When applying for the programme, 103 were current students and 140 were current employees with certain years of work experience. All participants had a Bachelor's degree, and 29 of them had a Master's degree. For their undergraduate studies, 69 studied in arts and humanities subjects, 132 in STEM subjects, 103 in the programme subject, and 163 in other social science subjects. Detailed breakdown can be found in Fig. 8.

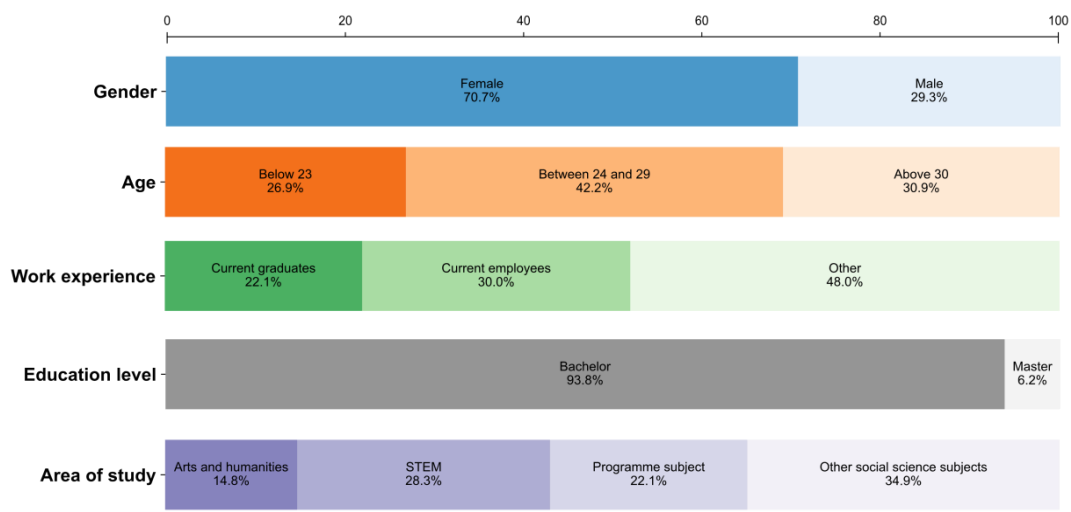

**Fig. 8. Demographic information of the human participants.**

The participants were randomly assigned to Test A (n = 237) or B (n = 230) and completed the tasks in the order specified above. No difference was found between the two groups in terms of age, gender, education level, work experience, and areas of study, as well as performances on the Creative Advert Writing Task and an additional Remote Association Task (see Tables S14 and S15) which was administered in the Master's admission assessment. All tasks were timed separately. After submitting the responses and moving onto subsequent tasks, they were not allowed to review or revise the responses to previous tasks. At the beginning of the test, they were informed that this test aimed to measure creativity. We received ethical approval from the Institutional Review Board of the Faculty of Psychology, Beijing Normal University (Ref: BNU202212090125).

*LLM data collection*

All 13 creative tasks were administered to five different LLMs, namely GPT-3.5 (GPT-3.5-turbo-0301), GPT-4 (GPT-4-32K-0315-preview), Claude, Qwen and SparkDesk (V1.5).

Responses were collected from GPT-3.5 using the public OpenAI API in April, 2023 and again in May, 2023. After rating the responses of humans and GPT-3.5 in the first administration, we realised that the comparison might not be fair, as the humans were fully aware of the purpose of the assessment while in certain tasks the LLMs were not explicitly instructed to be creative. Therefore, we applied additional instructions to encourage the LLMs to generate original and unique responses in the second administration, where these additional instructions were introduced as a system prompt in all divergent thinking and problem solving tasks and attached to the end of the Creative Advert Writing Task. We compared the responses of GPT-3.5 with and without these additional instructions, and the results are presented in Table S16, which shows no significant differences in all tasks except the Divergent Thinking Task 1 (Mean). All results presented in Results are based on the responses with the additional instructions.

We chose five temperature settings, i.e., 0, 0.25, 0.5, 0.75 and 1. When the responses were collected from GPT3.5 and GPT-4 in this study, the temperature ranged from 0 to 1. At each temperature collected 50 responses in separate sessions.

Responses from GPT-4 were collected in May, 2023 using the internal OpenAI API that was made available to Microsoft Research Asia. Additional instructions that encouraged original and unique responses were included as a system prompt in the divergent thinking and problem solving tasks and attached to the end of the Creative Advert Writing Task. Similar to GPT-3.5, five temperatures (0, 0.25, 0.5, 0.75 and 1) were tested and each task was administered 30 times in separate sessions at each temperature.

Responses were collected from Claude, Qwen and SparkDesk in June, 2023 using their web portals, respectively, under the default setting. Each task was administered 50 times in separate conversations. The additional instructions that were applied to GPT-3.5 and GPT-4 were attached to the beginning of each conversation before a creative task was presented.

*Rating procedure*

To ensure comparable number of responses across the LLMs and between the LLMs and the humans, we randomly picked 10 responses at each temperature in GPT-3.5 and in GPT-4 for divergent thinking tasks and creative writing tasks. For the Emoji-prompted Writing Tasks, due to the low rate of valid stories, we examined all responses collected from GPT-3.5 (50 rounds per temperature) and GPT-4 (30 rounds per temperature). If the number of valid stories exceeded 10 for a given temperature, we randomly chose 10 for the rating and subsequent analysis. If not, we kept all stories for the rating and subsequent analysis. It is also noted that in the Creative Advert Writing Task all responses at temperature 0 were identical. For problem solving tasks, we randomly picked 1 response at temperature 0 (as we noticed that responses at temperature 0 were considerably similar to each other), 10 responses at temperature 0.25, 0.5 and 0.75, and 20 responses at temperature 1 (as we would like to balance the number of responses between the GPT models and other LLMs), resulting in 51 responses in total.

Two rounds of rating were carried out. Five human raters (all psychology students) were recruited in each round, who, following the Consensual Assessment Technique[37], independently rated the responses based on the provided guidebook (which was developed in a procedure similar to the PISA creative thinking assessment; https://www.oecd.org/pisa/publications/PISA-2021-creative-thinking-framework.pdf). Specifically, in Round 1, the raters rated the human responses and the initial responses of GPT-3.5 without the additional instructions. In Round 2, the raters rated all responses of GPT-3.5 (both administrations with and without the additional instructions), GPT-4, Claude, Qwen and SparkDesk.

When rating the responses, the raters were provided with a list of responses without being informed who generated these responses. They were instructed to first screen out invalid responses (i.e., not meeting the requirements of the tasks or not relevant to the tasks; the number and percentage of invalid responses are shown in Table S1) and only provide ratings to valid responses on the specified dimensions using a scale from 1 to 5, except for the dimension of flexibility in the Social Problem Solving Task 1, which used a scale from 0 to 3. The inter-rater reliability was indicated by the Intraclass Correlation Coefficients (ICCs) or Fleiss' Kappa, all of which reached an acceptable level of .7 (see Table S1). The average rating across the five raters was taken as the final rating. The mean and SD of the ratings for all dimensions across 13 tasks are shown in Supplementary Table 2.

To align the ratings from different rounds onto the same scale, we conducted z-score based linear equating using the responses that were rated both in Round 1 and Round 2 (i.e., responses of GPT-3.5 without the additional instructions). Truncation was carried out to ensure the derived ratings would not go beyond the rating scale.

*Statistical Analysis*

The analyses were mostly performed in Python. Specifically, we used the statsmodels package (https://www.statsmodels.org/stable/index.html) to calculate Fleiss' Kappa and the *Pingouin* package (https://pingouin-stats.org/) to calculate intraclass correlation coefficients and perform one-way analysis of variance, which compared the performance of GPT-3.5 and GPT-4 under different temperatures. We used the *SciPy* package (https://www.scipy.org/) to run independent sample t-tests (two-sided), which examined the differences between the humans and the LLMs in their performance of the creative tasks and the differences between the two versions of prompts for GPT-3.5, and paired sample t-tests (two-sided), which examined the differences in the novelty and usefulness between the two halves of each response in the divergent thinking tasks. Independent sample t-tests (two-sided) and one-way analysis of variance were carried out to test the demographic differences in the humans' performance. Independent sample t-tests (two-sided) and chi-square tests, also using the SciPy package, were performed to examine potential differences between the human participants assigned to Test A and Test B. Correlations between the novelty and usefulness ratings were calculated with the SciPy package as well. We used the

gaussian_kde function from the SciPy package to estimate the kernel density which was depicted in Fig. 1.

To understand the relationship between the sub-scale ratings and the overall creativity ratings in the Keyword-prompted Writing Task and the Emoji-prompted Writing Tasks, we carried out a series of confirmatory factor analysis using the software Mplus (version 7.0).

To evaluate the text similarity between the ideas (or stories) within each response and between the responses, we used OpenAI's text-embedding-ada-002 (https://platform.openai.com/docs/guides/embeddings/) to transform the text into vector representations, based on which the cosine similarity was calculated with the Scikit-learn package. Additionally, we used the same package to calculate the Levenshtein distance, which is a string metric for measuring the difference between two sequences of text. A high distance indicates small similarity.

We ranked the human participants and benchmarked the LLMs against humans by locating the average performance of the LLMs in the humans' distributions. Focusing on the top creative performance, we examined the top 10 responses in each task after pooling together all responses we collected from the humans and the LLMs, and identified whether humans or LLMs contributed more responses.

To understand the LLMs' collective creativity from a practical perspective, we proposed a metric which is measured by the number of humans that contribute the same proportion of top responses as an LLM when it is asked 10 times. We developed an iterative sampling procedure using the pandas package (https://pandas.pydata.org/). First, we randomly split the responses of a certain LLM into groups of ten. Then we combined each group of responses with those from randomly drawn N human participants. By varying the number of N (each N was sampled 1000 times), we were able to identify the size of a group of humans, where half of the top 10 responses in the combined pool of responses came from the humans (and half came from the LLM), and this group size would indicate the collective creativity of the LLM. Going further with more than 10 LLM responses, we adopted a similar approach. For GPT-3.5 and GPT-4, we gathered five groups of LLM responses. For the mixed LLM analysis, we randomly selected a certain number of responses 100 times. Then we followed a similar procedure as described above to quantitatively assess LLMs' collective creativity associated with a certain number of responses. A linear regression was carried out using the SciPy package in the end to find out the number of LLM responses required additionally to compete with one extra human in the group.

**Acknowledgments**

**Funding:**
National Natural Science Foundation of China grant 62377003 (FL)
Invesco's philanthropic donation to Cambridge Judge Business School (LS)

**Author contributions:**
Conceptualization: LS, YYuan, FL, DS
Methodology: LS, YYuan, FL, DS
Investigation: LS, YYuan, XX, XW, FL
Formal analysis: LS, YYuan, YYao, YL, HZ, DS
Visualization: LS, YYuan, YYao, DS
Resources: XX, FL
Supervision: FL, DS
Writing-original draft: LS, YYuan
Writing-review & editing: LS, YYuan, FL, DS

**Competing interests:** At the time of this writing XX and XW were employees of Microsoft Research Asia and may own Microsoft stock as part of the standard compensation package. All other authors declare that they have no competing interests.

**Data and materials availability:** The datasets analysed during the current study are available from https://doi.org/10.6084/m9.figshare.24878421.

# Supplementary Materials for

## Large Language Models show both individual and collective creativity comparable to humans

Luning Sun *et al.*

*Corresponding author. Email: luof@bnu.edu.cn and d.stillwell@jbs.cam.ac.uk

**This PDF file includes:**

    Supplementary Text
    Tables S1 to S22
    Figs. S1 to S8

**Supplementary Text**

In order to find out how generalisable our results are, we tested the demographic differences in humans' performance. As shown in Table S17, on two out of the 16 creativity indicators (i.e., Scientific Problem Solving Task 1 and Emoji-prompted Writing Task 3), significant gender differences are detected, where males show higher performance than females. In terms of age, we split the sample into three groups, i.e., below the age of 23 (the age when students are expected to complete their undergraduate studies), between 24 and 29, and above 30. Age has a significant effect on three indicators (i.e., Scientific Problem Solving Task 3, Social Problem Solving Task 1, and Emoji-prompted Writing Task 2; See Table S18). Post-hoc Bonferroni tests (see Table S19) suggest that the group between 24 and 29 shows greater performance than the group below 23 in Scientific Problem Solving Task 3, and all other comparisons are not significant.

We also test the performance differences in terms of work experience and educational background. As shown in Tables S20 and S21, current employees, when applying for the Master's programme, show better performance in Scientific Problem Solving Task 3 and worse performance in Emoji-prompted Writing Task 2 than current students; those who have received a Master's degree perform poorer than those who have not received a Master's degree in Social Problem Solving Task 1. To examine if the tasks would benefit from training in certain areas of studies, we split the sample into four groups, including STEM subjects, arts and humanities subjects, the programme subject, and other social science subjects. One-way ANOVA shows non-significant results on all creativity indicators (see Table S22).

**Table S1. Statistics of the creative task responses from humans and LLMs and the inter-rater reliabilities (and the corresponding 95% confidence intervals) in the ratings.**

| Tasks | Statistics | Human | GPT-3.5 Prompt v1 | GPT-3.5 prompt v2 | GPT-4 | Claude | Qwen | SparkDesk |
|---|---|---|---|---|---|---|---|---|
| **Divergent Thinking** | | | | | | | | |
| Task 1 | Number of responses | 230 | 50 | 50 | 50 | 50 | 50 | 50 |
| | Number of ideas | 900 | 500 | 510 | 792 | 577 | 472 | 306 |
| | Number of valid ideas | 856 | 456 | 445 | 748 | 542 | 427 | 298 |
| | ICC Round 1 (Novelty) | 0.79 [0.76, 0.81] | | | | | | |
| | ICC Round 2 (Novelty) | | 0.91 [0.90, 0.92] | | | 0.89 [0.88, 0.90]* | | |
| | ICC Round 1 (Usefulness) | 0.72 [0.69, 0.75] | | | | | | |
| | ICC Round 2 (Usefulness) | | 0.80 [0.77, 0.82] | | | 0.86 [0.85, 0.87]* | | |
| Task 2 | Number of responses | 236† | 50 | 50 | 50 | 50 | 50 | 50 |
| | Number of ideas | 921 | 500 | 500 | 502 | 399 | 390 | 272 |
| | Number of valid ideas | 857 | 468 | 483 | 499 | 394 | 344 | 245 |
| | ICC Round 1 (Novelty) | 0.90 [0.89, 0.91] | | | | | | |
| | ICC Round 2 (Novelty) | | 0.89 [0.88, 0.90] | | | 0.92 [0.92, 0.93]* | | |
| | ICC Round 1 (Usefulness) | 0.84 [0.82, 0.85] | | | | | | |
| | ICC Round 2 (Usefulness) | | 0.90 [0.89, 0.91] | | | 0.91 [0.90, 0.91]* | | |
| **Scientific Problem Solving** | | | | | | | | |
| | Statistics | Human | GPT-3.5 Prompt v1 | GPT-3.5 prompt v2 | GPT-4 | Claude | Qwen | SparkDesk |
| Task 1 | Number of responses | 237 | 51 | 51 | 51 | 50 | 50 | 50 |
| | Number of ideas | 711 | 153 | 153 | 153 | 150 | 150 | 150 |
| | Number of valid ideas | 696 | 153 | 153 | 153 | 150 | 143 | 143 |
| | ICC Round 1 (Creativity) | 0.87 [0.85, 0.88] | | | | | | |
| | ICC Round 2 (Creativity) | | | | 0.88 [0.87, 0.90] | | | |
| Task 2 | Number of responses | 237 | 51 | 51 | 51 | 50 | 0 | 50 |
| | Number of valid responses | 236 | 51 | 51 | 51 | 50 | 0 | 48 |
| | ICC Round 1 (Creativity) | 0.89 [0.86, 0.91] | | | | | | |
| | ICC Round 2 (Creativity) | | | | 0.90 [0.88, 0.92] | | | |
| Task 3 | Number of responses | 237 | 51 | 51 | 51 | 50 | 50 | 50 |
| | Number of valid responses | 237 | 51 | 51 | 51 | 50 | 40 | 48 |
| | ICC Round 1 (Creativity) | 0.88 [0.86, 0.90] | | | | | | |
| | ICC Round 2 (Creativity) | | | | 0.92 [0.90, 0.93] | | | |
| **Social Problem Solving** | | | | | | | | |
| | Statistics | | GPT-3.5 | GPT-3.5 | | | | |

|  | | Human | Prompt v1 | prompt v2 | GPT-4 | Claude | Qwen | SparkDesk |
|---|---|---|---|---|---|---|---|---|
| Task 1 | Number of responses | 229† | 51 | 51 | 51 | 50 | 50 | 50 |
|  | Number of ideas | 685 | 153 | 153 | 153 | 150 | 150 | 150 |
|  | Number of valid ideas | 684 | 153 | 153 | 153 | 150 | 148 | 149 |
|  | Fleiss' kappa Round 1 (Flexibility) | 0.80 [0.77, 0.82] | | | | | | |
|  | Fleiss' kappa Round 2 (Flexibility) | | | | 0.78 [0.76, 0.81] | | | |
| Task 2 | Number of responses | 230 | 51 | 51 | 51 | 50 | 50 | 50 |
|  | Number of valid responses | 229 | 51 | 51 | 51 | 50 | 50 | 50 |
|  | ICC Round 1 (Creativity) | 0.93 [0.91, 0.94] | | | | | | |
|  | ICC Round 2 (Creativity) | | | | 0.92 [0.91, 0.93] | | | |
| Task 3 | Number of responses | 230 | 51 | 51 | 51 | 50 | 50 | 50 |
|  | Number of valid responses | 226 | 51 | 51 | 51 | 50 | 50 | 50 |
|  | ICC Round 1 (Creativity) | 0.90 [0.89, 0.92] | | | | | | |
|  | ICC Round 2 (Creativity) | | | | 0.91 [0.89, 0.93] | | | |

**Keyword-prompted Writing**

| Statistics | Human | GPT-3.5 Prompt v1 | GPT-3.5 prompt v2 | GPT-4 | Claude | Qwen | SparkDesk |
|---|---|---|---|---|---|---|---|
| Number of responses | 230 | 50 | - | 50 | 50 | 50 | 50 |
| Number of stories | 460 | 100 | - | 100 | 100 | 100 | 100 |
| Number of valid stories | 399 | 100 | - | 100 | 48 | 92 | 62 |
| ICC Round 1 (Creativity) | 0.85 [0.82, 0.87] | | | | | | |
| ICC Round 2 (Creativity) | | | | 0.84 [0.82, 0.87] | | | |
| ICC Round 1 (Diversity) | 0.85 [0.81, 0.88] | | | | | | |
| ICC Round 2 (Diversity) | | | | 0.85 [0.82, 0.88] | | | |

**Emoji-prompted Writing**

|  | Statistics | Human | GPT-3.5 Prompt v1 | GPT-3.5 prompt v2 | GPT-4 | Claude | Qwen | SparkDesk |
|---|---|---|---|---|---|---|---|---|
| Task 1 | Number of responses | 237 | 250‡ | - | 150[c] | 50 | 50 | 50 |
|  | Number of stories | 472 | 500 | - | 300 | 100 | 100 | 100 |
|  | Number of valid stories | 414 | 326 | - | 144 | 96 | 20 | 66 |
|  | ICC Round 1 (Diversity) | 0.81 [0.77, 0.85] | | | | | | |
|  | ICC Round 2 (Diversity) | | | | 0.87 [0.84, 0.90] | | | |
| Task 2 | Number of stories | 237 | 250‡ | - | 150[c] | 50 | 50 | 50 |
|  | Number of valid stories | 181 | 38 | - | 73 | 50 | 12 | 4 |
|  | ICC Round 1 (Creativity) | 0.87 [0.84, 0.90] | | | | | | |
|  | ICC Round 2 (Creativity) | | | | 0.85 [0.81, 0.89] | | | |

| | Statistics | Human | GPT-3.5 Prompt v1 | GPT-3.5 prompt v2 | GPT-4 | Claude | Qwen | SparkDesk |
|---|---|---|---|---|---|---|---|---|
| **Task 3** | Number of stories | 237 | 250‡ | - | 150ᶜ | 50 | 50 | 50 |
| | Number of valid stories | 189 | 86 | - | 109 | 50 | 0 | 0 |
| | ICC Round 1 (Creativity) | 0.81 [0.77, 0.84] | | | | | | |
| | ICC Round 2 (Creativity) | | | | 0.85 [0.81, 0.89] | | | |
| **Creative Advert Writing** | | | | | | | | |
| | Statistics | Human | GPT-3.5 Prompt v1 | GPT-3.5 prompt v2 | GPT-4 | Claude | Qwen | SparkDesk |
| | Number of responses | 467 | 41 | 41 | 41 | 50 | 50 | 50 |
| | Number of valid responses | 418 | 41 | 41 | 41 | 49 | 21 | 46 |
| | ICC Round 1 (Creativity) | 0.79 [0.76, 0.82] | | | | | | |
| | ICC Round 2 (Creativity) | | | | 0.88 [0.85, 0.90] | | | |

*Note.* *The ICCs were calculated based on the ratings of responses of the three LLMs as well as those of GPT-3.5 Prompt V1, which were included for equating purpose. †One participant failed to provide a response to this task within the time limit. ‡We examined all stories collected from GPT-3.5 (50 rounds per temperature) and GPT-4 (30 rounds per temperature). If the number of valid stories exceeded 10 for a given temperature, we randomly chose 10 and included them in the rating and subsequent analysis. If not, we kept all stories for rating and subsequent analysis.

**Table S2. Descriptive statistics and independent sample *t*-test* (two-sided) results of novelty and usefulness ratings for divergent thinking tasks between LLMs and human participants.**

| Tasks | Statistics | GPT-3.5 | GPT-4 | Claude | Qwen | SparkDesk | Human |
|---|---|---|---|---|---|---|---|
| **Divergent Thinking Task 1** | | | | | | | |
| | *n* | 445 | 748 | 542 | 427 | 298 | 856 |
| Novelty | *M* | 1.879 | 1.929 | 1.887 | 1.911 | 1.747 | 1.907 |
| | *SD* | 0.517 | 0.567 | 0.522 | 0.582 | 0.504 | 0.650 |
| | 95% CI (Δ) | [-0.093, 0.037] | [-0.038, 0.082] | [-0.082, 0.042] | [-0.067, 0.074] | [-0.232, -0.088] | - |
| | *t* | -0.851 | 0.724 | -0.623 | 0.100 | -4.366 | - |
| | *df* | 1091.906 | 1601.995 | 1320.037 | 939.683 | 664.262 | - |
| | *P* | 0.395 | 0.469 | 0.533 | 0.920 | < 0.001 | - |
| | Cohen's *d* | -0.046 | 0.036 | -0.033 | 0.006 | -0.260 | - |
| Usefulness | *M* | 2.558 | 2.516 | 2.527 | 2.381 | 2.625 | 2.389 |
| | *SD* | 0.515 | 0.525 | 0.515 | 0.559 | 0.515 | 0.559 |
| | 95% CI (Δ) | [0.107, 0.229] | [0.073, 0.179] | [0.080, 0.195] | [-0.074, 0.056] | [0.166, 0.305] | - |
| | *t* | 5.427 | 4.654 | 4.701 | -0.269 | 6.650 | - |
| | *df* | 964.805 | 1593.378 | 1218.542 | 1281 | 557.145 | - |
| | *P* | < 0.001 | < 0.001 | < 0.001 | 0.788 | < 0.001 | - |
| | Cohen's *d* | 0.309 | 0.232 | 0.253 | -0.016 | 0.430 | - |
| **Divergent Thinking Task 2** | | | | | | | |
| | *n* | 483 | 499 | 394 | 344 | 245 | 857 |
| Novelty | *M* | 2.001 | 2.052 | 2.194 | 2.568 | 2.160 | 2.046 |
| | *SD* | 0.652 | 0.769 | 0.713 | 0.880 | 0.835 | 0.816 |
| | 95% CI (Δ) | [-0.125, 0.035] | [-0.082, 0.095] | [0.054, 0.242] | [0.414, 0.631] | [-0.002, 0.231] | - |
| | *t* | -1.108 | 0.146 | 3.098 | 9.490 | 1.927 | - |
| | *df* | 1188.096 | 1354 | 1249 | 592.250 | 1100 | - |
| | *P* | 0.268 | 0.884 | 0.002 | < 0.001 | 0.054 | - |
| | Cohen's *d* | -0.059 | 0.008 | 0.189 | 0.626 | 0.140 | - |
| Usefulness | *M* | 2.560 | 2.657 | 2.523 | 2.161 | 2.415 | 2.521 |
| | *SD* | 0.632 | 0.776 | 0.674 | 0.823 | 0.823 | 0.693 |
| | 95% CI (Δ) | [-0.035, 0.115] | [0.054, 0.219] | [-0.080, 0.084] | [-0.458, -0.260] | [-0.219, 0.008] | - |
| | *t* | 1.042 | 3.254 | 0.053 | -7.141 | -1.824 | - |
| | *df* | 1338 | 948.557 | 1249 | 547.931 | 348.772 | - |
| | *P* | 0.298 | 0.001 | 0.958 | < 0.001 | 0.069 | - |

| | | | | | | |
|---|---|---|---|---|---|---|
| **Cohen's d** | 0.059 | 0.189 | 0.003 | -0.490 | -0.145 | - |

*Note.* *Before running the *t*-test, we tested if the populations had identical variances. If true, a standard independent sample *t*-test that assumes equal population variances was performed. If false, a Welch's *t*-test, which does not assume equal population variance, was performed.

**Table S3. Descriptive statistics and paired sample *t*-test (two-sided) results of novelty and usefulness ratings for the divergent thinking tasks between the first half and second half ideas from each response*.**

| Tasks | Model | First half | | | Second half | | | *t*- test results | | | | |
|---|---|---|---|---|---|---|---|---|---|---|---|---|
| | | *n* | M | SD | *n* | M | SD | 95% CI (Δ) | *t* | df | *P* | Cohen's *d* |
| **Divergent Thinking Task 1** | | | | | | | | | | | | |
| Novelty | GPT-3.5 | 50 | 1.801 | 0.186 | 50 | 1.966 | 0.280 | [0.059, 0.271] | 3.122 | 49 | 0.003 | 0.693 |
| | GPT-4 | 50 | 1.797 | 0.165 | 50 | 2.059 | 0.267 | [0.169, 0.355] | 5.648 | 49 | <0.001 | 1.180 |
| | Claude | 50 | 1.806 | 0.183 | 50 | 1.965 | 0.239 | [0.069, 0.249] | 3.550 | 49 | 0.001 | 0.748 |
| | Qwen | 50 | 1.791 | 0.243 | 50 | 2.006 | 0.285 | [0.112, 0.318] | 4.200 | 49 | <0.001 | 0.811 |
| | SparkDesk | 50 | 1.651 | 0.178 | 50 | 1.846 | 0.392 | [0.053, 0.337] | 2.767 | 49 | 0.008 | 0.641 |
| | Human | 225 | 1.814 | 0.531 | 225 | 2.004 | 0.480 | [0.100, 0.280] | 4.175 | 224 | <0.001 | 0.376 |
| Usefulness | GPT-3.5 | 50 | 2.734 | 0.222 | 50 | 2.398 | 0.271 | [-0.426, -0.246] | -7.531 | 49 | <0.001 | -1.357 |
| | GPT-4 | 50 | 2.669 | 0.148 | 50 | 2.367 | 0.213 | [-0.382, -0.223] | -7.633 | 49 | <0.001 | -1.649 |
| | Claude | 50 | 2.683 | 0.224 | 50 | 2.379 | 0.215 | [-0.381, -0.227] | -7.907 | 49 | <0.001 | -1.386 |
| | Qwen | 50 | 2.474 | 0.266 | 50 | 2.277 | 0.291 | [-0.312, -0.082] | -3.443 | 49 | 0.001 | -0.706 |
| | SparkDesk | 50 | 2.769 | 0.215 | 50 | 2.505 | 0.302 | [-0.378, -0.150] | -4.654 | 49 | <0.001 | -1.008 |
| | Human | 225 | 2.545 | 0.516 | 225 | 2.359 | 0.411 | [-0.263, -0.108] | -4.727 | 224 | <0.001 | -0.397 |
| **Divergent Thinking Task 2** | | | | | | | | | | | | |
| Novelty | GPT-3.5 | 50 | 1.838 | 0.208 | 50 | 2.169 | 0.313 | [0.216, 0.446] | 5.794 | 49 | <0.001 | 1.246 |
| | GPT-4 | 50 | 1.784 | 0.241 | 50 | 2.320 | 0.247 | [0.420, 0.653] | 9.249 | 49 | <0.001 | 2.197 |
| | Claude | 50 | 2.032 | 0.333 | 50 | 2.360 | 0.344 | [0.199, 0.458] | 5.090 | 49 | <0.001 | 0.969 |
| | Qwen | 49 | 2.243 | 0.450 | 49 | 2.814 | 0.580 | [0.374, 0.768] | 5.831 | 48 | <0.001 | 1.099 |
| | SparkDesk | 49 | 1.726 | 0.307 | 49 | 2.537 | 0.586 | [0.631, 0.991] | 9.072 | 48 | <0.001 | 1.735 |
| | Human | 225 | 1.886 | 0.610 | 225 | 2.177 | 0.635 | [0.174, 0.407] | 4.931 | 224 | <0.001 | 0.467 |
| Usefulness | GPT-3.5 | 50 | 2.703 | 0.201 | 50 | 2.412 | 0.294 | [-0.386, -0.196] | -6.178 | 49 | <0.001 | -1.154 |
| | GPT-4 | 50 | 2.971 | 0.240 | 50 | 2.341 | 0.289 | [-0.757, -0.503] | -9.983 | 49 | <0.001 | -2.370 |

| | | | | | | | | | | |
|---|---|---|---|---|---|---|---|---|---|---|
| **Claude** | 50 | 2.694 | 0.309 | 50 | 2.366 | 0.334 | [-0.443, -0.212] | -5.710 | 49 | <0.001 | -1.018 |
| **Qwen** | 49 | 2.484 | 0.472 | 49 | 1.937 | 0.493 | [-0.732, -0.362] | -5.939 | 48 | <0.001 | -1.134 |
| **SparkDesk** | 49 | 2.857 | 0.345 | 49 | 2.031 | 0.553 | [-1.008, -0.644] | -9.136 | 48 | <0.001 | -1.792 |
| **Human** | 225 | 2.681 | 0.540 | 225 | 2.468 | 0.511 | [-0.312, -0.114] | -4.233 | 224 | <0.001 | -0.404 |

*Note.* *If there are an odd number of ideas in one response, we would remove the idea in the middle and compare the ideas before and those after.

**Table S4. Descriptive statistics and independent sample *t*-test[*] (two-sided) results of creative task response ratings between LLMs and human participants.**

| Tasks | Statistics | GPT-3.5 | GPT-4 | Claude | Qwen | SparkDesk | Human |
|---|---|---|---|---|---|---|---|
| **Divergent Thinking** | | | | | | | |
| **Task 1 Creativity Mean** | *n* | 50 | 50 | 50 | 50 | 50 | 230 |
| | *M* | 4.738 | 4.725 | 4.665 | 4.392 | 4.443 | 4.770 |
| | *SD* | 0.495 | 0.309 | 0.336 | 0.377 | 0.316 | 1.320 |
| | 95% CI (Δ) | [-0.251, 0.189] | [-0.236, 0.147] | [-0.300, 0.090] | [-0.579, -0.177] | [-0.520, -0.134] | - |
| | *t* | -0.280 | -0.456 | -1.057 | -3.704 | -3.343 | - |
| | *df* | 210.241 | 276.777 | 272.664 | 261.264 | 276.019 | - |
| | *P* | 0.779 | 0.649 | 0.292 | < 0.001 | 0.001 | - |
| | Cohen's *d* | -0.026 | -0.037 | -0.087 | -0.313 | -0.271 | - |
| **Task 1 Creativity Max** | *n* | 50 | 50 | 50 | 50 | 50 | 230 |
| | *M* | 6.746 | 7.336 | 6.667 | 6.407 | 5.887 | 6.494 |
| | *SD* | 1.058 | 1.082 | 0.818 | 0.948 | 0.575 | 1.913 |
| | 95% CI (Δ) | [-0.135, 0.640] | [0.450, 1.235] | [-0.165, 0.511] | [-0.450, 0.277] | [-0.903, -0.312] | - |
| | *t* | 1.291 | 4.248 | 1.012 | -0.471 | -4.045 | - |
| | *df* | 129.325 | 125.693 | 180.114 | 149.085 | 253.791 | - |
| | *P* | 0.199 | < 0.001 | 0.313 | 0.638 | < 0.001 | - |
| | Cohen's *d* | 0.141 | 0.470 | 0.098 | -0.049 | -0.346 | - |
| **Task 2 Creativity Mean** | *n* | 50 | 50 | 50 | 50 | 50 | 235 |
| | *M* | 4.819 | 4.991 | 5.197 | 4.968 | 4.620 | 5.044 |
| | *SD* | 0.380 | 0.352 | 0.488 | 0.412 | 0.316 | 1.249 |
| | 95% CI (Δ) | [-0.417, -0.033] | [-0.240, 0.135] | [-0.058, 0.363] | [-0.273, 0.121] | [-0.607, -0.241] | - |
| | *t* | -2.305 | -0.550 | 1.433 | -0.759 | -4.566 | - |
| | *df* | 252.865 | 265.061 | 199.689 | 237.873 | 276.426 | - |
| | *P* | 0.022 | 0.583 | 0.153 | 0.449 | < 0.001 | - |
| | Cohen's *d* | -0.196 | -0.046 | 0.133 | -0.066 | -0.371 | - |
| **Task 2 Creativity Max** | *n* | 50 | 50 | 50 | 50 | 50 | 235 |
| | *M* | 6.262 | 7.667 | 6.850 | 6.531 | 5.806 | 6.682 |
| | *SD* | 0.903 | 1.305 | 1.184 | 0.771 | 0.918 | 1.996 |
| | 95% CI (Δ) | [-0.781, -0.060] | [0.404, 1.565] | [-0.252, 0.588] | [-0.486, 0.183] | [-1.240, -0.513] | - |
| | *t* | -2.305 | 3.337 | 0.792 | -0.892 | -4.766 | - |
| | *df* | 166.168 | 283 | 117.173 | 202.408 | 162.639 | - |
| | *P* | 0.022 | 0.001 | 0.430 | 0.374 | < 0.001 | - |

|  | | GPT-3.5 | GPT-4 | Claude | Qwen | SparkDesk | Human |
|---|---|---|---|---|---|---|---|
| | Cohen's d | -0.227 | 0.52 | 0.089 | -0.082 | -0.473 | - |

**Scientific Problem Solving**

|  | Statistics | GPT-3.5 | GPT-4 | Claude | Qwen | SparkDesk | Human |
|---|---|---|---|---|---|---|---|
| **Task 1** | n | 51 | 51 | 50 | 48 | 48 | 237 |
| **Creativity** | M | 1.834 | 2.497 | 2.347 | 1.964 | 2.187 | 1.948 |
| | SD | 0.364 | 0.301 | 0.411 | 0.39 | 0.376 | 0.374 |
| | 95% CI (Δ) | [-0.226, 0.000] | [0.440, 0.660] | [0.283, 0.516] | [-0.101, 0.134] | [0.123, 0.356] | - |
| | t | -1.975 | 9.836 | 6.749 | 0.281 | 4.044 | - |
| | df | 286 | 286 | 285 | 283 | 283 | - |
| | P | 0.049 | < 0.001 | < 0.001 | 0.779 | < 0.001 | - |
| | Cohen's d | -0.305 | 1.518 | 1.05 | 0.044 | 0.64 | - |
| **Task 2** | n | 51 | 51 | 50 | 0 | 48 | 236 |
| **Creativity** | M | 2.286 | 2.419 | 2.807 | - | 2.561 | 2.359 |
| | SD | 0.62 | 0.375 | 0.795 | - | 0.823 | 0.785 |
| | 95% CI (Δ) | [-0.304, 0.157] | [-0.085, 0.205] | [0.207, 0.689] | - | [-0.045, 0.449] | - |
| | t | -0.626 | 0.817 | 3.655 | - | 1.611 | - |
| | df | 285 | 159.181 | 284 | - | 282 | - |
| | P | 0.532 | 0.415 | <0.001 | - | 0.108 | - |
| | Cohen's d | -0.097 | 0.082 | 0.569 | - | 0.255 | - |
| **Task 3** | n | 51 | 51 | 50 | 40 | 48 | 237 |
| **Creativity** | M | 1.581 | 1.723 | 2.682 | 2.523 | 2.484 | 2.642 |
| | SD | 0.419 | 0.482 | 0.858 | 0.897 | 0.713 | 0.842 |
| | 95% CI (Δ) | [-1.219, -0.902] | [-1.091, -0.747] | [-0.219, 0.299] | [-0.406, 0.167] | [-0.390, 0.074] | - |
| | t | -13.221 | -10.579 | 0.303 | -0.822 | -1.357 | - |
| | df | 150.595 | 125.943 | 285 | 275 | 76.145 | - |
| | P | < 0.001 | < 0.001 | 0.762 | 0.412 | 0.179 | - |
| | Cohen's d | -1.351 | -1.161 | 0.047 | -0.14 | -0.192 | - |

**Social Problem Solving**

|  | Statistics | GPT-3.5 | GPT-4 | Claude | Qwen | SparkDesk | Human |
|---|---|---|---|---|---|---|---|
| **Task 1** | n | 51 | 51 | 50 | 50 | 50 | 229 |
| **Flexibility** | M | 2.157 | 2.843 | 2.1 | 2.04 | 2.36 | 2.306 |
| | SD | 0.612 | 0.367 | 0.678 | 0.727 | 0.563 | 0.58 |
| | 95% CI (Δ) | [-0.327, 0.030] | [0.410, 0.665] | [-0.389, -0.022] | [-0.453, -0.079] | [-0.123, 0.232] | - |
| | t | -1.641 | 8.381 | -2.203 | -2.798 | 0.603 | - |
| | df | 278 | 113.21 | 277 | 277 | 277 | - |
| | P | 0.102 | < 0.001 | 0.028 | 0.006 | 0.547 | - |

|  | Statistics | GPT-3.5 | GPT-4 | Claude | Qwen | SparkDesk | Human |
|---|---|---|---|---|---|---|---|
| | Cohen's d | -0.254 | 0.982 | -0.344 | -0.437 | 0.094 | - |
| **Task 2 Creativity** | $n$ | 51 | 51 | 50 | 50 | 50 | 229 |
| | $M$ | 2.857 | 2.492 | 2.228 | 2.343 | 1.841 | 2.117 |
| | $SD$ | 0.543 | 0.522 | 0.73 | 0.754 | 0.442 | 0.775 |
| | 95% CI (Δ) | [0.558, 0.921] | [0.198, 0.552] | [-0.125, 0.346] | [-0.011, 0.463] | [-0.436, 0.116] | - |
| | $t$ | 8.066 | 4.207 | 0.923 | 1.875 | -3.414 | - |
| | $df$ | 101.034 | 105.529 | 277 | 277 | 124.779 | - |
| | $P$ | < 0.001 | < 0.001 | 0.357 | 0.062 | 0.001 | - |
| | Cohen's d | 1.001 | 0.51 | 0.144 | 0.293 | -0.379 | - |
| **Task 3 Creativity** | $n$ | 51 | 51 | 50 | 50 | 50 | 226 |
| | $M$ | 2.445 | 3.045 | 2.355 | 2.36 | 3.277 | 2.692 |
| | $SD$ | 0.641 | 0.874 | 0.713 | 0.694 | 0.701 | 0.833 |
| | 95% CI (Δ) | [-0.457, 0.038] | [0.097, 0.610] | [-0.566, 0.108] | [-0.556, 0.108] | [0.335, 0.834] | - |
| | $t$ | -2.346 | 2.71 | -2.928 | -2.944 | 4.615 | - |
| | $df$ | 92.451 | 275 | 81.426 | 83.37 | 274 | - |
| | $P$ | 0.021 | 0.007 | 0.004 | 0.004 | < 0.001 | - |
| | Cohen's d | -0.309 | 0.42 | -0.415 | -0.41 | 0.721 | - |

**Keyword-prompted Writing**

|  | Statistics | GPT-3.5 | GPT-4 | Claude | Qwen | SparkDesk | Human |
|---|---|---|---|---|---|---|---|
| **Creativity** | $n$ | 50 | 50 | 24 | 46 | 31 | 221 |
| | $M$ | 2.02 | 3.341 | 2.594 | 2.31 | 2.395 | 2.417 |
| | $SD$ | 0.537 | 0.51 | 0.624 | 0.562 | 0.39 | 0.681 |
| | 95% CI (Δ) | [-0.573, 0.220] | [0.755, 1.094] | [-0.108, 0.464] | [-0.295, 0.082] | [-0.190, 0.145] | - |
| | $t$ | -4.473 | 10.825 | 1.223 | -1.124 | -0.265 | - |
| | $df$ | 88.471 | 93.123 | 243 | 75.203 | 59.633 | - |
| | $P$ | < 0.001 | < 0.001 | 0.223 | 0.265 | 0.792 | - |
| | Cohen's d | -0.604 | 1.416 | 0.263 | -0.161 | -0.034 | - |
| **Diversity** | $n$ | 50 | 50 | 24 | 46 | 31 | 178 |
| | $M$ | 2.12 | 2.85 | 2.928 | 2.336 | 2.191 | 3.301 |
| | $SD$ | 0.918 | 0.534 | 0.868 | 0.709 | 0.778 | 0.678 |
| | 95% CI (Δ) | [-1.460, 0.903] | [-0.656, 0.247] | [-0.752, 0.005] | [-1.188, 0.742] | [-1.376, 0.844] | - |
| | $t$ | -8.474 | -4.346 | -2.026 | -8.529 | -8.228 | - |
| | $df$ | 64.738 | 226 | 26.915 | 222 | 207 | - |
| | $P$ | < 0.001 | < 0.001 | 0.053 | < 0.001 | < 0.001 | - |
| | Cohen's d | -1.604 | -0.696 | -0.532 | -1.411 | -1.601 | - |

**Emoji-prompted Writing**

|  | Statistics | GPT-3.5 | GPT-4 | Claude | Qwen | SparkDesk | Human |
|---|---|---|---|---|---|---|---|
| **Task 1** | *n* | 50 | 50 | 48 | 10 | 33 | 201 |
| **Diversity** | *M* | 1.832 | 2.743 | 2.704 | 2.492 | 2.78 | 3.082 |
|  | *SD* | 0.731 | 0.67 | 0.973 | 0.693 | 0.727 | 0.725 |
|  | 95% CI (Δ) | [-1.476, -1.024] | [-0.561, -0.116] | [-0.677, -0.079] | [-1.051, -0.128] | [-0.570, -0.034] | - |
|  | *t* | -10.894 | -3.002 | -2.53 | -2.516 | -2.217 | - |
|  | *df* | 249 | 249 | 60.036 | 209 | 232 | - |
|  | *P* | < 0.001 | 0.003 | 0.014 | 0.013 | 0.028 | - |
|  | Cohen's *d* | -1.722 | -0.474 | -0.486 | -0.815 | -0.416 | - |
| **Task 2** | *n* | 38 | 50 | 50 | 12 | 4 | 181 |
| **Creativity** | *M* | 1.563 | 1.476 | 2.167 | 1.855 | 2.124 | 2.602 |
|  | *SD* | 0.341 | 0.258 | 0.45 | 0.483 | 0.383 | 0.777 |
|  | 95% CI (Δ) | [-1.197, -0.881] | [-1.261, -0.992] | [-0.605, -0.266] | [-1.069, -0.426] | [-1.248, -0.292] | - |
|  | *t* | -12.992 | -16.497 | -5.071 | -4.95 | -1.225 | - |
|  | *df* | 129.693 | 222.21 | 137.63 | 15.06 | 183 | - |
|  | *P* | < 0.001 | < 0.001 | < 0.001 | < 0.001 | 0.222 | - |
|  | Cohen's *d* | -1.441 | -1.612 | -0.605 | -0.98 | -0.619 | - |
| **Task 3** | *n* | 50 | 50 | 50 | - | - | 189 |
| **Creativity** | *M* | 2.032 | 2.083 | 2.421 | - | - | 2.557 |
|  | *SD* | 0.598 | 0.52 | 0.683 | - | - | 0.692 |
|  | 95% CI (Δ) | [-0.736, -0.314] | [-0.680, -0.267] | [-0.352, -0.081] | - | - | - |
|  | *t* | -4.899 | -4.514 | -1.234 | - | - | - |
|  | *df* | 237 | 237 | 237 | - | - | - |
|  | *P* | < 0.001 | < 0.001 | 0.218 | - | - | - |
|  | Cohen's *d* | -0.779 | -0.718 | -0.196 | - | - | - |
| **Creative Advert Writing** | | | | | | | |
|  | Statistics | GPT-3.5 | GPT-4 | Claude | Qwen | SparkDesk | Human |
| **Creativity** | *n* | 41 | 41 | 49 | 21 | 46 | 418 |
|  | *M* | 1.255 | 1.562 | 1.545 | 1.296 | 1.436 | 2.29 |
|  | *SD* | 0.166 | 0.143 | 0.209 | 0.106 | 0.196 | 0.612 |
|  | 95% CI (Δ) | [-1.113, -0.957] | [-0.801, -0.654] | [-0.828, -0.662] | [-1.069, -0.919] | [-0.936, -0.772] | - |
|  | *t* | -26.129 | -19.496 | -17.65 | -26.281 | -20.549 | - |
|  | *df* | 185.351 | 239.107 | 173.334 | 125.773 | 172.453 | - |
|  | *P* | < 0.001 | < 0.001 | < 0.001 | < 0.001 | < 0.001 | - |
|  | Cohen's *d* | -1.765 | -1.242 | -1.278 | -1.663 | -1.461 | - |

*Note.* *Before running the *t*-test, we tested if the populations had identical variances. If true, a standard independent sample *t*-test that assumes equal population variances was performed. If false, a Welch's *t*-test, which does not assume equal population variance, was performed.

**Table S5.** Descriptive statistics and independent sample *t*-test[*] (two-sided) results of diversity scores (the cosine similarity and the Levenshtein distance) between LLMs and human participants.

| Tasks | Statistics | GPT-3.5 | GPT-4 | Claude | Qwen | SparkDesk | Human |
|---|---|---|---|---|---|---|---|
| **Divergent Thinking** | | | | | | | |
| **Task 1** | *n* | 50 | 50 | 50 | 50 | 50 | 227 |
| **Cosine similarity** | *M* | 0.881 | 0.863 | 0.852 | 0.862 | 0.869 | 0.850 |
| | *SD* | 0.019 | 0.010 | 0.017 | 0.016 | 0.017 | 0.032 |
| | 95% CI (Δ) | [0.024, 0.038] | [0.007, 0.018] | [-0.005, 0.008] | [0.006, 0.018] | [0.012, 0.025] | - |
| | *t* | 9.040 | 4.896 | 0.537 | 3.768 | 5.805 | - |
| | *df* | 120.960 | 251.592 | 136.765 | 149.608 | 143.492 | - |
| | *P* | < 0.001 | < 0.001 | 0.592 | < 0.001 | < 0.001 | - |
| | Cohen's *d* | 1.022 | 0.423 | 0.058 | 0.391 | 0.612 | - |
| **Task 1** | *n* | 50 | 50 | 50 | 50 | 50 | 227 |
| **Levenshtein distance** | *M* | 0.741 | 0.853 | 0.897 | 0.833 | 0.841 | 0.885 |
| | *SD* | 0.099 | 0.045 | 0.027 | 0.059 | 0.053 | 0.067 |
| | 95% CI (Δ) | [-0.173, -0.114] | [-0.047, -0.016] | [0.001, 0.024] | [-0.071, -0.031] | [-0.063, -0.023] | - |
| | *t* | -9.716 | -4.073 | 2.087 | -4.984 | -4.266 | - |
| | *df* | 59.188 | 103.915 | 195.458 | 275 | 275 | - |
| | *P* | < 0.001 | < 0.001 | 0.038 | < 0.001 | < 0.001 | - |
| | Cohen's *d* | -1.940 | -0.495 | 0.198 | -0.779 | -0.666 | - |
| **Task 2** | *n* | 50 | 50 | 50 | 49 | 50 | 229 |
| **Cosine similarity** | *M* | 0.920 | 0.903 | 0.889 | 0.893 | 0.898 | 0.875 |
| | *SD* | 0.021 | 0.016 | 0.011 | 0.013 | 0.015 | 0.028 |
| | 95% CI (Δ) | [0.037, 0.054] | [0.023, 0.034] | [0.010, 0.019] | [0.013, 0.023] | [0.018, 0.029] | - |
| | *t* | 10.829 | 9.772 | 5.975 | 6.930 | 8.303 | - |
| | *df* | 277 | 126.195 | 193.227 | 159.293 | 136.061 | - |
| | *P* | < 0.001 | < 0.001 | < 0.001 | < 0.001 | < 0.001 | - |
| | Cohen's *d* | 1.690 | 1.081 | 0.566 | 0.696 | 0.890 | - |
| **Task 2** | *n* | 50 | 50 | 50 | 49 | 50 | 229 |
| **Levenshtein distance** | *M* | 0.614 | 0.734 | 0.849 | 0.800 | 0.771 | 0.836 |
| | *SD* | 0.160 | 0.089 | 0.036 | 0.057 | 0.067 | 0.082 |
| | 95% CI (Δ) | [-0.268, -0.175] | [-0.127, -0.076] | [-0.001, 0.028] | [-0.060, -0.011] | [-0.089, -0.040] | - |
| | *t* | -9.499 | -7.818 | 1.797 | -2.884 | -5.209 | - |
| | *df* | 54.731 | 277 | 177.069 | 276 | 277 | - |

|  | Statistics | GPT-3.5 | GPT-4 | Claude | Qwen | SparkDesk | Human |
|---|---|---|---|---|---|---|---|
|  | P | < 0.001 | < 0.001 | 0.074 | 0.004 | < 0.001 | - |
|  | Cohen's d | -2.204 | -1.220 | 0.175 | -0.454 | -0.813 | - |

**Scientific Problem Solving Task 1**

|  | Statistics | GPT-3.5 | GPT-4 | Claude | Qwen | SparkDesk | Human |
|---|---|---|---|---|---|---|---|
| **Cosine similarity** | n | 51 | 51 | 50 | 48 | 48 | 237 |
|  | M | 0.882 | 0.877 | 0.874 | 0.895 | 0.886 | 0.869 |
|  | SD | 0.015 | 0.015 | 0.016 | 0.024 | 0.014 | 0.026 |
|  | 95% CI (Δ) | [0.008, 0.019] | [0.003, 0.013] | [-0.001, 0.010] | [0.019, 0.035] | [0.012, 0.022] | - |
|  | t | 5.113 | 3.059 | 1.800 | 6.639 | 6.640 | - |
|  | df | 123.699 | 126.997 | 112.359 | 283 | 122.591 | - |
|  | P | < 0.001 | 0.003 | 0.074 | < 0.001 | < 0.001 | - |
|  | Cohen's d | 0.566 | 0.335 | 0.206 | 1.051 | 0.718 | - |
| **Levenshtein distance** | n | 51 | 51 | 50 | 48 | 48 | 237 |
|  | M | 0.805 | 0.814 | 0.872 | 0.770 | 0.776 | 0.861 |
|  | SD | 0.059 | 0.063 | 0.033 | 0.083 | 0.092 | 0.054 |
|  | 95% CI (Δ) | [-0.073, -0.040] | [-0.063, -0.030] | [-0.001, 0.022] | [-0.117, -0.066] | [-0.113, -0.058] | - |
|  | t | -6.656 | -5.462 | 1.835 | -7.307 | -6.223 | - |
|  | df | 286 | 286 | 111.951 | 55.149 | 53.627 | - |
|  | P | < 0.001 | < 0.001 | 0.069 | < 0.001 | < 0.001 | - |
|  | Cohen's d | -1.027 | -0.843 | 0.211 | -1.535 | -1.386 | - |

**Social Problem Solving Task 1**

|  | Statistics | GPT-3.5 | GPT-4 | Claude | Qwen | SparkDesk | Human |
|---|---|---|---|---|---|---|---|
| **Cosine similarity** | n | 51 | 51 | 50 | 50 | 50 | 229 |
|  | M | 0.886 | 0.885 | 0.869 | 0.877 | 0.874 | 0.856 |
|  | SD | 0.020 | 0.014 | 0.020 | 0.020 | 0.021 | 0.023 |
|  | 95% CI (Δ) | [0.023, 0.037] | [0.024, 0.034] | [0.006, 0.020] | [0.014, 0.028] | [0.011, 0.025] | - |
|  | t | 8.659 | 11.483 | 3.820 | 6.125 | 5.232 | - |
|  | df | 278 | 113.172 | 277 | 277 | 277 | - |
|  | P | < 0.001 | < 0.001 | < 0.001 | < 0.001 | < 0.001 | - |
|  | Cohen's d | 1.341 | 1.345 | 0.596 | 0.956 | 0.817 | - |
| **Levenshtein distance** | n | 51 | 51 | 50 | 50 | 50 | 229 |
|  | M | 0.841 | 0.873 | 0.878 | 0.846 | 0.844 | 0.899 |
|  | SD | 0.059 | 0.038 | 0.039 | 0.052 | 0.053 | 0.038 |
|  | 95% CI (Δ) | [-0.075, -0.040] | [-0.037, -0.014] | [-0.032, -0.009] | [-0.068, -0.037] | [-0.070, -0.038] | - |
|  | *t* | -6.635 | -4.318 | -3.445 | -6.688 | -6.828 | - |
|  | *df* | 59.392 | 278 | 277 | 60.619 | 60.190 | - |
|  | *P* | < 0.001 | < 0.001 | 0.001 | < 0.001 | < 0.001 | - |
|  | Cohen's *d* | -1.353 | -0.669 | -0.538 | -1.282 | -1.324 | - |
| **Keyword-prompted Writing** | | | | | | | |
|  | Statistics | **GPT-3.5** | **GPT-4** | **Claude** | **Qwen** | **SparkDesk** | **Human** |
| **Cosine similarity** | *n* | 50 | 50 | 24 | 46 | 31 | 178 |
|  | *M* | 0.930 | 0.928 | 0.918 | 0.931 | 0.950 | 0.883 |
|  | *SD* | 0.027 | 0.016 | 0.017 | 0.019 | 0.027 | 0.026 |
|  | 95% CI (Δ) | [0.038, 0.055] | [0.038, 0.051] | [0.027, 0.043] | [0.041, 0.055] | [0.057, 0.077] | - |
|  | *t* | 11.094 | 14.639 | 8.850 | 13.794 | 13.105 | - |
|  | *df* | 226 | 125.280 | 39.652 | 91.403 | 207 | - |
|  | *P* | < 0.001 | < 0.001 | < 0.001 | < 0.001 | < 0.001 | - |
|  | Cohen's *d* | 1.776 | 1.829 | 1.392 | 1.925 | 2.550 | - |
| **Levenshtein distance** | *n* | 50 | 50 | 24 | 46 | 31 | 178 |
|  | *M* | 0.716 | 0.882 | 0.911 | 0.815 | 0.680 | 0.921 |
|  | *SD* | 0.215 | 0.048 | 0.027 | 0.109 | 0.304 | 0.023 |
|  | 95% CI (Δ) | [-0.267, -0.144] | [-0.054, -0.026] | [-0.021, -0.000] | [-0.139, -0.074] | [-0.352, -0.130] | - |
|  | *t* | -6.751 | -5.670 | -2.060 | -6.612 | -4.419 | - |
|  | *df* | 49.321 | 55.554 | 200 | 46.065 | 30.061 | - |
|  | *P* | < 0.001 | < 0.001 | 0.041 | < 0.001 | < 0.001 | - |
|  | Cohen's *d* | -2.012 | -1.310 | -0.448 | -2.006 | -2.051 | - |
| **Emoji-prompted Writing Task 1** | | | | | | | |
|  | Statistics | **GPT-3.5** | **GPT-4** | **Claude** | **Qwen** | **SparkDesk** | **Human** |
| **Cosine similarity** | *n* | 50 | 50 | 48 | 10 | 33 | 194 |
|  | *M* | 0.893 | 0.860 | 0.869 | 0.872 | 0.869 | 0.860 |
|  | *SD* | 0.029 | 0.034 | 0.026 | 0.033 | 0.028 | 0.034 |
|  | 95% CI (Δ) | [0.023, 0.043] | [-0.010, 0.011] | [-0.001, 0.017] | [-0.010, 0.033] | [-0.004, 0.021] | - |
|  | *t* | 6.303 | 0.022 | 1.863 | 1.053 | 1.401 | - |
|  | *df* | 242 | 242 | 89.825 | 202 | 225 | - |
|  | *P* | < 0.001 | 0.982 | 0.066 | 0.294 | 0.163 | - |
|  | Cohen's *d* | 1.000 | 0.003 | 0.258 | 0.341 | 0.264 | - |
| **Levenshtein distance** | *n* | 50 | 50 | 48 | 10 | 33 | 194 |
|  | *M* | 0.754 | 0.898 | 0.911 | 0.787 | 0.898 | 0.907 |

| | | | | | | |
|---|---|---|---|---|---|---|
| *SD* | 0.196 | 0.035 | 0.035 | 0.104 | 0.037 | 0.041 |
| **95% CI (Δ)** | [-0.209, -0.097] | [-0.021, 0.004] | [-0.008, 0.017] | [-0.194, -0.045] | [-0.023, 0.007] | - |
| *t* | -5.500 | -1.355 | 0.653 | -3.612 | -1.095 | - |
| *df* | 50.110 | 242 | 240 | 9.143 | 225 | - |
| *P* | < 0.001 | 0.177 | 0.515 | 0.005 | 0.275 | - |
| **Cohen's d** | -1.606 | -0.215 | 0.105 | -2.620 | -0.206 | - |

*Note*. [*]Before running the *t*-test, we tested if the populations had identical variances. If true, a standard independent sample *t*-test that assumes equal population variances was performed. If false, a Welch's *t*-test, which does not assume equal population variance, was performed.

**Table S6. Text similarity indicated by the cosine similarity and the Levenshtein distance in the stories generated by LLMs and humans (averaged across all pairs).**

| Statistics | GPT-3.5 | GPT-4 | Claude | Qwen | SparkDesk | LLMs | Human |
|---|---|---|---|---|---|---|---|
| **Divergent Thinking Task 1** | | | | | | | |
| *n* | 445 | 748 | 542 | 427 | 298 | 2460 | 856 |
| *n* (pairs) | 98790 | 279378 | 146611 | 90951 | 44253 | 3024570 | 365940 |
| **Cosine similarity** | 0.868 | 0.859 | 0.841 | 0.850 | 0.865 | 0.849 | 0.830 |
| **Levenshtein distance** | 0.830 | 0.870 | 0.927 | 0.889 | 0.873 | 0.900 | 0.939 |
| **Divergent Thinking Task 2** | | | | | | | |
| *n* | 483 | 499 | 394 | 344 | 245 | 1965 | 857 |
| *n* (pairs) | 116403 | 124251 | 77421 | 58996 | 29890 | 1929630 | 366796 |
| **Cosine similarity** | 0.904 | 0.898 | 0.882 | 0.880 | 0.894 | 0.883 | 0.860 |
| **Levenshtein distance** | 0.789 | 0.774 | 0.878 | 0.871 | 0.840 | 0.866 | 0.897 |
| **Scientific Problem Solving Task 1** | | | | | | | |
| *n* | 153 | 153 | 150 | 143 | 143 | 742 | 696 |
| *n* (pairs) | 11628 | 11628 | 11175 | 10153 | 10153 | 274911 | 241860 |
| **Cosine similarity** | 0.892 | 0.883 | 0.876 | 0.886 | 0.889 | 0.877 | 0.851 |
| **Levenshtein distance** | 0.864 | 0.871 | 0.899 | 0.889 | 0.867 | 0.896 | 0.908 |
| **Scientific Problem Solving Task 2** | | | | | | | |
| *n* | 51 | 51 | 50 | - | 48 | 200 | 236 |
| *n* (pairs) | 1275 | 1275 | 1225 | - | 1128 | 19900 | 27730 |
| **Cosine similarity** | 0.944 | 0.967 | 0.952 | - | 0.954 | 0.940 | 0.898 |
| **Levenshtein distance** | 0.854 | 0.835 | 0.905 | - | 0.845 | 0.884 | 0.917 |
| **Scientific Problem Solving Task 3** | | | | | | | |
| *n* | 51 | 51 | 50 | 40 | 48 | 240 | 237 |
| *n* (pairs) | 1275 | 1275 | 1225 | 780 | 1128 | 28680 | 27966 |
| **Cosine similarity** | 0.927 | 0.942 | 0.908 | 0.901 | 0.944 | 0.912 | 0.871 |
| **Levenshtein distance** | 0.873 | 0.853 | 0.911 | 0.873 | 0.811 | 0.887 | 0.912 |
| **Social Problem Solving Task 1** | | | | | | | |
| *n* | 153 | 153 | 150 | 148 | 149 | 753 | 684 |
| *n* (pairs) | 11628 | 11628 | 11175 | 10878 | 11026 | 283128 | 233586 |
| **Cosine similarity** | 0.895 | 0.893 | 0.868 | 0.875 | 0.891 | 0.877 | 0.846 |
| **Levenshtein distance** | 0.856 | 0.883 | 0.905 | 0.890 | 0.847 | 0.894 | 0.917 |
| **Social Problem Solving Task 2** | | | | | | | |
| *n* | 51 | 51 | 50 | 50 | 50 | 252 | 229 |
| *n* (pairs) | 1275 | 1275 | 1225 | 1225 | 1225 | 31626 | 26106 |
| **Cosine similarity** | 0.921 | 0.932 | 0.923 | 0.919 | 0.944 | 0.914 | 0.872 |
| **Levenshtein distance** | 0.886 | 0.871 | 0.899 | 0.869 | 0.832 | 0.892 | 0.931 |
| **Social Problem Solving Task 3** | | | | | | | |
| *n* | 51 | 51 | 50 | 50 | 50 | 252 | 226 |
| *n* (pairs) | 1275 | 1275 | 1225 | 1225 | 1225 | 31626 | 25425 |
| **Cosine similarity** | 0.948 | 0.956 | 0.945 | 0.933 | 0.955 | 0.939 | 0.875 |
| **Levenshtein distance** | 0.849 | 0.860 | 0.896 | 0.841 | 0.820 | 0.874 | 0.929 |
| **Keyword-prompted Writing** | | | | | | | |
| *n* | 100 | 100 | 48 | 92 | 62 | 402 | 399 |
| *n* (pairs) | 4950 | 4950 | 1128 | 4186 | 1891 | 80601 | 79401 |
| **Cosine similarity** | 0.911 | 0.929 | 0.911 | 0.911 | 0.933 | 0.907 | 0.870 |
| **Levenshtein distance** | 0.872 | 0.887 | 0.915 | 0.887 | 0.867 | 0.897 | 0.922 |
| **Emoji-prompted Writing Task 1** | | | | | | | |
| *n* | 100 | 100 | 96 | 20 | 66 | 382 | 414 |
| *n* (pairs) | 4950 | 4950 | 4560 | 190 | 2145 | 72771 | 85491 |
| **Cosine similarity** | 0.888 | 0.871 | 0.864 | 0.856 | 0.871 | 0.857 | 0.836 |
| **Levenshtein distance** | 0.851 | 0.886 | 0.920 | 0.883 | 0.887 | 0.906 | 0.929 |
| **Emoji-prompted Writing Task 2** | | | | | | | |
| *n* | 38 | 50 | 50 | 12 | 4 | 154 | 181 |
| *n* (pairs) | 703 | 1225 | 1225 | 66 | 6 | 11781 | 16290 |
| **Cosine similarity** | 0.898 | 0.902 | 0.890 | 0.861 | 0.883 | 0.880 | 0.844 |
| **Levenshtein distance** | 0.878 | 0.868 | 0.913 | 0.901 | 0.869 | 0.906 | 0.927 |
| **Emoji-prompted Writing Task 3** | | | | | | | |
| *n* | 50 | 50 | 50 | - | - | 150 | 189 |
| *n* (pairs) | 1225 | 1225 | 1225 | - | - | 11175 | 17766 |
| **Cosine similarity** | 0.927 | 0.930 | 0.930 | - | - | 0.922 | 0.882 |
| **Levenshtein distance** | 0.873 | 0.866 | 0.905 | - | - | 0.895 | 0.919 |
| **Creative Advert Writing** | | | | | | | |
| *n* | 41 | 41 | 49 | 21 | 46 | 198 | 418 |
| *n* (pairs) | 820 | 820 | 1176 | 210 | 1035 | 19503 | 87153 |
| **Cosine similarity** | 0.895 | 0.906 | 0.859 | 0.881 | 0.900 | 0.877 | 0.825 |
| **Levenshtein distance** | 0.682 | 0.793 | 0.857 | 0.747 | 0.712 | 0.836 | 0.935 |

**Table S7. Descriptive statistics and one-way analysis of variance (ANOVA) results of GPT-3.5's performance at different temperatures.**

| Tasks | Statistics | Temperature | | | | | ANOVA Results | | |
|---|---|---|---|---|---|---|---|---|---|
| | | 0 | 0.25 | 0.5 | 0.75 | 1 | $F(4, 45)$ | $P$ | Partial $\eta^2$ |
| **Divergent Thinking** | | | | | | | | | |
| Task 1 Creativity Mean | M | 4.699 | 4.737 | 5.023 | 4.461 | 4.772 | 1.736 | 0.159 | 0.134 |
| | SD | 0.352 | 0.59 | 0.637 | 0.352 | 0.391 | | | |
| | n | 10 | 10 | 10 | 10 | 10 | | | |
| Task 1 Creativity Max | M | 6.62 | 6.825 | 6.926 | 6.321 | 7.04 | 0.7 | 0.596 | 0.059 |
| | SD | 0.88 | 1.402 | 1.079 | 0.458 | 1.276 | | | |
| | n | 10 | 10 | 10 | 10 | 10 | | | |
| Task 2 Creativity Mean | M | 4.917 | 4.674 | 4.899 | 4.723 | 4.881 | 0.851 | 0.501 | 0.07 |
| | SD | 0.176 | 0.237 | 0.488 | 0.39 | 0.505 | | | |
| | n | 10 | 10 | 10 | 10 | 10 | | | |
| Task 2 Creativity Max | M | 6.032 | 5.713 | 6.516 | 6.542 | 6.505 | 1.819 | 0.142 | 0.139 |
| | SD | 0.299 | 0.584 | 1.064 | 0.98 | 1.141 | | | |
| | n | 10 | 10 | 10 | 10 | 10 | | | |
| **Scientific Problem Solving** | | | | | | | | | |
| | Statistics | 0 | 0.25 | 0.5 | 0.75 | 1 | $F(3, 46)$ | $P$ | Partial $\eta^2$ |
| Task 1 Creativity | M | 1.545 | 1.6 | 1.567 | 1.844 | 2.094 | 10.282 | < 0.001 | 0.401 |
| | SD | - | 0.192 | 0.151 | 0.442 | 0.291 | | | |
| | n | 1 | 10 | 10 | 10 | 20 | | | |
| Task 2 Creativity | M | 1.772 | 1.977 | 2.182 | 2.284 | 2.519 | 1.96 | 0.133 | 0.113 |
| | SD | - | 0.528 | 0.492 | 0.62 | 0.675 | | | |
| | n | 1 | 10 | 10 | 10 | 20 | | | |
| Task 3 Creativity | M | 1.42 | 1.357 | 1.441 | 1.483 | 1.821 | 4.38 | 0.009 | 0.222 |
| | SD | - | 0.174 | 0.156 | 0.245 | 0.552 | | | |
| | n | 1 | 10 | 10 | 10 | 20 | | | |
| **Social Problem Solving** | | | | | | | | | |
| | Statistics | 0 | 0.25 | 0.5 | 0.75 | 1 | $F(3, 46)$ | $P$ | Partial $\eta^2$ |
| Task 1 Flexibility | M | 2 | 2.1 | 2.4 | 2.3 | 2 | 1.163 | 0.334 | 0.071 |
| | SD | - | 0.568 | 0.699 | 0.483 | 0.649 | | | |
| | n | 1 | 10 | 10 | 10 | 20 | | | |
| Task 2 Creativity | M | 2.922 | 2.996 | 2.996 | 2.736 | 2.774 | 0.734 | 0.537 | 0.046 |
| | SD | - | 0.422 | 0.393 | 0.753 | 0.561 | | | |
| | n | 1 | 10 | 10 | 10 | 20 | | | |

| | Statistics | 0 | 0.25 | 0.5 | 0.75 | 1 | F(4, 45) | P | Partial η² |
|---|---|---|---|---|---|---|---|---|---|
| Task 3 Creativity | M | 2.09 | 2.462 | 2.462 | 2.169 | 2.582 | 0.908 | 0.445 | 0.056 |
| | SD | - | 0.655 | 0.745 | 0.18 | 0.729 | | | |
| | n | 1 | 10 | 10 | 10 | 20 | | | |
| **Keyword-prompted Writing** | | | | | | | | | |
| | Statistics | **0** | **0.25** | **0.5** | **0.75** | **1** | **F(4, 45)** | **P** | **Partial η²** |
| Creativity | M | 1.5 | 1.87 | 2.08 | 2.36 | 2.29 | 5.873 | 0.001 | 0.343 |
| | SD | 0.249 | 0.374 | 0.561 | 0.515 | 0.5 | | | |
| | n | 10 | 10 | 10 | 10 | 10 | | | |
| Diversity | M | 1.4 | 1.62 | 2.12 | 2.76 | 2.7 | 6.519 | < 0.001 | 0.367 |
| | SD | 0.298 | 0.614 | 1.08 | 0.771 | 0.823 | | | |
| | n | 10 | 10 | 10 | 10 | 10 | | | |
| **Emoji-prompted Writing** | | | | | | | | | |
| | Statistics | **0** | **0.25** | **0.5** | **0.75** | **1** | **F(4, 45)** | **P** | **Partial η²** |
| Task 1 Diversity | M | 1.1 | 1.88 | 2.1 | 1.9 | 2.18 | 4.395 | 0.004 | 0.281 |
| | SD | 0.141 | 0.738 | 0.492 | 0.392 | 1.064 | | | |
| | n | 10 | 10 | 10 | 10 | 10 | | | |
| Task 2 Creativity | M | 1.6 | 1.725 | 1.54 | 1.56 | 1.444 | 0.948* | 0.429 | 0.079 |
| | SD | - | 0.477 | 0.232 | 0.263 | 0.397 | | | |
| | n | 1 | 8 | 10 | 10 | 9 | | | |
| Task 3 Creativity | M | 2.04 | 1.86 | 1.82 | 2.42 | 2.02 | 1.657 | 0.177 | 0.128 |
| | SD | 0.564 | 0.212 | 0.371 | 0.835 | 0.708 | | | |
| | n | 10 | 10 | 10 | 10 | 10 | | | |
| **Creative Advert Writing** | | | | | | | | | |
| | Statistics | **0** | **0.25** | **0.5** | **0.75** | **1** | **F(3, 36)** | **P** | **Partial η²** |
| Creativity | M | 1.741 | 1.309 | 1.212 | 1.221 | 1.23 | 0.894 | 0.454 | 0.069 |
| | SD | - | 0.183 | 0.125 | 0.128 | 0.154 | | | |
| | n | 1 | 10 | 10 | 10 | 10 | | | |

*Note.* *$df = (3, 33)$.

**Table S8. Descriptive statistics and one-way ANOVA results of GPT-4's performance at different temperatures.**

| Tasks | Statistics | Temperature | | | | | ANOVA Results | | |
|---|---|---|---|---|---|---|---|---|---|
| | | 0 | 0.25 | 0.5 | 0.75 | 1 | $F(4, 45)$ | $P$ | Partial $\eta^2$ |
| **Divergent Thinking** | | | | | | | | | |
| Task 1 Creativity Mean | $M$ | 4.787 | 4.687 | 4.672 | 4.74 | 4.741 | 0.211 | 0.931 | 0.018 |
| | $SD$ | 0.385 | 0.219 | 0.254 | 0.34 | 0.367 | | | |
| | $n$ | 10 | 10 | 10 | 10 | 10 | | | |
| Task 1 Creativity Max | $M$ | 7.626 | 7.488 | 7.45 | 7.182 | 6.934 | 0.634 | 0.641 | 0.053 |
| | $SD$ | 1.062 | 0.949 | 1.364 | 1.076 | 0.996 | | | |
| | $n$ | 10 | 10 | 10 | 10 | 10 | | | |
| Task 2 Creativity Mean | $M$ | 4.964 | 5.039 | 5.129 | 5.077 | 4.747 | 1.934 | 0.121 | 0.147 |
| | $SD$ | 0.314 | 0.323 | 0.23 | 0.396 | 0.402 | | | |
| | $n$ | 10 | 10 | 10 | 10 | 10 | | | |
| Task 2 Creativity Max | $M$ | 8.214 | 7.848 | 7.655 | 7.841 | 6.776 | 1.81 | 0.143 | 0.139 |
| | $SD$ | 1.094 | 1.146 | 1.048 | 1.652 | 1.283 | | | |
| | $n$ | 10 | 10 | 10 | 10 | 10 | | | |
| **Scientific Problem Solving** | | | | | | | | | |
| | Statistics | 0 | 0.25 | 0.5 | 0.75 | 1 | $F(3, 46)$ | $P$ | Partial $\eta^2$ |
| Task 1 Creativity | $M$ | 2.642 | 2.471 | 2.503 | 2.452 | 2.523 | 0.137 | 0.937 | 0.009 |
| | $SD$ | - | 0.271 | 0.255 | 0.28 | 0.365 | | | |
| | $n$ | 1 | 10 | 10 | 10 | 20 | | | |
| Task 2 Creativity | $M$ | 1.977 | 2.223 | 2.267 | 2.386 | 2.632 | 4.425 | 0.008 | 0.224 |
| | $SD$ | - | 0.345 | 0.09 | 0.386 | 0.385 | | | |
| | $n$ | 1 | 10 | 10 | 10 | 20 | | | |
| Task 3 Creativity | $M$ | 1.42 | 1.546 | 1.744 | 1.546 | 1.905 | 1.937 | 0.137 | 0.112 |
| | $SD$ | - | 0.302 | 0.468 | 0.436 | 0.548 | | | |
| | $n$ | 1 | 10 | 10 | 10 | 20 | | | |
| **Social Problem Solving** | | | | | | | | | |
| | Statistics | 0 | 0.25 | 0.5 | 0.75 | 1 | $F(3, 46)$ | $P$ | Partial $\eta^2$ |
| Task 1 Flexibility | $M$ | 3 | 2.9 | 2.9 | 2.7 | 2.85 | 0.642 | 0.592 | 0.04 |
| | $SD$ | - | 0.316 | 0.316 | 0.483 | 0.366 | | | |
| | $n$ | 1 | 10 | 10 | 10 | 20 | | | |

| | Statistics | 0 | 0.25 | 0.5 | 0.75 | 1 | F(4, 45) | P | Partial η² |
|---|---|---|---|---|---|---|---|---|---|
| Task 2 Creativity | M | 2.551 | 2.142 | 2.402 | 2.421 | 2.746 | 3.692 | 0.018 | 0.194 |
| | SD | - | 0.325 | 0.494 | 0.48 | 0.551 | | | |
| | n | 1 | 10 | 10 | 10 | 20 | | | |
| Task 3 Creativity | M | 3.687 | 2.809 | 3.102 | 3.075 | 3.088 | 0.26 | 0.854 | 0.017 |
| | SD | - | 0.754 | 1.064 | 0.94 | 0.855 | | | |
| | n | 1 | 10 | 10 | 10 | 20 | | | |

**Keyword-prompted Writing**

| | Statistics | 0 | 0.25 | 0.5 | 0.75 | 1 | F(4, 45) | P | Partial η² |
|---|---|---|---|---|---|---|---|---|---|
| Creativity | M | 3.342 | 3.287 | 2.992 | 3.386 | 3.7 | 2.816 | 0.036 | 0.2 |
| | SD | 0.523 | 0.345 | 0.541 | 0.541 | 0.393 | | | |
| | n | 10 | 10 | 10 | 10 | 10 | | | |
| Diversity | M | 2.683 | 2.522 | 2.914 | 3.122 | 3.007 | 2.315 | 0.072 | 0.171 |
| | SD | 0.639 | 0.614 | 0.276 | 0.52 | 0.393 | | | |
| | n | 10 | 10 | 10 | 10 | 10 | | | |

**Emoji-prompted Writing**

| | Statistics | 0 | 0.25 | 0.5 | 0.75 | 1 | F(4, 45) | P | Partial η² |
|---|---|---|---|---|---|---|---|---|---|
| Task 1 Diversity | M | 2.415 | 2.722 | 2.594 | 3.004 | 2.978 | 1.463 | 0.229 | 0.115 |
| | SD | 0.319 | 0.669 | 0.54 | 0.853 | 0.77 | | | |
| | n | 10 | 10 | 10 | 10 | 10 | | | |
| Task 2 Creativity | M | 1.407 | 1.447 | 1.629 | 1.488 | 1.407 | 1.307 | 0.282 | 0.104 |
| | SD | 0.192 | 0.26 | 0.327 | 0.293 | 0.166 | | | |
| | n | 10 | 10 | 10 | 10 | 10 | | | |
| Task 3 Creativity | M | 2.166 | 1.935 | 2.329 | 2.027 | 1.958 | 0.997 | 0.419 | 0.081 |
| | SD | 0.479 | 0.466 | 0.536 | 0.598 | 0.513 | | | |
| | n | 10 | 10 | 10 | 10 | 10 | | | |

**Creative Advert Writing**

| | Statistics | 0 | 0.25 | 0.5 | 0.75 | 1 | F(3, 36) | P | Partial η² |
|---|---|---|---|---|---|---|---|---|---|
| Creativity | M | 1.476 | 1.529 | 1.547 | 1.512 | 1.67 | 2.857 | 0.051 | 0.192 |
| | SD | - | 0.062 | 0.123 | 0.126 | 0.194 | | | |
| | n | 1 | 10 | 10 | 10 | 10 | | | |

**Table S9. Descriptive statistics and one-way ANOVA results of GPT-3.5's diversity at different temperatures.**

| Tasks | Statistics | Temperature | | | | | ANOVA Results | | |
|---|---|---|---|---|---|---|---|---|---|
| | | 0 | 0.25 | 0.5 | 0.75 | 1 | $F(4, 45)$ | $P$ | Partial $\eta^2$ |
| **Divergent Thinking Task 1** | | | | | | | | | |
| Cosine similarity | $M$ | 0.891 | 0.895 | 0.890 | 0.873 | 0.857 | 15.841 | < 0.001 | 0.585 |
| | $SD$ | 0.008 | 0.013 | 0.018 | 0.009 | 0.013 | | | |
| | $n$ | 10 | 10 | 10 | 10 | 10 | | | |
| Levenshtein distance | $M$ | 0.689 | 0.646 | 0.702 | 0.805 | 0.866 | 24.430 | < 0.001 | 0.685 |
| | $SD$ | 0.069 | 0.063 | 0.066 | 0.049 | 0.038 | | | |
| | $n$ | 10 | 10 | 10 | 10 | 10 | | | |
| **Divergent Thinking Task 2** | | | | | | | | | |
| Cosine similarity | $M$ | 0.942 | 0.931 | 0.922 | 0.911 | 0.895 | 17.696 | < 0.001 | 0.611 |
| | $SD$ | 0.006 | 0.017 | 0.016 | 0.015 | 0.010 | | | |
| | $n$ | 10 | 10 | 10 | 10 | 10 | | | |
| Levenshtein distance | $M$ | 0.408 | 0.528 | 0.618 | 0.705 | 0.811 | 37.922 | < 0.001 | 0.771 |
| | $SD$ | 0.014 | 0.095 | 0.122 | 0.077 | 0.043 | | | |
| | $n$ | 10 | 10 | 10 | 10 | 10 | | | |
| **Scientific Problem Solving Task 1** | | | | | | | | | |
| | Statistics | 0 | 0.25 | 0.5 | 0.75 | 1 | $F(3, 46)$ | $P$ | Partial $\eta^2$ |
| Cosine similarity | $M$ | 0.873 | 0.874 | 0.880 | 0.888 | 0.885 | 1.904 | 0.142 | 0.110 |
| | $SD$ | - | 0.019 | 0.012 | 0.014 | 0.014 | | | |
| | $n$ | 1 | 10 | 10 | 10 | 20 | | | |
| Levenshtein distance | $M$ | 0.808 | 0.749 | 0.784 | 0.800 | 0.846 | 10.312 | < 0.001 | 0.402 |
| | $SD$ | | 0.056 | 0.063 | 0.039 | 0.037 | | | |
| | $n$ | 1 | 10 | 10 | 10 | 20 | | | |
| **Social Problem Solving Task 1** | | | | | | | | | |
| | Statistics | 0 | 0.25 | 0.5 | 0.75 | 1 | $F(3, 46)$ | $P$ | Partial $\eta^2$ |
| Cosine similarity | $M$ | 0.898 | 0.895 | 0.882 | 0.887 | 0.883 | 0.947 | 0.426 | 0.058 |
| | $SD$ | - | 0.007 | 0.014 | 0.014 | 0.028 | | | |
| | $n$ | 1 | 10 | 10 | 10 | 20 | | | |
| Levenshtein distance | $M$ | 0.834 | 0.781 | 0.845 | 0.843 | 0.869 | 6.449 | 0.001 | 0.296 |
| | $SD$ | - | 0.077 | 0.031 | 0.054 | 0.043 | | | |

|  | *n* | 1 | 10 | 10 | 10 | 20 | | | |
|---|---|---|---|---|---|---|---|---|---|

**Keyword-prompted Writing**

| | Statistics | 0 | 0.25 | 0.5 | 0.75 | 1 | $F(4, 45)$ | $P$ | Partial $\eta^2$ |
|---|---|---|---|---|---|---|---|---|---|
| Cosine similarity | *M* | 0.948 | 0.948 | 0.927 | 0.911 | 0.914 | 6.635 | < 0.001 | 0.371 |
| | *SD* | 0.011 | 0.023 | 0.030 | 0.023 | 0.017 | | | |
| | *n* | 10 | 10 | 10 | 10 | 10 | | | |
| Levenshtein distance | *M* | 0.530 | 0.555 | 0.741 | 0.856 | 0.897 | 11.247 | < 0.001 | 0.500 |
| | *SD* | 0.164 | 0.223 | 0.213 | 0.057 | 0.021 | | | |
| | *n* | 10 | 10 | 10 | 10 | 10 | | | |

**Emoji-prompted Writing Task 1**

| | Statistics | 0 | 0.25 | 0.5 | 0.75 | 1 | $F(4, 45)$ | $P$ | Partial $\eta^2$ |
|---|---|---|---|---|---|---|---|---|---|
| Cosine similarity | *M* | 0.924 | 0.885 | 0.883 | 0.893 | 0.880 | 4.923 | 0.002 | 0.304 |
| | *SD* | 0.016 | 0.036 | 0.021 | 0.022 | 0.029 | | | |
| | *n* | 10 | 10 | 10 | 10 | 10 | | | |
| Levenshtein distance | *M* | 0.432 | 0.762 | 0.827 | 0.877 | 0.870 | 30.948 | < 0.001 | 0.733 |
| | *SD* | 0.185 | 0.107 | 0.082 | 0.034 | 0.044 | | | |
| | n | 10 | 10 | 10 | 10 | 10 | | | |

**Table S10. Descriptive statistics and one-way ANOVA results of GPT-4's diversity at different temperatures.**

| Tasks | Statistics | Temperature | | | | | ANOVA Results | | |
|---|---|---|---|---|---|---|---|---|---|
| | | **0** | **0.25** | **0.5** | **0.75** | **1** | $F(4, 45)$ | $P$ | Partial $\eta^2$ |
| **Divergent Thinking Task 1** | | | | | | | | | |
| Cosine similarity | $M$ | 0.869 | 0.869 | 0.862 | 0.859 | 0.855 | 5.400 | 0.001 | 0.324 |
| | $SD$ | 0.010 | 0.007 | 0.009 | 0.009 | 0.006 | | | |
| | $n$ | 10 | 10 | 10 | 10 | 10 | | | |
| Levenshtein distance | $M$ | 0.802 | 0.832 | 0.855 | 0.878 | 0.899 | 15.967 | < 0.001 | 0.587 |
| | $SD$ | 0.055 | 0.025 | 0.013 | 0.021 | 0.015 | | | |
| | $n$ | 10 | 10 | 10 | 10 | 10 | | | |
| **Divergent Thinking Task 2** | | | | | | | | | |
| Cosine similarity | $M$ | 0.915 | 0.916 | 0.908 | 0.896 | 0.881 | 24.007 | < 0.001 | 0.681 |
| | $SD$ | 0.003 | 0.006 | 0.014 | 0.012 | 0.007 | | | |
| | $n$ | 10 | 10 | 10 | 10 | 10 | | | |
| Levenshtein distance | $M$ | 0.672 | 0.658 | 0.701 | 0.780 | 0.860 | 31.904 | < 0.001 | 0.739 |
| | $SD$ | 0.038 | 0.034 | 0.068 | 0.053 | 0.033 | | | |
| | $n$ | 10 | 10 | 10 | 10 | 10 | | | |
| **Scientific Problem Solving Task 1** | | | | | | | | | |
| | Statistics | **0** | **0.25** | **0.5** | **0.75** | **1** | $F(3, 46)$ | $P$ | Partial $\eta^2$ |
| Cosine similarity | $M$ | 0.864 | 0.885 | 0.869 | 0.878 | 0.876 | 2.202 | 0.101 | 0.126 |
| | $SD$ | | 0.012 | 0.011 | 0.012 | 0.017 | | | |
| | $n$ | 1 | 10 | 10 | 10 | 20 | | | |
| Levenshtein distance | $M$ | 0.737 | 0.748 | 0.777 | 0.837 | 0.859 | 18.587 | < 0.001 | 0.548 |
| | $SD$ | | 0.062 | 0.053 | 0.029 | 0.031 | | | |
| | $n$ | 1 | 10 | 10 | 10 | 20 | | | |
| **Social Problem Solving Task 1** | | | | | | | | | |
| | Statistics | **0** | **0.25** | **0.5** | **0.75** | **1** | $F(3, 46)$ | $P$ | Partial $\eta^2$ |
| Cosine similarity | $M$ | 0.888 | 0.886 | 0.893 | 0.874 | 0.886 | 3.219 | 0.031 | 0.174 |
| | $SD$ | | 0.007 | 0.008 | 0.017 | 0.016 | | | |
| | $n$ | 1 | 10 | 10 | 10 | 20 | | | |
| Levenshtein distance | $M$ | 0.881 | 0.858 | 0.867 | 0.874 | 0.883 | 1.043 | 0.383 | 0.064 |
| | $SD$ | | 0.026 | 0.023 | 0.063 | 0.035 | | | |

|  |  | n | 1 | 10 | 10 | 10 | 20 |  |  |  |
|---|---|---|---|---|---|---|---|---|---|---|
| **Keyword-prompted Writing** | | | | | | | | | | |
| | **Statistics** | | **0** | **0.25** | **0.5** | **0.75** | **1** | $F(4, 45)$ | $P$ | Partial $\eta^2$ |
| **Cosine similarity** | $M$ | | 0.930 | 0.936 | 0.925 | 0.927 | 0.919 | 1.495 | 0.220 | 0.117 |
| | $SD$ | | 0.015 | 0.014 | 0.017 | 0.018 | 0.017 | | | |
| | $n$ | | 10 | 10 | 10 | 10 | 10 | | | |
| **Levenshtein distance** | $M$ | | 0.862 | 0.845 | 0.881 | 0.901 | 0.919 | 4.990 | 0.002 | 0.307 |
| | $SD$ | | 0.034 | 0.074 | 0.033 | 0.021 | 0.022 | | | |
| | $n$ | | 10 | 10 | 10 | 10 | 10 | | | |
| **Emoji-prompted Writing Task 1** | | | | | | | | | | |
| | **Statistics** | | **0** | **0.25** | **0.5** | **0.75** | **1** | $F(4, 45)$ | $P$ | Partial $\eta^2$ |
| **Cosine similarity** | $M$ | | 0.880 | 0.874 | 0.869 | 0.843 | 0.837 | 4.021 | 0.007 | 0.263 |
| | $SD$ | | 0.017 | 0.035 | 0.023 | 0.038 | 0.034 | | | |
| | $n$ | | 10 | 10 | 10 | 10 | 10 | | | |
| **Levenshtein distance** | $M$ | | 0.875 | 0.892 | 0.894 | 0.909 | 0.921 | 3.021 | 0.027 | 0.212 |
| | $SD$ | | 0.029 | 0.037 | 0.030 | 0.026 | 0.036 | | | |
| | $n$ | | 10 | 10 | 10 | 10 | 10 | | | |

**Table S11. Statistics of top 10 ideas from pooled responses of LLMs and human participants.**

| Tasks | Numbers and proportions of total responses | | | | Proportions of top 10 responses | | | | | |
|---|---|---|---|---|---|---|---|---|---|---|
| | Human | LLM | Human % | LLM% | Human | GPT-3.5 | GPT-4 | Claude | Qwen | SparkDesk |
| Divergent Thinking Task 1 | 856 | 2460 | 25.81% | 74.19% | 90.00% | 10.00% | 0.00% | 0.00% | 0.00% | 0.00% |
| Divergent Thinking Task 2 | 857 | 1965 | 30.37% | 69.63% | 100.00% | 0.00% | 0.00% | 0.00% | 0.00% | 0.00% |
| Scientific Problem Solving Task 1 | 696 | 742 | 48.40% | 51.60% | 42.86% | 7.14% | 14.29% | 28.57% | 0.00% | 7.14% |
| Scientific Problem Solving Task 2 | 236 | 200 | 54.13% | 45.87% | 60.0% | 0.00% | 0.00% | 30.00% | 0.00% | 10.00% |
| Scientific Problem Solving Task 3 | 237 | 240 | 49.69% | 50.31% | 45.45% | 0.00% | 0.00% | 18.18% | 18.18% | 18.18% |
| Social Problem Solving Task 2 | 229 | 252 | 47.61% | 52.39% | 66.67% | 0.00% | 0.00% | 8.33% | 25.0% | 0.00% |
| Social Problem Solving Task 3 | 226 | 252 | 47.28% | 52.72% | 35.71% | 7.14% | 21.43% | 7.14% | 0.00% | 28.57% |
| Keyword-prompted Writing | 399 | 402 | 49.81% | 50.19% | 21.43% | 0.00% | 64.29% | 14.29% | 0.00% | 0.00% |
| Emoji-prompted Writing Task 2 | 181 | 177 | 50.56% | 49.44% | 100.00% | 0.00% | 0.00% | 0.00% | 0.00% | 0.00% |
| Emoji-prompted Writing Task 3 | 189 | 245 | 43.55% | 56.45% | 80.00% | 0.00% | 0.00% | 20.00% | 0.00% | 0.00% |
| Creative Advert Writing | 418 | 198 | 67.86% | 32.14% | 100.00% | 0.00% | 0.00% | 0.00% | 0.00% | 0.00% |

**Table S12. Percentages of the top 10 responses contributed by humans in a pool combining responses of a certain number of humans with those from one LLM that is asked 10 times.**

**GPT-3.5**

| Tasks | N=1 | N=2 | N=3 | N=4 | N=5 | N=6 | N=7 | N=8 | N=9 | N=10 | N=11 | N=12 | N=13 | N=14 | N=15 |
|---|---|---|---|---|---|---|---|---|---|---|---|---|---|---|---|
| Divergent Thinking Task 1 | 8.04% | 14.86% | 20.86% | 26.52% | 31.85% | 36.58% | 40.85% | 45.08% | 48.78% | 52.08% | 55.04% | 57.86% | 60.76% | 62.94% | 64.87% |
| Divergent Thinking Task 2 | 8.92% | 16.40% | 22.66% | 28.54% | 33.72% | 38.11% | 43.20% | 47.07% | 50.97% | 54.62% | 58.46% | 61.30% | 64.09% | 66.33% | 68.62% |
| Scientific Problem Solving Task 1 | 7.80% | 15.02% | 21.28% | 26.46% | 30.62% | 34.61% | 37.56% | 40.57% | 43.60% | 46.31% | 48.62% | 51.26% | 53.43% | 55.25% | 57.35% |
| Scientific Problem Solving Task 2 | 7.44% | 15.27% | 22.08% | 28.54% | 34.27% | 39.12% | 43.84% | 48.60% | 52.46% | 55.79% | 58.70% | 61.36% | 63.42% | 64.97% | 66.50% |
| Scientific Problem Solving Task 3 | 8.86% | 18.80% | 26.19% | 31.83% | 38.16% | 44.85% | 51.92% | 62.81% | 72.61% | 80.13% | 85.06% | 88.01% | 89.94% | 91.30% | 92.28% |
| Social Problem Solving Task 2 | 6.60% | 10.39% | 13.12% | 15.43% | 17.26% | 18.83% | 20.08% | 21.58% | 23.08% | 24.30% | 25.73% | 26.95% | 28.07% | 29.24% | 30.62% |
| Social Problem Solving Task 3 | 7.08% | 13.16% | 18.35% | 23.02% | 27.48% | 32.69% | 38.72% | 45.09% | 50.82% | 56.22% | 60.61% | 64.97% | 68.25% | 70.97% | 72.56% |
| Keyword-prompted Writing | 11.08% | 21.24% | 29.49% | 36.32% | 41.60% | 47.06% | 51.49% | 55.85% | 60.14% | 63.57% | 67.32% | 70.25% | 73.43% | 76.11% | 77.80% |
| Emoji-prompted Writing Task 2 | 9.74% | 18.93% | 27.42% | 33.19% | 42.31% | 49.96% | 57.50% | 65.43% | 72.17% | 77.74% | 83.08% | 87.74% | 91.07% | 93.75% | 95.68% |
| Emoji-prompted Writing Task 3 | 9.64% | 17.94% | 27.56% | 34.29% | 40.59% | 47.28% | 52.97% | 57.76% | 61.82% | 65.58% | 68.98% | 71.98% | 74.53% | 76.50% | 78.05% |
| Creative Advert Writing | 9.11% | 16.74% | 23.21% | 31.09% | 46.89% | 52.35% | 68.46% | 78.13% | 88.69% | 94.65% | 98.04% | 99.11% | 99.63% | 99.84% | 99.93% |
| **Overall** | 8.57% | 16.25% | 22.93% | 28.66% | 34.98% | 40.13% | 46.05% | 51.63% | 56.83% | 61.00% | 64.51% | 67.34% | 69.69% | 71.56% | 73.11% |

**GPT-4**

| Tasks | N=1 | N=2 | N=3 | N=4 | N=5 | N=6 | N=7 | N=8 | N=9 | N=10 | N=11 | N=12 | N=13 | N=14 | N=15 |
|---|---|---|---|---|---|---|---|---|---|---|---|---|---|---|---|
| Divergent Thinking Task 1 | 6.98% | 12.90% | 18.33% | 23.23% | 27.13% | 31.25% | 34.61% | 37.35% | 40.38% | 42.87% | 45.27% | 47.72% | 49.83% | 51.47% | 53.63% |

| Tasks | N=1 | N=2 | N=3 | N=4 | N=5 | N=6 | N=7 | N=8 | N=9 | N=10 | N=11 | N=12 | N=13 | N=14 | N=15 |
|---|---|---|---|---|---|---|---|---|---|---|---|---|---|---|---|
| Divergent Thinking Task 2 | 5.54% | 9.56% | 13.05% | 15.84% | 18.62% | 20.93% | 23.22% | 25.38% | 26.97% | 29.04% | 30.49% | 32.08% | 33.95% | 35.10% | 36.62% |
| Scientific Problem Solving Task 1 | 2.13% | 4.25% | 6.42% | 8.33% | 10.52% | 12.28% | 14.18% | 15.78% | 17.46% | 19.22% | 20.62% | 22.09% | 23.79% | 25.12% | 26.73% |
| Scientific Problem Solving Task 2 | 7.02% | 12.85% | 17.83% | 22.74% | 26.86% | 31.02% | 35.09% | 38.69% | 42.31% | 45.64% | 48.97% | 51.90% | 55.28% | 57.84% | 60.53% |
| Scientific Problem Solving Task 3 | 9.15% | 17.29% | 26.19% | 34.52% | 42.30% | 50.57% | 57.84% | 64.64% | 69.87% | 73.97% | 77.26% | 79.64% | 82.00% | 84.03% | 85.89% |
| Social Problem Solving Task 2 | 5.52% | 9.38% | 12.81% | 15.99% | 18.95% | 21.74% | 24.79% | 27.53% | 30.46% | 33.12% | 35.59% | 37.99% | 40.73% | 43.01% | 44.64% |
| Social Problem Solving Task 3 | 7.97% | 13.81% | 20.28% | 24.28% | 27.74% | 30.63% | 33.31% | 36.13% | 39.07% | 41.54% | 43.51% | 45.93% | 47.76% | 49.27% | 50.91% |
| Keyword-prompted Writing | 3.08% | 6.51% | 9.62% | 12.56% | 15.50% | 17.89% | 20.25% | 22.55% | 24.67% | 26.58% | 28.69% | 30.15% | 31.65% | 32.84% | 34.70% |
| Emoji-prompted Writing Task 2 | 8.86% | 16.31% | 23.98% | 31.61% | 42.30% | 49.47% | 61.62% | 68.51% | 75.61% | 82.64% | 88.07% | 92.07% | 94.82% | 96.84% | 98.24% |
| Emoji-prompted Writing Task 3 | 9.52% | 19.37% | 27.40% | 34.72% | 42.10% | 46.75% | 49.99% | 53.41% | 56.35% | 59.52% | 62.39% | 65.41% | 69.01% | 71.88% | 74.20% |
| Creative Advert Writing | 9.64% | 18.15% | 25.65% | 33.17% | 38.62% | 45.66% | 55.17% | 65.62% | 71.86% | 81.68% | 88.78% | 93.24% | 95.66% | 97.02% | 97.87% |
| **Overall** | 6.85% | 12.76% | 18.32% | 23.36% | 28.24% | 32.56% | 37.28% | 41.42% | 45.00% | 48.71% | 51.78% | 54.38% | 56.77% | 58.58% | 60.36% |

**Claude**

| Tasks | N=1 | N=2 | N=3 | N=4 | N=5 | N=6 | N=7 | N=8 | N=9 | N=10 | N=11 | N=12 | N=13 | N=14 | N=15 |
|---|---|---|---|---|---|---|---|---|---|---|---|---|---|---|---|
| Divergent Thinking Task 1 | 6.23% | 12.57% | 18.21% | 24.01% | 29.16% | 34.79% | 39.66% | 44.25% | 48.64% | 52.11% | 55.82% | 59.25% | 62.20% | 65.19% | 67.77% |
| Divergent Thinking Task 2 | 6.91% | 13.06% | 18.69% | 24.61% | 29.03% | 33.43% | 37.52% | 40.50% | 43.76% | 46.18% | 48.75% | 50.91% | 53.21% | 54.69% | 56.67% |
| Scientific Problem Solving Task 1 | 3.24% | 6.20% | 8.98% | 11.62% | 14.21% | 16.45% | 18.45% | 20.83% | 22.91% | 24.66% | 26.68% | 28.30% | 30.30% | 31.95% | 33.52% |
| Scientific Problem Solving Task 2 | 7.52% | 14.24% | 19.17% | 23.32% | 27.46% | 30.46% | 33.56% | 35.88% | 38.16% | 40.41% | 42.16% | 44.20% | 45.93% | 47.79% | 49.42% |
| Scientific Problem Solving Task 3 | 8.46% | 14.21% | 20.16% | 25.70% | 31.53% | 36.59% | 41.36% | 44.47% | 47.12% | 49.97% | 52.57% | 54.42% | 56.51% | 58.27% | 59.96% |
| Social Problem Solving Task 2 | 8.17% | 14.87% | 21.28% | 27.54% | 32.64% | 36.53% | 39.90% | 42.83% | 45.16% | 47.49% | 49.60% | 51.48% | 52.70% | 54.30% | 55.44% |

| Tasks | N=1 | N=2 | N=3 | N=4 | N=5 | N=6 | N=7 | N=8 | N=9 | N=10 | N=11 | N=12 | N=13 | N=14 | N=15 |
|---|---|---|---|---|---|---|---|---|---|---|---|---|---|---|---|
| Social Problem Solving Task 3 | 8.19% | 14.60% | 20.46% | 26.01% | 31.25% | 36.51% | 41.80% | 46.97% | 51.86% | 55.92% | 59.77% | 63.17% | 66.51% | 69.53% | 72.08% |
| Keyword-prompted Writing | 5.94% | 11.14% | 15.64% | 19.79% | 23.66% | 27.22% | 30.71% | 33.95% | 36.82% | 39.83% | 41.92% | 44.17% | 46.27% | 48.41% | 49.98% |
| Emoji-prompted Writing Task 2 | 8.94% | 16.95% | 24.76% | 30.14% | 35.68% | 40.89% | 45.93% | 50.85% | 55.58% | 59.69% | 63.87% | 67.53% | 71.09% | 74.21% | 76.61% |
| Emoji-prompted Writing Task 3 | 8.93% | 17.53% | 25.22% | 30.97% | 36.20% | 40.24% | 43.72% | 46.59% | 49.36% | 51.74% | 53.97% | 56.08% | 58.34% | 60.42% | 62.37% |
| Creative Advert Writing | 9.90% | 18.93% | 28.86% | 37.01% | 43.90% | 49.95% | 57.11% | 62.65% | 74.40% | 82.29% | 86.97% | 90.28% | 92.47% | 94.05% | 95.22% |
| **Overall** | 7.49% | 14.03% | 20.13% | 25.52% | 30.43% | 34.82% | 39.07% | 42.71% | 46.71% | 50.03% | 52.92% | 55.44% | 57.78% | 59.89% | 61.73% |

**Qwen**

| Tasks | N=1 | N=2 | N=3 | N=4 | N=5 | N=6 | N=7 | N=8 | N=9 | N=10 | N=11 | N=12 | N=13 | N=14 | N=15 |
|---|---|---|---|---|---|---|---|---|---|---|---|---|---|---|---|
| Divergent Thinking Task 1 | 7.74% | 15.33% | 22.51% | 28.73% | 34.72% | 39.63% | 44.10% | 48.46% | 52.05% | 55.86% | 58.90% | 61.81% | 64.33% | 66.62% | 68.80% |
| Divergent Thinking Task 2 | 6.85% | 13.07% | 18.69% | 23.91% | 28.84% | 32.80% | 36.36% | 40.95% | 44.14% | 47.61% | 50.81% | 53.72% | 56.78% | 59.69% | 62.47% |
| Scientific Problem Solving Task 1 | 6.17% | 12.16% | 17.40% | 21.55% | 25.29% | 28.74% | 31.80% | 34.40% | 36.81% | 39.33% | 41.88% | 44.29% | 46.43% | 48.95% | 50.76% |
| Scientific Problem Solving Task 3 | 8.30% | 15.59% | 22.62% | 29.72% | 35.39% | 40.03% | 43.79% | 47.43% | 50.70% | 54.14% | 57.08% | 59.30% | 61.51% | 63.68% | 65.12% |
| Social Problem Solving Task 2 | 7.50% | 13.19% | 18.09% | 22.61% | 26.44% | 29.68% | 32.96% | 35.87% | 38.74% | 41.56% | 44.34% | 47.00% | 49.77% | 51.99% | 54.13% |
| Social Problem Solving Task 3 | 9.25% | 16.06% | 21.77% | 27.67% | 33.03% | 38.76% | 44.21% | 49.29% | 53.84% | 57.31% | 60.53% | 63.11% | 65.46% | 67.59% | 70.20% |
| Keyword-prompted Writing | 9.24% | 16.96% | 24.10% | 30.59% | 36.95% | 42.08% | 46.74% | 51.22% | 54.72% | 57.88% | 61.05% | 63.79% | 66.41% | 67.99% | 69.95% |
| Creative Advert Writing | 9.66% | 18.91% | 27.87% | 37.69% | 43.20% | 49.73% | 57.43% | 65.73% | 84.17% | 97.77% | 99.66% | 99.94% | 99.99% | 100.00% | 100.00% |
| **Overall** | 8.09% | 15.16% | 21.63% | 27.81% | 32.98% | 37.68% | 42.17% | 46.67% | 51.90% | 56.43% | 59.28% | 61.62% | 63.84% | 65.81% | 67.68% |

**SparkDesk**

| Tasks | N=1 | N=2 | N=3 | N=4 | N=5 | N=6 | N=7 | N=8 | N=9 | N=10 | N=11 | N=12 | N=13 | N=14 | N=15 |
|---|---|---|---|---|---|---|---|---|---|---|---|---|---|---|---|

| Task | | | | | | | | | | | | | | |
|---|---|---|---|---|---|---|---|---|---|---|---|---|---|---|
| Divergent Thinking Task 1 | 8.71% | 16.46% | 24.13% | 30.71% | 36.56% | 41.89% | 47.41% | 52.35% | 57.63% | 62.73% | 67.44% | 71.28% | 75.49% | 78.96% | 82.07% |
| Divergent Thinking Task 2 | 9.52% | 18.57% | 26.51% | 33.71% | 40.93% | 47.37% | 53.19% | 58.58% | 62.98% | 67.21% | 70.47% | 73.58% | 75.65% | 78.07% | 79.47% |
| Scientific Problem Solving Task 1 | 4.02% | 7.84% | 10.96% | 13.93% | 16.97% | 19.23% | 21.92% | 23.92% | 26.11% | 28.49% | 30.22% | 32.05% | 34.07% | 35.96% | 36.87% |
| Scientific Problem Solving Task 2 | 7.69% | 15.23% | 20.90% | 27.66% | 33.73% | 38.93% | 43.57% | 47.65% | 50.54% | 53.02% | 55.33% | 56.87% | 58.18% | 59.46% | 60.43% |
| Scientific Problem Solving Task 3 | 8.02% | 14.63% | 22.01% | 27.49% | 32.33% | 37.16% | 42.26% | 47.02% | 51.30% | 55.38% | 58.72% | 61.84% | 64.57% | 66.65% | 68.43% |
| Social Problem Solving Task 2 | 7.99% | 15.76% | 21.70% | 27.29% | 32.59% | 37.72% | 43.07% | 47.83% | 52.43% | 56.57% | 60.64% | 64.18% | 67.49% | 70.77% | 73.00% |
| Social Problem Solving Task 3 | 5.81% | 10.51% | 14.44% | 18.17% | 21.34% | 24.77% | 27.51% | 30.07% | 33.28% | 35.46% | 37.78% | 40.01% | 41.86% | 44.00% | 45.55% |
| Keyword-prompted Writing | 7.48% | 14.11% | 20.94% | 27.17% | 32.94% | 37.55% | 42.12% | 46.29% | 50.04% | 53.50% | 57.06% | 60.25% | 63.78% | 67.03% | 69.60% |
| Creative Advert Writing | 9.61% | 18.56% | 27.18% | 33.91% | 47.96% | 56.58% | 65.78% | 71.24% | 78.76% | 87.34% | 92.06% | 94.35% | 95.78% | 97.00% | 98.13% |
| **Overall** | 7.65% | 14.63% | 20.98% | 26.67% | 32.82% | 37.91% | 42.98% | 47.22% | 51.45% | 55.52% | 58.86% | 61.60% | 64.10% | 66.43% | 68.17% |

**Table S13. Collective creativity for the LLMs in each creative task and each creativity domain.**

| N | DT 1 | DT 2 | DT | SCI_PS 1 | SCI_PS 2 | SCI_PS 3 | SCI_PS | SOC_PS 2 | SOC_PS 3 | SOC_PS | PS | KW | EW_2 | EW_3 | AW | CW | Overall |
|---|---|---|---|---|---|---|---|---|---|---|---|---|---|---|---|---|---|
| **GPT-3.5** | | | | | | | | | | | | | | | | | |
| 10 | 9.369 | 8.751 | 9.035 | 11.523 | 8.362 | 6.729 | 7.893 | 42.322 | 8.857 | 13.947 | 9.368 | 6.664 | 6.006 | 6.478 | 5.570 | 6.099 | 7.707 |
| 15 | 11.470 | 11.078 | 11.302 | 13.572 | 11.632 | 6.734 | 8.894 | 56.576 | 8.945 | 21.754 | 11.473 | 7.606 | 5.541 | 7.701 | 5.129 | 6.232 | 8.706 |
| 20 | 15.587 | 12.574 | 14.042 | 17.027 | 15.895 | 6.692 | 10.612 | 60.684 | 10.713 | 34.438 | 14.671 | 9.059 | 6.284 | 8.991 | 5.263 | 6.829 | 9.967 |
| 25 | 16.232 | 13.721 | 15.116 | 20.225 | 22.427 | 6.954 | 13.171 | 74.578 | 12.119 | 29.924 | 17.337 | 9.736 | 6.482 | 9.212 | 5.527 | 7.341 | 10.908 |
| 30 | 17.052 | 19.021 | 17.930 | 20.427 | 22.627 | 7.025 | 13.003 | 72.505 | 13.658 | 38.822 | 17.910 | 10.730 | 6.368 | 10.920 | 5.704 | 7.784 | 11.758 |
| 35 | 20.471 | 18.802 | 19.620 | 23.125 | 23.962 | 6.915 | 15.156 | 79.517 | 18.077 | 55.294 | 21.940 | 11.364 | 7.344 | 13.627 | 5.854 | 8.431 | 12.969 |
| **GPT-4** | | | | | | | | | | | | | | | | | |
| 10 | 13.106 | 27.368 | 18.415 | 44.506 | 11.352 | 5.931 | 11.465 | 17.736 | 14.442 | 16.229 | 13.046 | 32.883 | 6.044 | 7.002 | 6.456 | 7.562 | 10.419 |
| 15 | 17.415 | 37.342 | 26.832 | 71.200 | 13.157 | 6.873 | 14.142 | 21.791 | 33.454 | 25.121 | 17.703 | 55.519 | 6.236 | 10.002 | 5.814 | 8.352 | 13.495 |
| 20 | 31.579 | 39.262 | 36.222 | 81.354 | 16.634 | 7.316 | 18.402 | 25.675 | 40.677 | 30.293 | 22.626 | 105.942 | 6.046 | 11.735 | 6.585 | 9.282 | 16.542 |
| 25 | 30.711 | 45.524 | 39.866 | 83.173 | 17.074 | 7.985 | 18.445 | 30.377 | 59.614 | 44.416 | 26.660 | 107.899 | 6.470 | 11.763 | 6.603 | 9.828 | 18.732 |
| 30 | 39.709 | 47.336 | 44.238 | 96.986 | 27.493 | 8.712 | 21.715 | 40.456 | 86.858 | 68.001 | 34.792 | 155.212 | 6.599 | 11.949 | 6.534 | 10.091 | 22.174 |
| 35 | 41.009 | 46.663 | 44.726 | 122.870 | 34.177 | 8.901 | 25.048 | 56.250 | 82.466 | 70.448 | 40.365 | 171.078 | 6.844 | 13.415 | 6.638 | 10.436 | 24.511 |
| **Claude** | | | | | | | | | | | | | | | | | |
| 10 | 9.392 | 11.577 | 10.272 | 29.301 | 15.370 | 10.010 | 16.484 | 11.212 | 8.620 | 9.466 | 12.812 | 15.012 | 7.826 | 9.268 | 6.007 | 8.269 | 9.992 |
| **Qwen** | | | | | | | | | | | | | | | | | |
| 10 | 8.429 | 10.747 | 9.523 | 14.581 | 8.786 | 11.225 | 13.102 | 8.157 | 10.189 | 10.667 | 7.727 | 6.035 | 6.663 | 8.638 | 6.007 | 8.269 | 9.992 |
| **SparkDesk** | | | | | | | | | | | | | | | | | |
| 10 | 7.524 | 6.451 | 6.947 | 24.510 | 8.814 | 8.696 | 11.883 | 8.472 | 17.628 | 11.275 | 11.596 | 8.990 | 5.237 | 6.426 | 8.657 | 8.269 | 9.992 |

| LLMs | | | | | | | | | | | | | | | | | |
|---|---|---|---|---|---|---|---|---|---|---|---|---|---|---|---|---|---|
| 10 | 9.130 | 10.872 | 9.870 | 23.460 | 10.321 | 7.189 | 11.478 | 14.019 | 9.906 | 11.514 | 11.492 | 10.222 | 5.965 | 7.074 | 5.513 | 6.684 | 8.716 |
| 15 | 10.788 | 13.919 | 12.144 | 32.995 | 16.008 | 8.998 | 16.207 | 21.201 | 13.923 | 16.921 | 16.515 | 14.602 | 6.539 | 9.964 | 6.076 | 7.906 | 11.087 |
| 20 | 13.028 | 17.824 | 15.158 | 48.982 | 23.195 | 11.732 | 22.115 | 29.047 | 18.629 | 23.611 | 22.688 | 21.471 | 7.250 | 11.481 | 6.449 | 8.769 | 13.786 |
| 25 | 14.927 | 21.479 | 17.866 | 68.809 | 34.184 | 13.861 | 28.888 | 37.448 | 22.746 | 30.037 | 29.330 | 29.705 | 7.828 | 13.038 | 6.662 | 9.533 | 16.312 |
| 30 | 17.240 | 24.672 | 20.609 | 76.549 | 47.239 | 16.056 | 36.726 | 44.269 | 30.479 | 37.254 | 36.956 | 36.919 | 8.303 | 14.109 | 6.495 | 10.310 | 18.983 |
| 35 | 19.370 | 29.642 | 24.104 | 82.718 | 58.137 | 17.001 | 42.881 | 52.714 | 41.186 | 47.450 | 44.715 | 43.957 | 8.726 | 14.930 | 6.543 | 10.869 | 22.047 |
| 40 | 21.360 | 30.348 | 25.743 | 85.764 | 61.783 | 17.938 | 45.895 | 59.758 | 47.518 | 54.518 | 49.414 | 53.757 | 9.340 | 16.258 | 6.507 | 11.693 | 24.279 |
| 45 | 24.541 | 32.489 | 28.609 | 91.983 | 71.315 | 19.978 | 51.618 | 61.931 | 58.569 | 60.415 | 55.467 | 68.716 | 9.616 | 17.278 | 6.506 | 12.157 | 27.035 |
| 50 | 26.452 | 35.205 | 31.031 | 94.297 | 75.588 | 21.143 | 55.656 | 65.714 | 62.960 | 64.620 | 59.556 | 76.835 | 9.784 | 17.782 | 6.645 | 12.564 | 29.154 |

*Note.* DT 1: Divergent thinking task 1, DT 2: Divergent thinking task 2; SCI_PS 1: Scientific problem solving task 1; SCI_PS 2: Scientific problem solving task 2; SCI_PS 3: Scientific problem solving task 3; SOC_PS 2: Social problem solving task 2; SOC_PS 3: Social problem solving task 3; KW: Keyword-prompted writing; EW 2: Emoji-prompted writing 2; EW 3: Emoji-prompted writing 3; AW: Creative advert writing; DT: Divergent thinking; SCI_PS: Scientific problem solving; SOC_PS: Social problem solving; PS: Problem solving; CW: Creative writing.

**Table S14. Descriptive statistics and independent sample *t*-test* (two-sided) results of age, performances on the Creative Advert Writing Task and Remote Association Task creativity performance between the human participants assigned to Test Form A and Test Form B.**

| Tasks | Group A | | | Group B | | | *t*- test results | | | | |
|---|---|---|---|---|---|---|---|---|---|---|---|
| | *n* | *M* | *SD* | *n* | *M* | *SD* | 95% CI (Δ) | *t* | *df* | *P* | Cohen's *d* |
| **Remote Association Task** | 237 | 8.987 | 3.068 | 230 | 8.839 | 3.316 | [-0.434, 0.730] | 0.500 | 465 | 0.617 | 0.046 |
| **Creative Advert Writing** | 213 | 2.275 | 0.629 | 205 | 2.305 | 0.591 | [-0.148, 0.087] | -0.505 | 416 | 0.614 | -0.050 |
| **Age** | 237 | 27.224 | 6.070 | 230 | 27.913 | 5.963 | [-1.786, 0.407] | -1.235 | 465 | 0.217 | -0.115 |

*Note*. *Before running the *t*-tests, we tested and confirmed that both groups had equal variances on all indicators.

**Table S15. Results of Chi-square tests on gender, work experience, education level and areas of study differences between human participants assigned to Test Form A and Test Form B.**

| Variable | *Chi-Square* | *P* | *df* | *Cramér's V* |
|---|---|---|---|---|
| **Gender** | 1.044 | 0.307 | 1 | 0.047 |
| **Work experience** | 0.413 | 0.813 | 2 | 0.030 |
| **Education level** | 1.523 | 0.217 | 1 | 0.057 |
| **Areas of study** | 7.231 | 0.065 | 3 | 0.124 |

**Table S16. Descriptive statistics and independent sample *t*-test* (two-sided) results comparing the responses of GPT-3.5 in certain creative tasks with and without the additional instructions.**

| Tasks | Prompt v2 | | | Prompt v1 | | | *t*- test results | | | | Cohen's *d* |
|---|---|---|---|---|---|---|---|---|---|---|---|
| | *n* | *M* | *SD* | *n* | *M* | *SD* | 95% CI (Δ) | *t* | *df* | *P* | |
| **Divergent Thinking** | | | | | | | | | | | |
| Task 1 Creativity Mean | 50 | 4.738 | 0.49 | 50 | 4.547 | 0.401 | [0.012, 0.371] | 2.119 | 98 | 0.037 | 0.428 |
| Task 1 Creativity Max | 50 | 6.746 | 1.048 | 50 | 6.389 | 0.817 | [-0.020, 0.734] | 1.88 | 98 | 0.063 | 0.38 |
| Task 2 Creativity Mean | 50 | 4.819 | 0.377 | 50 | 4.73 | 0.322 | [-0.052, 0.229] | 1.247 | 98 | 0.215 | 0.252 |
| Task 2 Creativity Max | 50 | 6.262 | 0.894 | 50 | 6.515 | 0.894 | [-0.612, 0.105] | -1.402 | 98 | 0.164 | -0.283 |
| **Scientific Problem Solving** | | | | | | | | | | | |
| Task1 Creativity | 51 | 1.834 | 0.36 | 51 | 1.733 | 0.278 | [-0.027, 0.228] | 1.56 | 100 | 0.12 | 0.313 |
| Task2 Creativity | 51 | 1.702 | 0.6 | 51 | 1.714 | 0.659 | [-0.262, 0.238] | -0.093 | 100 | 0.926 | -0.019 |
| Task3 Creativity | 51 | 1.353 | 0.393 | 51 | 1.337 | 0.393 | [-0.140, 0.172] | 0.199 | 100 | 0.842 | 0.04 |
| **Social Problem Solving** | | | | | | | | | | | |
| Table 1 Flexibility | 51 | 2.157 | 0.606 | 51 | 1.961 | 0.593 | [-0.042, 0.434] | 1.635 | 100 | 0.105 | 0.327 |
| Task 2 Creativity | 51 | 2.729 | 0.579 | 51 | 2.698 | 0.719 | [-0.228, 0.290] | 0.24 | 100 | 0.811 | 0.048 |
| Task 3 Creativity | 51 | 2.267 | 0.476 | 51 | 2.38 | 0.539 | [-0.316, 0.088] | -1.118 | 100 | 0.266 | -0.224 |
| **Creative Advert Writing** | | | | | | | | | | | |
| Creativity | 41 | 1.298 | 0.373 | 41 | 1.361 | 0.401 | [-0.236, 0.109] | -0.733 | 80 | 0.466 | -0.164 |

*Note.* *Before running the *t*-tests, we tested and confirmed that both groups had equal variances on all indicators.

**Table S17. Descriptive statistics and independent sample *t*-test* (two-sided) results of gender differences in humans' creativity performance.**

| Tasks | Female | | | Male | | | *t*- test results | | | | |
|---|---|---|---|---|---|---|---|---|---|---|---|
| | *n* | *M* | *SD* | *n* | *M* | *SD* | 95% CI (Δ) | *t* | *df* | *P* | Cohen's *d* |
| **Divergent Thinking** | | | | | | | | | | | |
| Task 1 Creativity Mean | 157 | 4.723 | 1.289 | 73 | 4.87 | 1.369 | [-0.515, 0.222] | -0.782 | 228 | 0.435 | -0.111 |
| Task 1 Creativity Max | 157 | 6.398 | 1.879 | 73 | 6.699 | 1.955 | [-0.834, 0.233] | -1.109 | 228 | 0.269 | -0.158 |
| **Scientific Problem Solving** | | | | | | | | | | | |
| Task 2 Creativity Mean | 171 | 5.073 | 1.362 | 64 | 4.966 | 0.857 | [-0.254, 0.468] | 0.584 | 233 | 0.56 | 0.086 |
| Task 2 Creativity Max | 171 | 6.745 | 2.106 | 64 | 6.515 | 1.637 | [-0.347, 0.807] | 0.785 | 233 | 0.433 | 0.116 |
| **Scientific Problem Solving** | | | | | | | | | | | |
| Task1 Creativity | 173 | 1.912 | 0.363 | 64 | 2.043 | 0.387 | [-0.237, -0.024] | -2.41 | 235 | 0.017 | -0.353 |
| Task2 Creativity | 172 | 2.381 | 0.803 | 64 | 2.3 | 0.736 | [-0.145, 0.308] | 0.707 | 234 | 0.48 | 0.104 |
| Task3 Creativity | 173 | 2.602 | 0.877 | 64 | 2.75 | 0.738 | [-0.390, 0.095] | -1.199 | 235 | 0.232 | -0.175 |
| **Social Problem Solving** | | | | | | | | | | | |
| Task 1 Flexibility | 156 | 2.321 | 0.601 | 73 | 2.274 | 0.534 | [-0.116, 0.209] | 0.565 | 227 | 0.572 | 0.08 |
| Task 2 Creativity | 156 | 2.051 | 0.706 | 73 | 2.258 | 0.894 | [-0.422, 0.009] | -1.888 | 227 | 0.06 | -0.268 |
| Task 3 Creativity | 154 | 2.753 | 0.855 | 72 | 2.561 | 0.773 | [-0.041, 0.426] | 1.621 | 224 | 0.106 | 0.231 |
| **Keyword-prompted Writing** | | | | | | | | | | | |
| Creativity | 150 | 2.382 | 0.662 | 71 | 2.49 | 0.71 | [-0.301, 0.085] | -1.103 | 219 | 0.271 | -0.16 |
| Diversity | 119 | 3.276 | 0.666 | 59 | 3.353 | 0.693 | [-0.290, 0.136] | -0.712 | 176 | 0.478 | -0.114 |
| **Emoji-prompted Writing** | | | | | | | | | | | |
| Task 1 Diversity | 147 | 3.061 | 0.697 | 54 | 3.137 | 0.786 | [-0.303, 0.152] | -0.657 | 199 | 0.512 | -0.105 |
| Task 2 Creativity | 133 | 2.54 | 0.776 | 48 | 2.775 | 0.743 | [-0.492, 0.021] | -1.81 | 179 | 0.072 | -0.306 |
| Task 3 Creativity | 138 | 2.472 | 0.687 | 51 | 2.784 | 0.643 | [-0.531, -0.092] | -2.801 | 187 | 0.006 | -0.461 |
| **Creative Advert Writing** | | | | | | | | | | | |
| Creativity | 295 | 2.285 | 0.618 | 123 | 2.301 | 0.599 | [-0.145, 0.114] | -0.234 | 416 | 0.815 | -0.025 |

*Note*. *Before running the *t*-tests, we tested and confirmed that both groups had equal variances on all indicators.

**Table S18. Descriptive statistics and one-way analysis of variance (ANOVA) of humans' creativity performance among different age groups.**

| Tasks | Statistics | Age groups | | | ANOVA Results | | |
|---|---|---|---|---|---|---|---|
| **Divergent Thinking** | | | | | | | |
| | Statistics | ~23 | 24~29 | 30~ | $F(2, 227)$ | $P$ | Partial $\eta^2$ |
| **Task 1 Creativity Mean** | $M$ | 4.789 | 4.796 | 4.719 | 0.076 | 0.926 | 0.001 |
| | $SD$ | 1.141 | 1.465 | 1.302 | | | |
| | $n$ | 68 | 90 | 72 | | | |
| **Task 2 Creativity Max** | $M$ | 6.457 | 6.462 | 6.569 | 0.08 | 0.923 | 0.001 |
| | $SD$ | 1.787 | 1.972 | 1.975 | | | |
| | $n$ | 68 | 90 | 72 | | | |
| | Statistics | ~23 | 24~29 | 30~ | $F(2, 232)$ | $P$ | Partial $\eta^2$ |
| **Task 2 Creativity Mean** | $M$ | 5.015 | 5.164 | 4.903 | 0.811 | 0.446 | 0.007 |
| | $SD$ | 0.859 | 1.455 | 1.384 | | | |
| | $n$ | 86 | 90 | 59 | | | |
| **Task 2 Creativity Max** | $M$ | 6.669 | 6.742 | 6.611 | 0.079 | 0.924 | 0.001 |
| | $SD$ | 1.696 | 2.066 | 2.302 | | | |
| | $n$ | 86 | 90 | 59 | | | |
| **Scientific Problem Solving** | | | | | | | |
| **Task 1 Creativity** | Statistics | ~23 | 24~29 | 30~ | $F(2, 234)$ | $P$ | Partial $\eta^2$ |
| | $M$ | 1.916 | 1.87 | 1.909 | 0.408 | 0.665 | 0.003 |
| | $SD$ | 0.389 | 0.361 | 0.309 | | | |
| | $n$ | 87 | 90 | 60 | | | |
| **Task 2 Creativity** | Statistics | ~23 | 24~29 | 30~ | $F(2, 233)$ | $P$ | Partial $\eta^2$ |
| | $M$ | 2.402 | 2.351 | 2.31 | 0.253 | 0.777 | 0.002 |
| | $SD$ | 0.752 | 0.816 | 0.794 | | | |
| | $n$ | 87 | 89 | 60 | | | |
| **Task 3 Creativity** | Statistics | ~23 | 24~29 | 30~ | $F(2, 234)$ | $P$ | Partial $\eta^2$ |
| | $M$ | 2.497 | 2.816 | 2.593 | 3.373 | 0.036 | 0.028 |
| | $SD$ | 0.785 | 0.869 | 0.85 | | | |
| | $n$ | 87 | 90 | 60 | | | |
| **Social Problem Solving** | | | | | | | |
| **Task 1 Flexibility** | Statistics | ~23 | 24~29 | 30~ | $F(2, 226)$ | $P$ | Partial $\eta^2$ |
| | $M$ | 2.373 | 2.367 | 2.167 | 3.078 | 0.048 | 0.027 |
| | $SD$ | 0.599 | 0.507 | 0.628 | | | |
| | $n$ | 67 | 90 | 72 | | | |

| | Statistics | ~23 | 24~29 | 30~ | F(2, 226) | P | Partial η² |
|---|---|---|---|---|---|---|---|
| Task 2 Creativity | | | | | | | |
| | M | 2.018 | 2.216 | 2.086 | 1.337 | 0.265 | 0.012 |
| | SD | 0.778 | 0.832 | 0.689 | | | |
| | n | 67 | 90 | 72 | | | |
| Task 3 Creativity | Statistics | ~23 | 24~29 | 30~ | F(2, 223) | P | Partial η² |
| | M | 2.621 | 2.66 | 2.8 | 0.892 | 0.411 | 0.008 |
| | SD | 0.803 | 0.863 | 0.823 | | | |
| | n | 66 | 90 | 70 | | | |
| **Keyword-prompted Writing** | | | | | | | |
| Creativity | Statistics | ~23 | 24~29 | 30~ | F(2, 218) | P | Partial η² |
| | M | 2.463 | 2.429 | 2.357 | 0.435 | 0.648 | 0.004 |
| | SD | 0.691 | 0.667 | 0.693 | | | |
| | n | 67 | 85 | 69 | | | |
| Diversity | Statistics | ~23 | 24~29 | 30~ | F(2, 175) | P | Partial η² |
| | M | 3.368 | 3.254 | 3.302 | 0.416 | 0.66 | 0.005 |
| | SD | 0.661 | 0.754 | 0.593 | | | |
| | n | 50 | 71 | 57 | | | |
| **Emoji-prompted Writing** | | | | | | | |
| Task 1 Diversity | Statistics | ~23 | 24~29 | 30~ | F(2, 198) | P | Partial η² |
| | M | 3.083 | 3.104 | 3.045 | 0.098 | 0.907 | 0.001 |
| | SD | 0.796 | 0.681 | 0.685 | | | |
| | n | 77 | 75 | 49 | | | |
| Task 2 Creativity | Statistics | ~23 | 24~29 | 30~ | F(2, 178) | P | Partial η² |
| | M | 2.809 | 2.504 | 2.458 | 3.765 | 0.025 | 0.041 |
| | SD | 0.776 | 0.777 | 0.728 | | | |
| | n | 65 | 71 | 45 | | | |
| Task 3 Creativity | Statistics | ~23 | 24~29 | 30~ | F(2, 186) | P | Partial η² |
| | M | 2.63 | 2.494 | 2.533 | 0.721 | 0.488 | 0.008 |
| | SD | 0.733 | 0.675 | 0.651 | | | |
| | n | 74 | 70 | 45 | | | |
| **Creative Advert Writing** | | | | | | | |
| Creativity | Statistics | ~23 | 24~29 | 30~ | F(2,415) | P | Partial η² |
| | M | 2.312 | 2.341 | 2.193 | 2.132 | 0.12 | 0.01 |
| | SD | 0.613 | 0.597 | 0.624 | | | |
| | n | 141 | 160 | 117 | | | |

**Table S19. Post-hoc multiple comparison results (Bonferroni correction) following one-way analysis of variance of humans' creativity performance among different age groups.**

|  | *Age_group (I)* | *Age_group (J)* | *Mean Difference (I-J)* | *Std. Error* | *P* | *95% CI (Δ)* |
|---|---|---|---|---|---|---|
| **Scientific Problem Solving** | | | | | | |
| Task 3 Creativity | ~23 | 24~29 | -0.319 | 0.125 | 0.035 | [-0.627, -0.017] |
|  | ~23 | 30~ | -0.097 | 0.140 | 1.000 | [-0.434, 0.241] |
|  | 24~29 | 30~ | 0.222 | 0.139 | 0.334 | [-0.113, 0.557] |
| **Social Problem Solving** | | | | | | |
| Task 1 Flexibility | ~23 | 24~29 | 0.016 | 0.092 | 1.000 | [-0.206, 0.238] |
|  | ~23 | 30~ | 0.227 | 0.097 | 0.061 | [-0.007, 0.462] |
|  | 24~29 | 30~ | 0.212 | 0.091 | 0.063 | [-0.008, 0.431] |
| **Emoji-prompted Writing** | | | | | | |
| Task 2 Creativity | ~23 | 24~29 | 0.305 | 0.131 | 0.064 | [-0.012, 0.622] |
|  | ~23 | 30~ | 0.351 | 0.148 | 0.057 | [-0.007, 0.710] |
|  | 24~29 | 30~ | 0.046 | 0.146 | 1.000 | [-0.306, 0.399] |

**Table S20. Descriptive statistics and independent sample *t*-test* (two-sided) results of creativity performance between human participants without (current students) and with work experience (current employees).**

| Tasks | Current graduates | | | Current employees | | | *t*- test results | | | | Cohen's *d* |
|---|---|---|---|---|---|---|---|---|---|---|---|
| | *n* | *M* | *SD* | *n* | *M* | *SD* | 95% CI (Δ) | *t* | *df* | *P* | |
| **Divergent Thinking** | | | | | | | | | | | |
| Task 1 Creativity Mean | 49 | 4.782 | 1.211 | 72 | 4.834 | 1.445 | [-0.552, 0.450] | -0.202 | 119 | 0.84 | -0.038 |
| Task 1 Creativity Max | 49 | 6.614 | 1.921 | 72 | 6.572 | 2.006 | [-0.688, 0.771] | 0.113 | 119 | 0.91 | 0.021 |
| Task 2 Creativity Mean | 53 | 5.068 | 0.847 | 68 | 5.089 | 1.359 | [-0.447, 0.404] | -0.099 | 119 | 0.921 | -0.018 |
| Task 2 Creativity Max | 53 | 6.846 | 1.752 | 68 | 6.704 | 2.178 | [-0.591, 0.875] | 0.383 | 119 | 0.702 | 0.071 |
| **Scientific Problem Solving** | | | | | | | | | | | |
| Task1 Creativity | 54 | 1.932 | 0.395 | 68 | 1.918 | 0.347 | [-0.119, 0.147] | 0.207 | 120 | 0.836 | 0.038 |
| Task2 Creativity | 54 | 2.385 | 0.729 | 67 | 2.412 | 0.862 | [-0.318, 0.265] | -0.182 | 119 | 0.856 | -0.033 |
| Task3 Creativity | 54 | 2.481 | 0.777 | 68 | 2.782 | 0.792 | [-0.584, 0.017] | -2.102 | 120 | 0.038 | -0.383 |
| **Social Problem Solving** | | | | | | | | | | | |
| Task 1 Flexibility | 49 | 2.367 | 0.566 | 71 | 2.211 | 0.583 | [-0.056, 0.368] | 1.458 | 118 | 0.148 | 0.271 |
| Task 2 Creativity | 49 | 2.049 | 0.846 | 71 | 2.203 | 0.787 | [-0.452, 0.145] | -1.021 | 118 | 0.309 | -0.19 |
| Task 3 Creativity | 48 | 2.596 | 0.814 | 70 | 2.749 | 0.795 | [-0.451, 0.145] | -1.015 | 116 | 0.312 | -0.19 |
| **Keyword-prompted Writing** | | | | | | | | | | | |
| Creativity | 48 | 2.406 | 0.65 | 68 | 2.365 | 0.685 | [-0.211, 0.294] | 0.326 | 114 | 0.745 | 0.062 |
| Diversity | 37 | 3.232 | 0.585 | 55 | 3.211 | 0.576 | [-0.226, 0.269] | 0.173 | 90 | 0.863 | 0.037 |
| **Emoji-prompted Writing** | | | | | | | | | | | |
| Task 1 Diversity | 46 | 3.1 | 0.8 | 56 | 3.1 | 0.643 | [-0.286, 0.286] | 0 | 100 | 1 | 0 |
| Task 2 Creativity | 45 | 2.836 | 0.803 | 54 | 2.481 | 0.798 | [0.030, 0.678] | 2.171 | 97 | 0.032 | 0.443 |
| Task 3 Creativity | 49 | 2.629 | 0.786 | 45 | 2.524 | 0.654 | [-0.197, 0.405] | 0.688 | 92 | 0.493 | 0.144 |
| **Creative Advert Writing** | | | | | | | | | | | |
| Creativity | 95 | 2.406 | 0.625 | 124 | 2.248 | 0.609 | [-0.008, 0.323] | 1.881 | 217 | 0.061 | 0.256 |

*Note.* *Before running the *t*-tests, we tested and confirmed that both groups had equal variances on all indicators.

**Table S21. Descriptive statistics and independent sample *t*-test* (two-sided) results of creativity performance between human participants with a Bachelor's degree and those with a Master's degree (when applying).**

| Tasks | Bachelor degree | | | Master degree | | | *t*- test results | | | | Cohen's d |
|---|---|---|---|---|---|---|---|---|---|---|---|
| | *n* | *M* | *SD* | *n* | *M* | *SD* | 95% CI (Δ) | *t* | *df* | *P* | |
| **Divergent Thinking** | | | | | | | | | | | |
| Task 1 Creativity Mean | 212 | 4.771 | 1.334 | 18 | 4.75 | 1.092 | [-0.618, 0.662] | 0.067 | 228 | 0.946 | 0.017 |
| Task 1 Creativity Max | 212 | 6.476 | 1.927 | 18 | 6.701 | 1.659 | [-1.152, 0.701] | -0.479 | 228 | 0.632 | -0.118 |
| Task 2 Creativity Mean | 224 | 5.064 | 1.265 | 11 | 4.634 | 0.654 | [-0.329, 1.190] | 1.117 | 233 | 0.265 | 0.346 |
| Task 2 Creativity Max | 224 | 6.687 | 2.015 | 11 | 6.578 | 1.451 | [-1.108, 1.326] | 0.177 | 233 | 0.86 | 0.055 |
| **Scientific Problem Solving** | | | | | | | | | | | |
| Task1 Creativity | 226 | 1.95 | 0.379 | 11 | 1.903 | 0.245 | [-0.181, 0.274] | 0.404 | 235 | 0.687 | 0.125 |
| Task2 Creativity | 225 | 2.37 | 0.787 | 11 | 2.145 | 0.738 | [-0.253, 0.702] | 0.925 | 234 | 0.356 | 0.286 |
| Task3 Creativity | 226 | 2.636 | 0.839 | 11 | 2.764 | 0.937 | [-0.641, 0.386] | -0.489 | 235 | 0.625 | -0.151 |
| **Social Problem Solving** | | | | | | | | | | | |
| Table 1 Flexibility | 211 | 2.332 | 0.572 | 18 | 2 | 0.594 | [0.054, 0.609] | 2.354 | 227 | 0.019 | 0.578 |
| Task 2 Creativity | 211 | 2.122 | 0.79 | 18 | 2.056 | 0.577 | [-0.309, 0.442] | 0.35 | 227 | 0.727 | 0.086 |
| Task 3 Creativity | 208 | 2.672 | 0.823 | 18 | 2.922 | 0.933 | [-0.653, 0.153] | -1.223 | 224 | 0.223 | -0.301 |
| **Keyword-prompted Writing** | | | | | | | | | | | |
| Creativity | 203 | 2.415 | 0.672 | 18 | 2.433 | 0.759 | [-0.349, 0.313] | -0.108 | 219 | 0.914 | -0.027 |
| Diversity | 163 | 3.31 | 0.681 | 15 | 3.2 | 0.607 | [-0.251, 0.472] | 0.603 | 176 | 0.547 | 0.163 |
| **Emoji-prompted Writing** | | | | | | | | | | | |
| Task 1 Diversity | 192 | 3.091 | 0.727 | 9 | 2.889 | 0.597 | [-0.286, 0.689] | 0.816 | 199 | 0.416 | 0.279 |
| Task 2 Creativity | 171 | 2.602 | 0.766 | 10 | 2.6 | 0.908 | [-0.498, 0.502] | 0.009 | 179 | 0.993 | 0.003 |
| Task 3 Creativity | 179 | 2.553 | 0.701 | 10 | 2.62 | 0.433 | [-0.511, 0.377] | -0.297 | 187 | 0.767 | -0.097 |
| **Creative Advert Writing** | | | | | | | | | | | |
| Creativity | 392 | 2.299 | 0.617 | 26 | 2.154 | 0.51 | [-0.098, 0.388] | 1.172 | 416 | 0.242 | 0.237 |

*Note.* *Before running the *t*-tests, we tested and confirmed that both groups had equal variances on all indicators.

**Table S22. Descriptive statistics and one-way analysis of variance (ANOVA) of humans' creativity performance across different areas of studies.**

| Tasks | Statistics | Academic disciplines | | | | ANOVA Results | | |
|---|---|---|---|---|---|---|---|---|
| **Divergent Thinking** | | | | | | | | |
| | Statistics | Arts and humanities | STEM | Programme subject | Other social science subjects | $F(3,226)$ | $P$ | Partial $\eta^2$ |
| Task 1 Creativity Mean | M | 4.417 | 4.883 | 4.687 | 4.83 | 0.896 | 0.444 | 0.012 |
| | SD | 1.112 | 1.479 | 1.241 | 1.273 | | | |
| | n | 25 | 74 | 52 | 79 | | | |
| Task 1 Creativity Max | M | 5.991 | 6.602 | 6.482 | 6.559 | 0.684 | 0.563 | 0.009 |
| | SD | 1.736 | 1.932 | 2.062 | 1.856 | | | |
| | n | 25 | 74 | 52 | 79 | | | |
| | Statistics | Arts and humanities | STEM | Programme subject | Other social science subjects | $F(3,231)$ | $P$ | Partial $\eta^2$ |
| Task 2 Creativity Mean | M | 5.02 | 4.94 | 5.041 | 5.13 | 0.267 | 0.849 | 0.003 |
| | SD | 0.9 | 1.089 | 1.001 | 1.609 | | | |
| | n | 44 | 57 | 51 | 83 | | | |
| Task 2 Creativity Max | M | 6.643 | 6.556 | 6.66 | 6.803 | 0.184 | 0.907 | 0.002 |
| | SD | 1.834 | 1.975 | 1.662 | 2.288 | | | |
| | n | 44 | 57 | 51 | 83 | | | |
| **Scientific Problem Solving** | | | | | | | | |
| Task 1 Creativity | Statistics | Arts and humanities | STEM | Programme subject | Other social science subjects | $F(3,233)$ | $P$ | Partial $\eta^2$ |
| | M | 1.895 | 1.942 | 1.992 | 1.952 | 0.535 | 0.659 | 0.007 |
| | SD | 0.369 | 0.359 | 0.355 | 0.399 | | | |
| | n | 44 | 58 | 51 | 84 | | | |
| Task 2 Creativity | Statistics | Arts and humanities | STEM | Programme subject | Other social science subjects | $F(3,232)$ | $P$ | Partial $\eta^2$ |
| | M | 2.309 | 2.345 | 2.392 | 2.376 | 0.107 | 0.956 | 0.001 |
| | SD | 0.648 | 0.893 | 0.695 | 0.832 | | | |
| | n | 44 | 58 | 50 | 84 | | | |
| Task 3 Creativity | Statistics | Arts and humanities | STEM | Programme subject | Other social science subjects | $F(3,233)$ | $P$ | Partial $\eta^2$ |
| | M | 2.741 | 2.634 | 2.706 | 2.557 | 0.582 | 0.627 | 0.007 |
| | SD | 0.869 | 0.781 | 0.787 | 0.906 | | | |
| | n | 44 | 58 | 51 | 84 | | | |

| Social Problem Solving | | | | | | | | |
|---|---|---|---|---|---|---|---|---|
| Task 1 Flexibility | Statistics | Arts and humanities | STEM | Programme subject | Other social science subjects | $F(3,225)$ | $P$ | Partial $\eta^2$ |
| | $M$ | 2.28 | 2.257 | 2.346 | 2.333 | 0.333 | 0.802 | 0.004 |
| | $SD$ | 0.614 | 0.598 | 0.59 | 0.55 | | | |
| | $n$ | 25 | 74 | 52 | 78 | | | |
| Task 2 Creativity | Statistics | Arts and humanities | STEM | Programme subject | Other social science subjects | $F(3,225)$ | $P$ | Partial $\eta^2$ |
| | $M$ | 1.992 | 2.081 | 2.077 | 2.218 | 0.755 | 0.52 | 0.01 |
| | $SD$ | 0.692 | 0.794 | 0.703 | 0.828 | | | |
| | $n$ | 25 | 74 | 52 | 78 | | | |
| Task 3 Creativity | Statistics | Arts and humanities | STEM | Programme subject | Other social science subjects | $F(3,222)$ | $P$ | Partial $\eta^2$ |
| | $M$ | 2.904 | 2.676 | 2.529 | 2.747 | 1.314 | 0.271 | 0.017 |
| | $SD$ | 0.837 | 0.813 | 0.891 | 0.806 | | | |
| | $n$ | 25 | 74 | 51 | 76 | | | |

| Keyword-prompted Writing | | | | | | | | |
|---|---|---|---|---|---|---|---|---|
| Creativity | Statistics | Arts and humanities | STEM | Programme subject | Other social science subjects | $F(3,217)$ | $P$ | Partial $\eta^2$ |
| | $M$ | 2.467 | 2.361 | 2.447 | 2.432 | 0.24 | 0.868 | 0.003 |
| | $SD$ | 0.789 | 0.669 | 0.644 | 0.69 | | | |
| | $n$ | 24 | 70 | 51 | 76 | | | |
| Diversity | Statistics | Arts and humanities | STEM | Programme subject | Other social science subjects | $F(3,174)$ | $P$ | Partial $\eta^2$ |
| | $M$ | 3.27 | 3.224 | 3.346 | 3.356 | 0.449 | 0.718 | 0.008 |
| | $SD$ | 0.757 | 0.728 | 0.687 | 0.597 | | | |
| | $n$ | 20 | 58 | 41 | 59 | | | |

| Emoji-prompted Writing | | | | | | | | |
|---|---|---|---|---|---|---|---|---|
| Task 1 Diversity | Statistics | Arts and humanities | STEM | Programme subject | Other social science subjects | $F(3,197)$ | $P$ | Partial $\eta^2$ |
| | $M$ | 3.04 | 3.109 | 3.173 | 3.031 | 0.421 | 0.738 | 0.006 |
| | $SD$ | 0.702 | 0.724 | 0.721 | 0.748 | | | |
| | $n$ | 40 | 44 | 45 | 72 | | | |
| Task 2 Creativity | Statistics | Arts and humanities | STEM | Programme subject | Other social science subjects | $F(3,177)$ | $P$ | Partial $\eta^2$ |
| | $M$ | 2.565 | 2.7 | 2.62 | 2.531 | 0.445 | 0.721 | 0.007 |
| | $SD$ | 0.799 | 0.744 | 0.845 | 0.751 | | | |
| | $n$ | 34 | 48 | 41 | 58 | | | |
| Task 3 Creativity | Statistics | Arts and humanities | STEM | Programme subject | Other social science subjects | $F(3,185)$ | $P$ | Partial $\eta^2$ |

|  | M | 2.572 | 2.612 | 2.537 | 2.516 | 0.197 | 0.898 | 0.003 |
|  | SD | 0.725 | 0.682 | 0.754 | 0.654 |  |  |  |
|  | n | 36 | 51 | 38 | 64 |  |  |  |

**Creative Advert Writing**

| Creativity | Statistics | Arts and humanities | STEM | Programme subject | Other social science subjects | $F(3,414)$ | $P$ | Partial $\eta^2$ |
|---|---|---|---|---|---|---|---|---|
|  | M | 2.294 | 2.282 | 2.358 | 2.249 | 0.62 | 0.602 | 0.004 |
|  | SD | 0.572 | 0.595 | 0.673 | 0.601 |  |  |  |
|  | n | 62 | 120 | 96 | 140 |  |  |  |

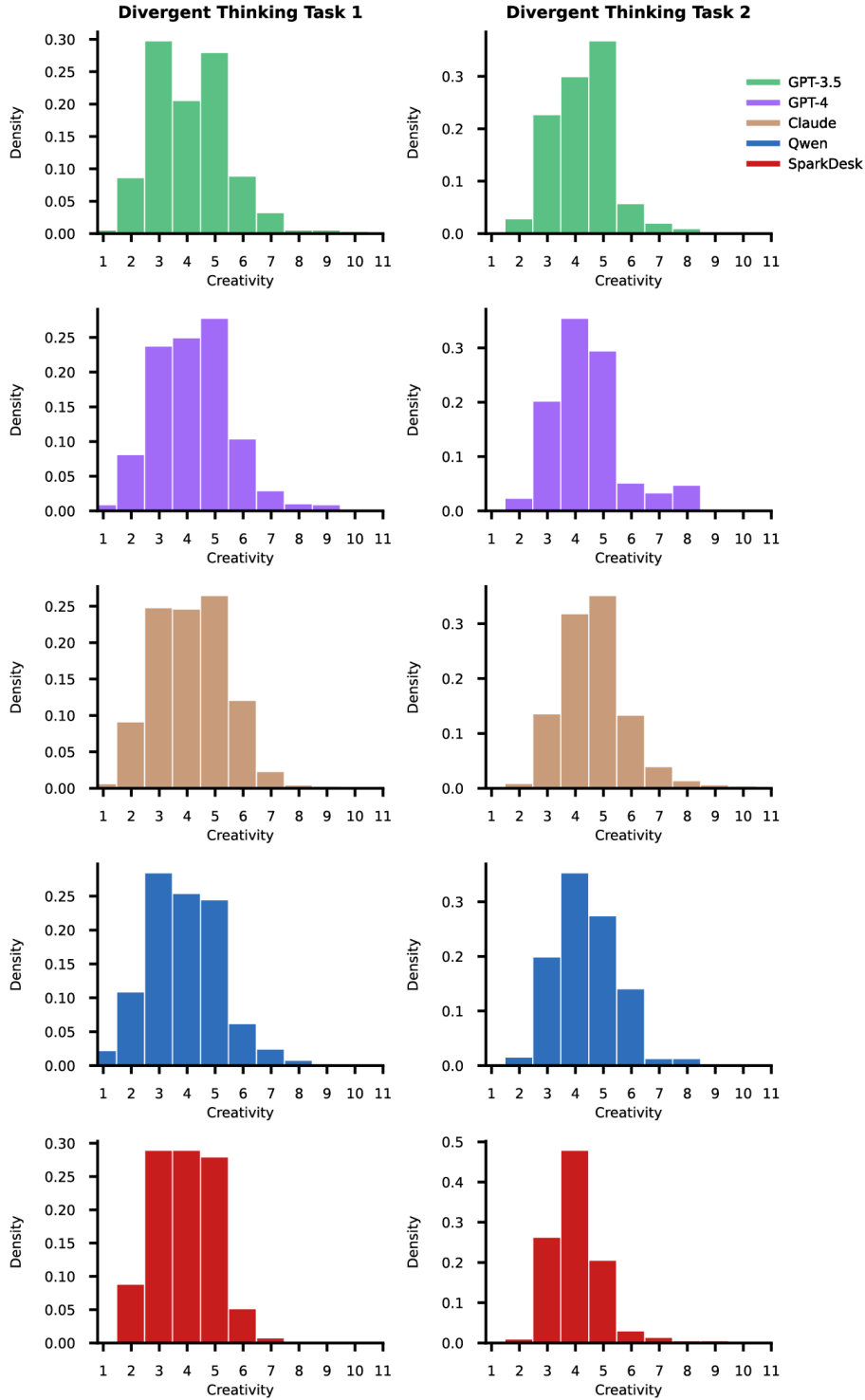

**Fig. S1. Densities of the creativity ratings for LLM responses in the divergent thinking tasks.**

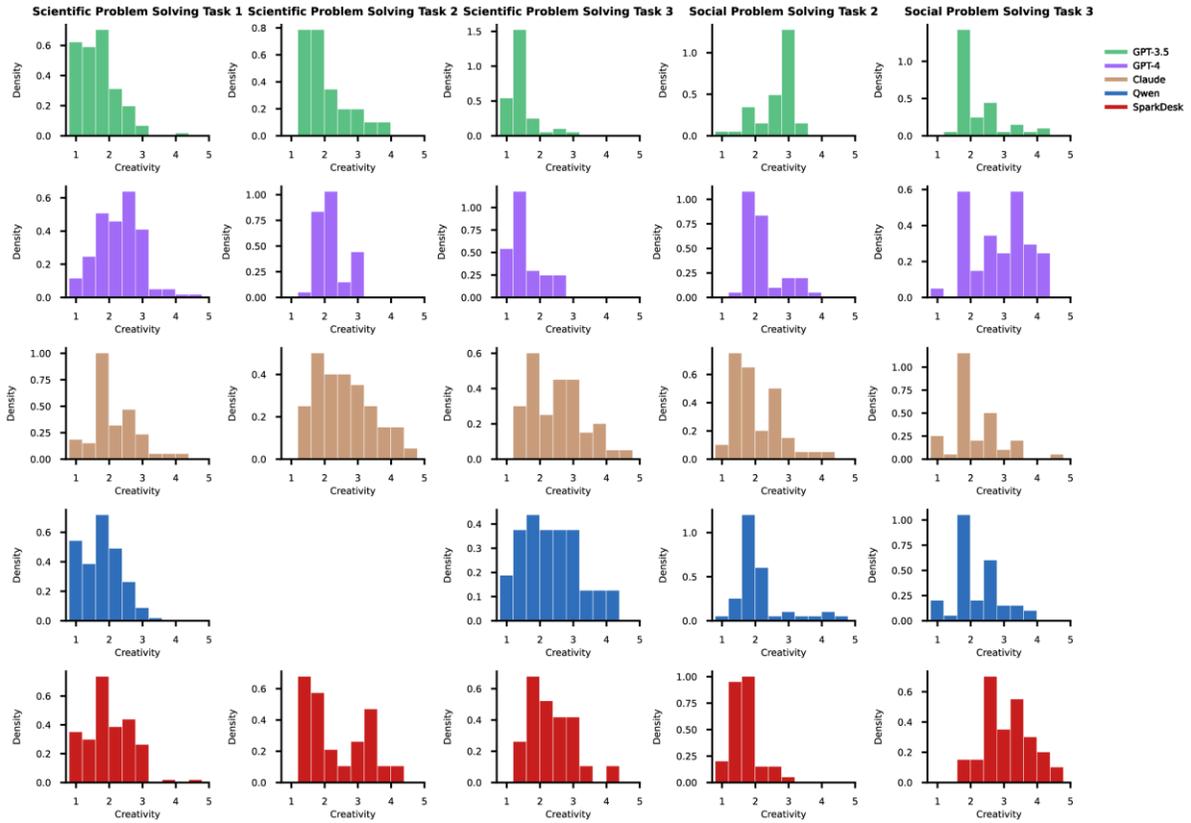

**Fig. S2. Densities of the creativity ratings for LLM responses in the problem solving tasks.**

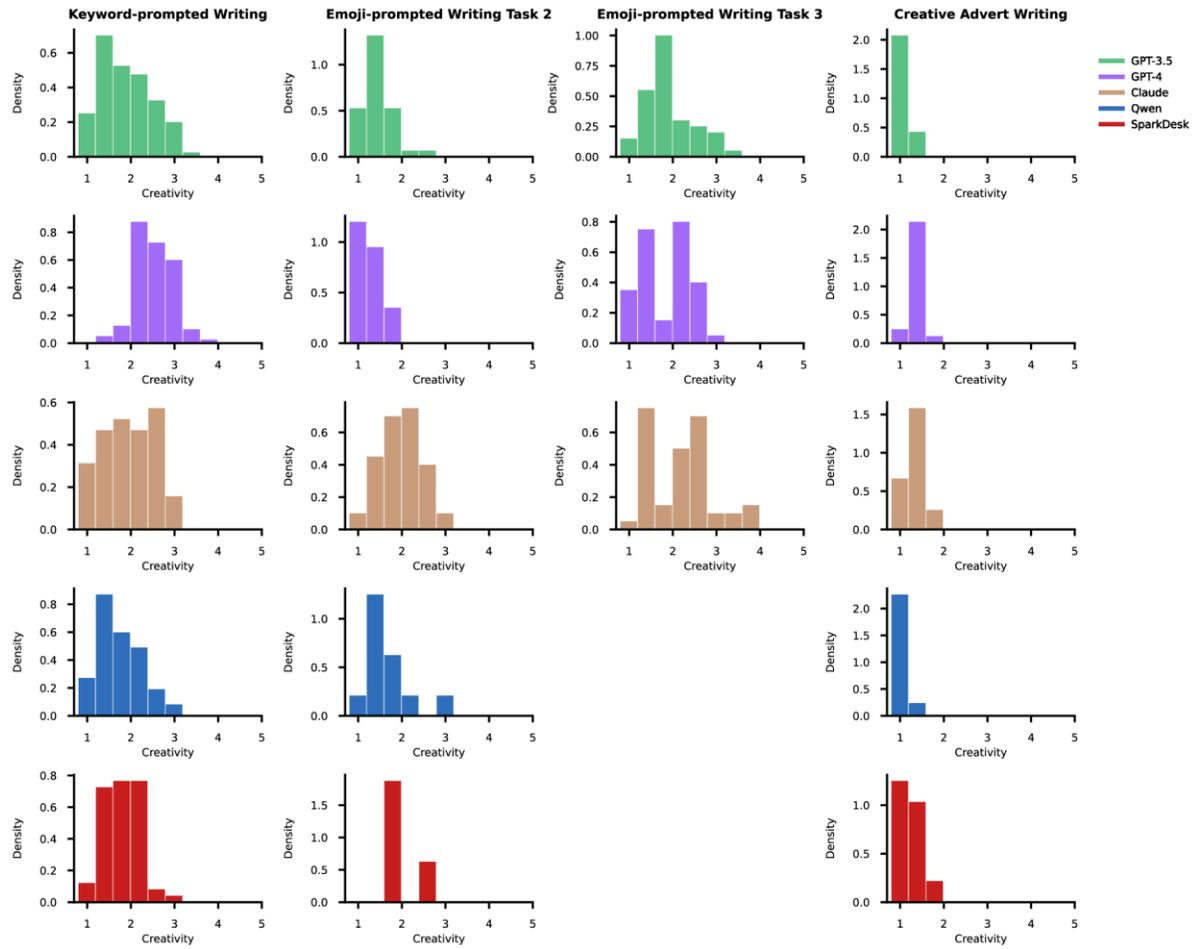

**Fig. S3. Densities of the creativity ratings for LLM responses in the creative writing tasks.**

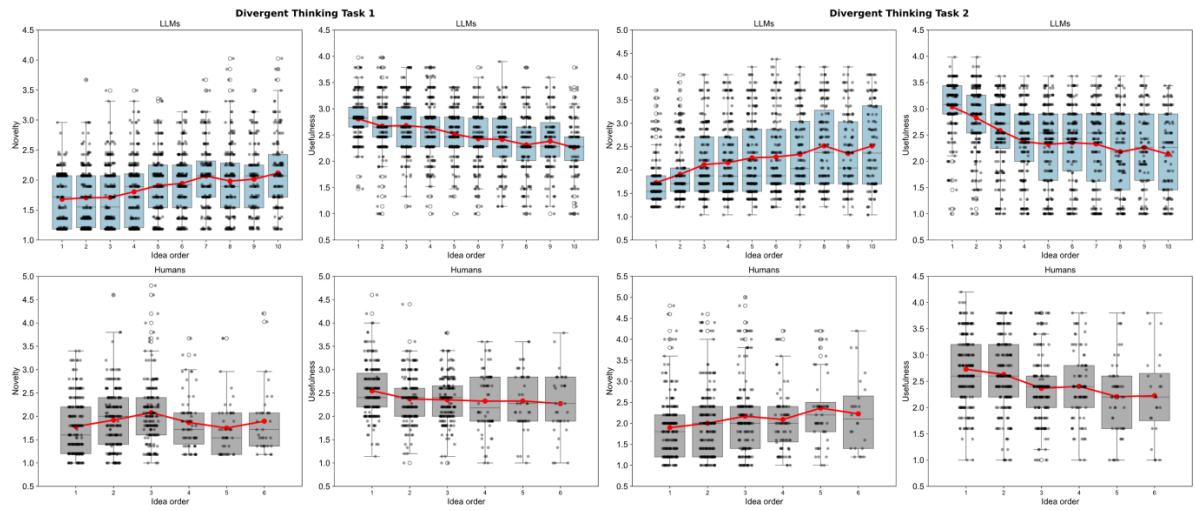

**Fig. S4. Ratings of novelty and usefulness for different ideas in each response for the divergent thinking tasks.**

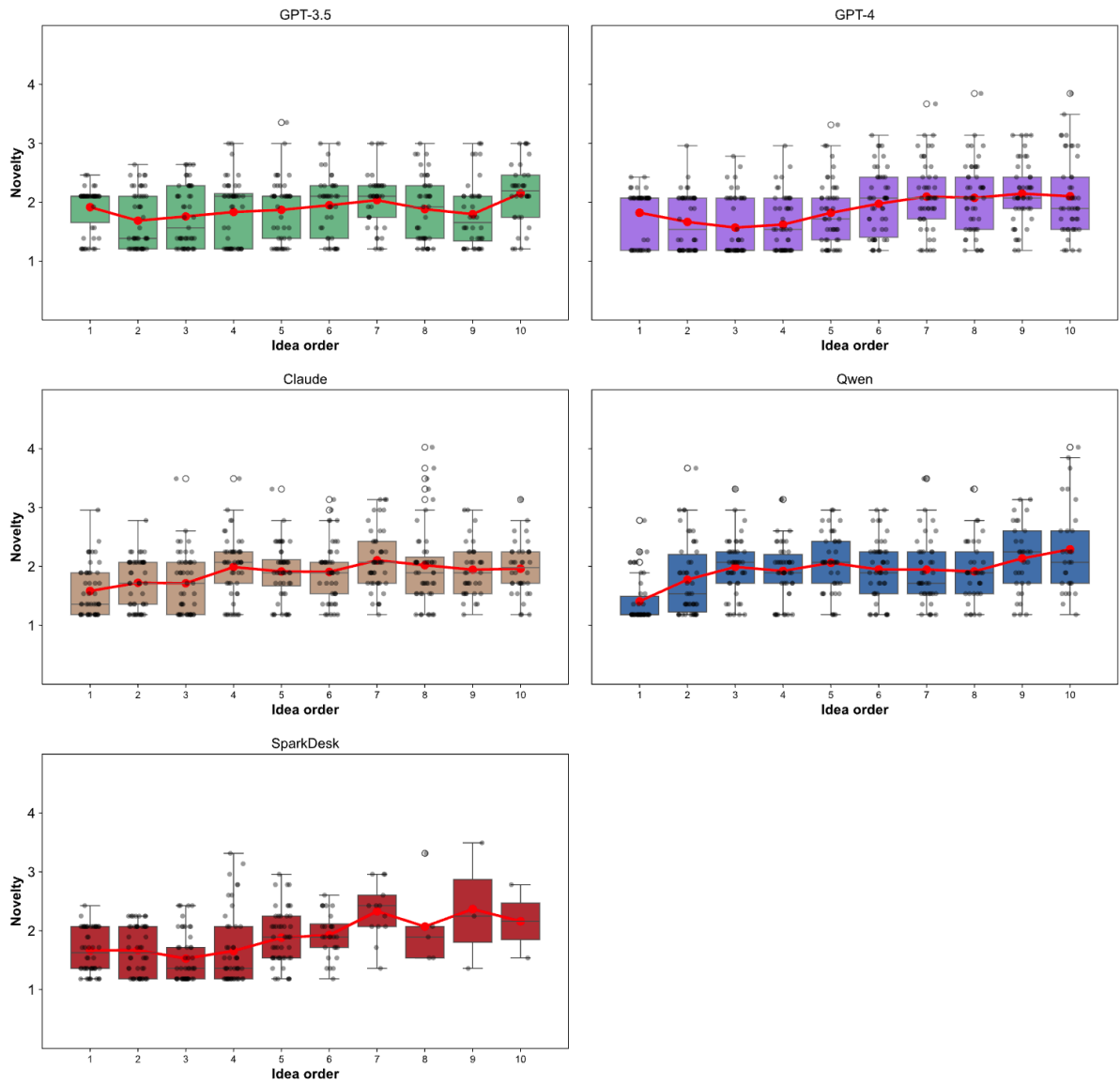

**Fig. S5. Novelty ratings for different ideas in each response for the Divergent Thinking Task 1.**

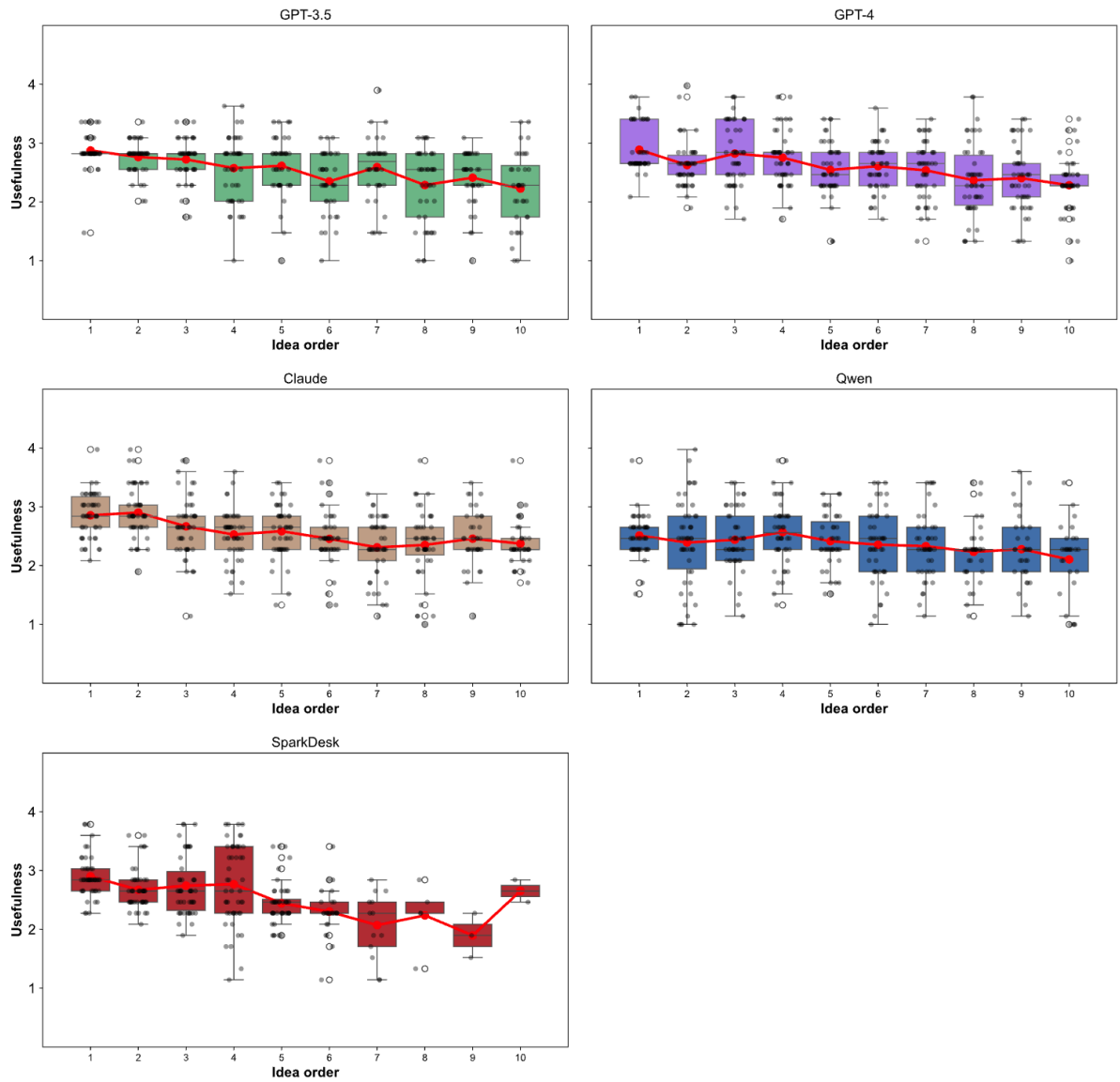

**Fig. S6. Usefulness ratings for different ideas in each response for the Divergent Thinking Task 1.**

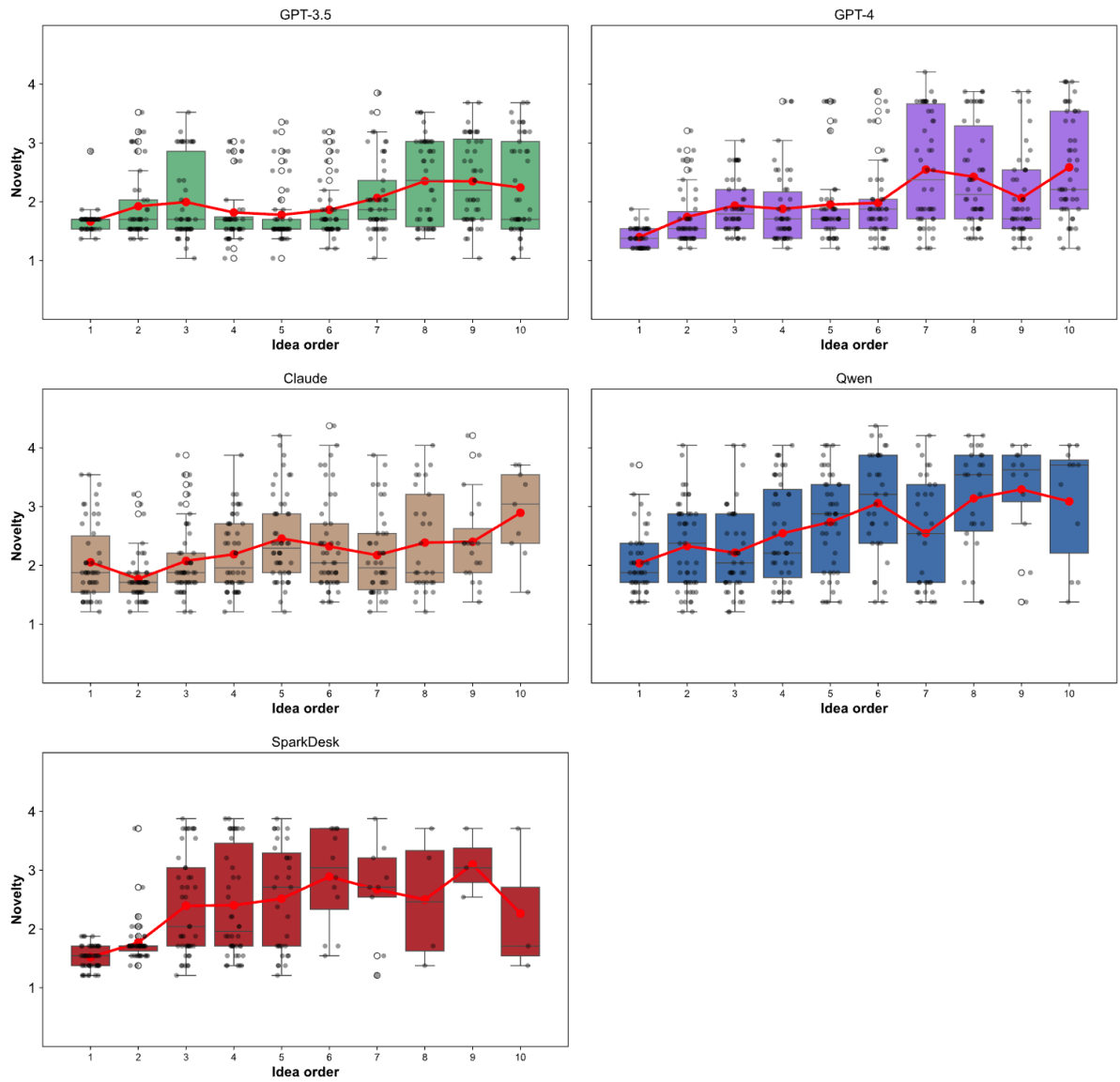

**Fig. S7. Novelty ratings for different ideas in each response for the Divergent Thinking Task 2.**

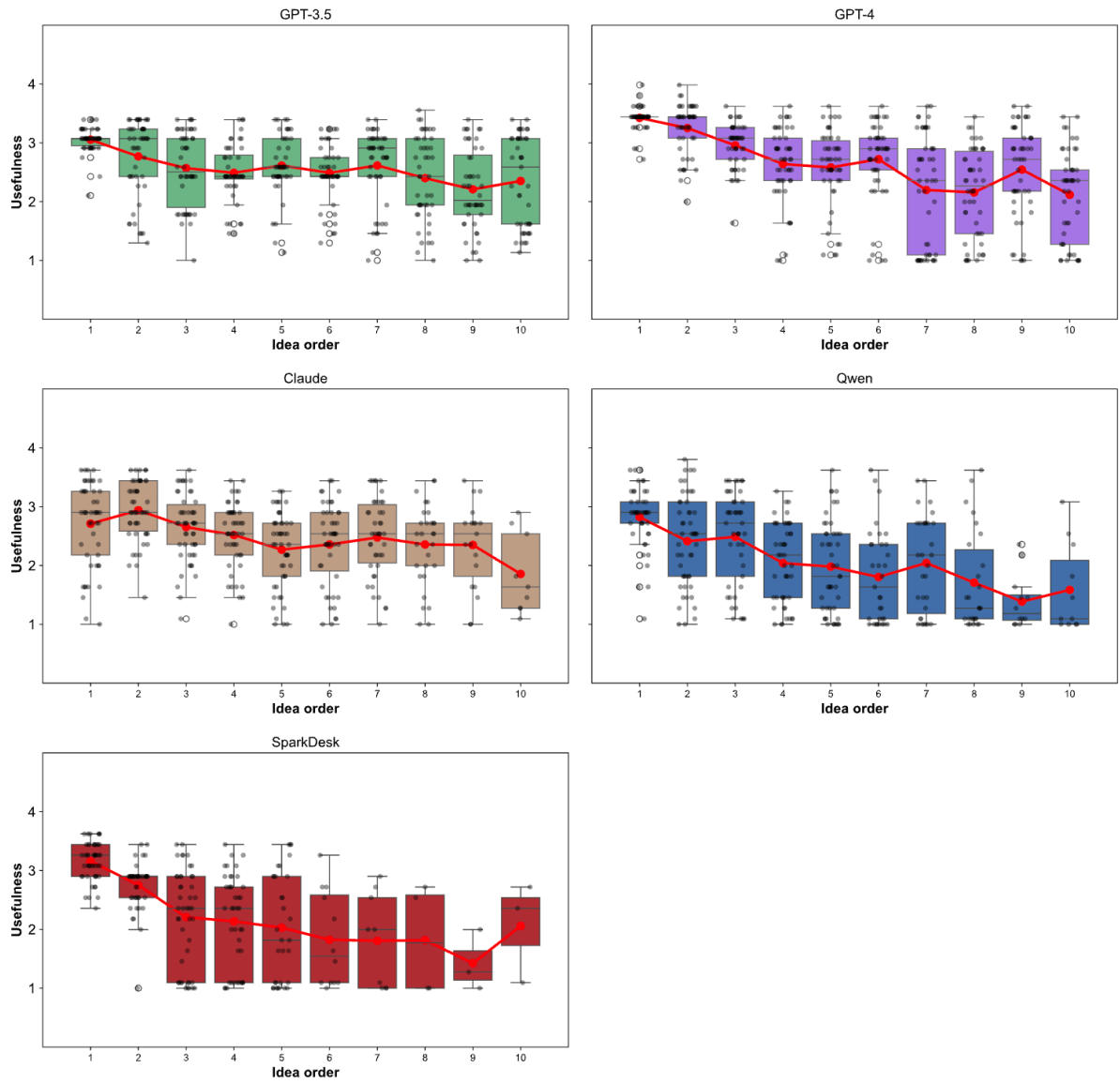

**Fig. S8. Usefulness ratings for different ideas in each response for the Divergent Thinking Task 2.**